%% file: bare_jrnl.tex
\documentclass[journal]{IEEEtran}
\usepackage{url}
\usepackage{cite}
\usepackage{balance}
\usepackage{graphicx}
\usepackage{amsfonts}
\usepackage{fancyhdr}
\usepackage{comment}
\usepackage{times}
\usepackage{amsmath}
\usepackage{changepage}
\usepackage{amssymb}
\usepackage{enumerate} 
\usepackage{algorithmic}
\usepackage{bm}
\usepackage{subfigure}
\usepackage{calligra}
\usepackage{multirow}
\usepackage{tabularx}
\usepackage{booktabs}
\usepackage{mathtools}
\usepackage{arydshln}
\usepackage{latexsym} 
\usepackage{amsmath}
\usepackage{amssymb}
\usepackage{color}
\usepackage[ruled,vlined]{algorithm2e}
\usepackage{subfigure}
\usepackage{ulem}
\usepackage{float}
\usepackage{tabstackengine}
\usepackage{siunitx}
\usepackage{colortbl}
\usepackage{hhline}
\usepackage{bookmark}

\newcommand{\darkgraytext}{\textcolor[rgb]{0.4,0.4,0.4}}

\begin{document}
\title{Robust Odometry and Mapping for Multi-LiDAR Systems with Online Extrinsic Calibration}
\author{Jianhao Jiao, Haoyang Ye, Yilong Zhu, and Ming Liu
\thanks{This work was supported by Collaborative Research Fund by Research Grants Council Hong Kong, 
under Project No. C4063-18G, Department of Science and Technology of Guangdong Province Fund, 
under Project No. GDST20EG54 and Zhongshan Municipal Science and Technology Bureau Fund, under project ZSST21EG06, awarded to Prof. Ming Liu.}
\thanks{The authors are with the Department of Electrical and Computer Engineering, 
Hong Kong University of Science and Technology, Hong Kong, China (email: \{jjiao, hyeab, yzhubr\}@connect.ust.hk, eelium@ust.hk).}
}
\markboth{}%
{Shell \MakeLowercase{\textit{et al.}}: Bare Demo of IEEEtran.cls for IEEE Journals}
\maketitle

\input{abstract}
\input{introduction}

\input{related_work}

\input{problem_statement}

\input{system_overview}
\input{measurement_preprocessing}

\input{initialization}

\input{multilo}

\input{mapping}

\input{experiment}

    \input{exper_implement}

    \input{exper_calibration}

    \input{exper_overview_localization}

        \input{exper_sr_localization}

        \input{exper_rhd_localization} 
        \input{exper_rv_localization}

    \input{exper_inject_perturbation}
    \input{exper_ablation_study}
\input{discussion}

\input{conclusion}

\input{appendix}

\input{acknowledgment}

\bibliographystyle{IEEEtran}
\bibliography{reference}{}
\end{document}

%% file: abstract.tex
\begin{abstract}
Combining multiple LiDARs enables a robot to maximize its perceptual awareness of environments and obtain sufficient measurements, which is promising for simultaneous localization and mapping (SLAM). 
This paper proposes a system to achieve robust and simultaneous extrinsic calibration, odometry, and mapping for multiple LiDARs.
Our approach starts with measurement preprocessing to extract edge and planar features from raw measurements.
After a motion and extrinsic initialization procedure, a sliding window-based multi-LiDAR odometry runs onboard to estimate poses with an online calibration refinement and convergence identification.
We further develop a mapping algorithm to construct a global map and optimize poses with sufficient features together with a method to capture and reduce data uncertainty.
We validate our approach's performance with extensive experiments on ten sequences (4.60km total length) for the calibration and SLAM and compare them against the state-of-the-art.
We demonstrate that the proposed work is a complete, robust, and extensible system for various multi-LiDAR setups.
The source code, datasets, and demonstrations are available at: \url{https://ram-lab.com/file/site/m-loam}.
\end{abstract}

\begin{IEEEkeywords}
Simultaneous localization and mapping, calibration and identification, sensor fusion, autonomous driving.
\end{IEEEkeywords}
\IEEEpeerreviewmaketitle

%% file: introduction.tex
\section{Introduction}
\label{sec:introduction}

\subsection{Motivation}
\IEEEPARstart{S}{imultaneous} Localization and Mapping (SLAM) is essential to a wide range of applications, 
such as scene reconstruction, robotic exploration, and autonomous driving \cite{thrun2002probabilistic,cadena2016past,barfoot2017state}.
Approaches that use only a LiDAR have attracted much attention from the research community due to their accuracy and reliability in range measurements.
However, LiDAR-based methods commonly suffer from data sparsity and limited vertical field of view (FOV) in real-world applications \cite{jiao2019automatic}.
For instance, LiDARs' points distribute loosely, which induces a mass of empty regions between two nearby scans. 
This characteristic usually causes state estimation to degenerate in structureless environments, such as narrow corridors and stairs \cite{ye2019tightly}. 
Recently, owing to the decreasing price of sensors, we have seen a growing trend of deploying multi-LiDAR systems on practical robotic platforms \cite{lyft2019,sun2020scalability,agarwal2020ford,barnes2020oxford,geyer2020a2d2,liu2021role}.
Compared with a single-LiDAR setup, the primary improvement of multi-LiDAR systems is the significant enhancement on the sensing range and density of measurements.
This benefit is practically useful for self-driving cars since we have to address the critical blind spots created by the vehicle body.
Thus, we consider multi-LiDAR systems in this paper.

\subsection{Challenges}
Despite their great advantages in environmental perception, a number of issues affect the development of SLAM using a multi-LiDAR setup.
\subsubsection{Precise and Flexible Extrinsic Calibration} 
Recovering the multi-LiDAR transformations for a new robotic platform is complicated. 
In many cases, professional users have to calibrate sensors in human-made surroundings \cite{jiao2019a} carefully. 
This requirement increases the cost to deploy and maintain a multi-LiDAR system for field robots.

It is desirable that the system can \textit{self-calibrate} the extrinsics in various environments online.
As shown in \cite{li2014high,yang2015monocular,qin2018vins}, benefiting from the simultaneous estimation of extrinsics and ego-motion, 
the working scope of visual-inertial systems has been expanded to drones and vessels in outdoor scenes.    
These approaches continuously perform calibration during the mission to guarantee the objective function to be always `optimal'. 
However, this process typically requires environmental or motion constraints to be fully observable. 
Otherwise, the resulting extrinsics may become suboptimal or unreliable. 
Thus, we have to fix the extrinsics if they are accurate, which creates a demand for online calibration with a convergence identification. 
In order to cope with the centimeter-level calibration error and unexpected changes on extrinsics over time, 
it is also beneficial to model the extrinsic perturbation for multi-LiDAR systems.

\subsubsection{Low Pose Drift}
To provide accurate poses in real time, 
state-of-the-art (SOTA) LiDAR-based methods \cite{zhang2014loam,shan2018lego,lin2020loam} commonly solve SLAM by two algorithms: 
\textit{odometry} and \textit{mapping}.
These algorithms are designed to estimate poses in a coarse-to-fine fashion.
Based on the original odometry algorithm, an approach that fully exploits multi-LiDAR measurements within a local window is required.
The increasing constraints help to prevent degeneracy or failure of frame-to-frame registration.
The subsequent mapping algorithm runs at a relatively low frequency and is given plenty of feature points and many iterations for better results. 
However, as we identify in Section \ref{sec:mapping}, many SOTA approaches neglect the fact that uncertain points in the global map limit the accuracy. 
To minimize this adverse effect, we must develop a method to capture map points' uncertainties and reject outliers.

\subsection{Contributions}
\label{sec:introduction_contribution}
To tackle these challenges, we propose M-LOAM, a robust system for \textbf{M}ulti-\textbf{L}iDAR extrinsic calibration, real-time \textbf{O}dometry, and \textbf{M}apping. 
Without manual intervention, our system can start with several extrinsic-uncalibrated LiDARs, 
automatically calibrate their extrinsics, and provide accurate poses as well as a globally consistent map.
Our previous work \cite{ye2019tightly} proposed sliding window-based odometry to fuse LiDAR points with high-frequency IMU measurements.
That framework inspires this paper, where we try to solve the problem of multi-LiDAR fusion.
In addition, we introduce a motion-based approach \cite{jiao2019automatic} to initialize extrinsics, 
and employ the tools in \cite{barfoot2014associating} to represent uncertain quantities.
Our design of M-LOAM presents the following \textit{contributions}:
\begin{enumerate}
	\item Automatic initialization that computes all critical states, including motion between consecutive frames as well as extrinsics for subsequent phases. 
	It can start at arbitrary positions without any prior knowledge of the mechanical configuration or calibration objects (Section \ref{sec:initialization}).
	\item Online self-calibration with a general convergence criterion is executed simultaneously with the odometry. 
	It has the capability to monitor the convergence and trigger termination in a fully unsupervised manner (Section \ref{sec:multilo_with_online_calibration}).
	\item Sliding window-based odometry that fully exploits information from multiple LiDARs.
	This implementation can be explained as small-scale frame-to-map registration, 
	which further reduces the drift accumulated by the consecutive frame-to-frame odometry (Section \ref{sec:multilo_with_pure_odometry}).	
	\item Mapping with a two-stage approach that captures and propagates uncertain quantities from sensor noise, degenerate pose estimation, and extrinsic perturbation. 
	This approach enables the mapping process with an awareness of uncertainty and helps us to maintain the consistency of a global map as well as boost the robustness of a system for long-duration navigation tasks (Section \ref{sec:mapping}).
\end{enumerate}

\begin{figure}
	\centering
	\includegraphics[width=0.98\linewidth]{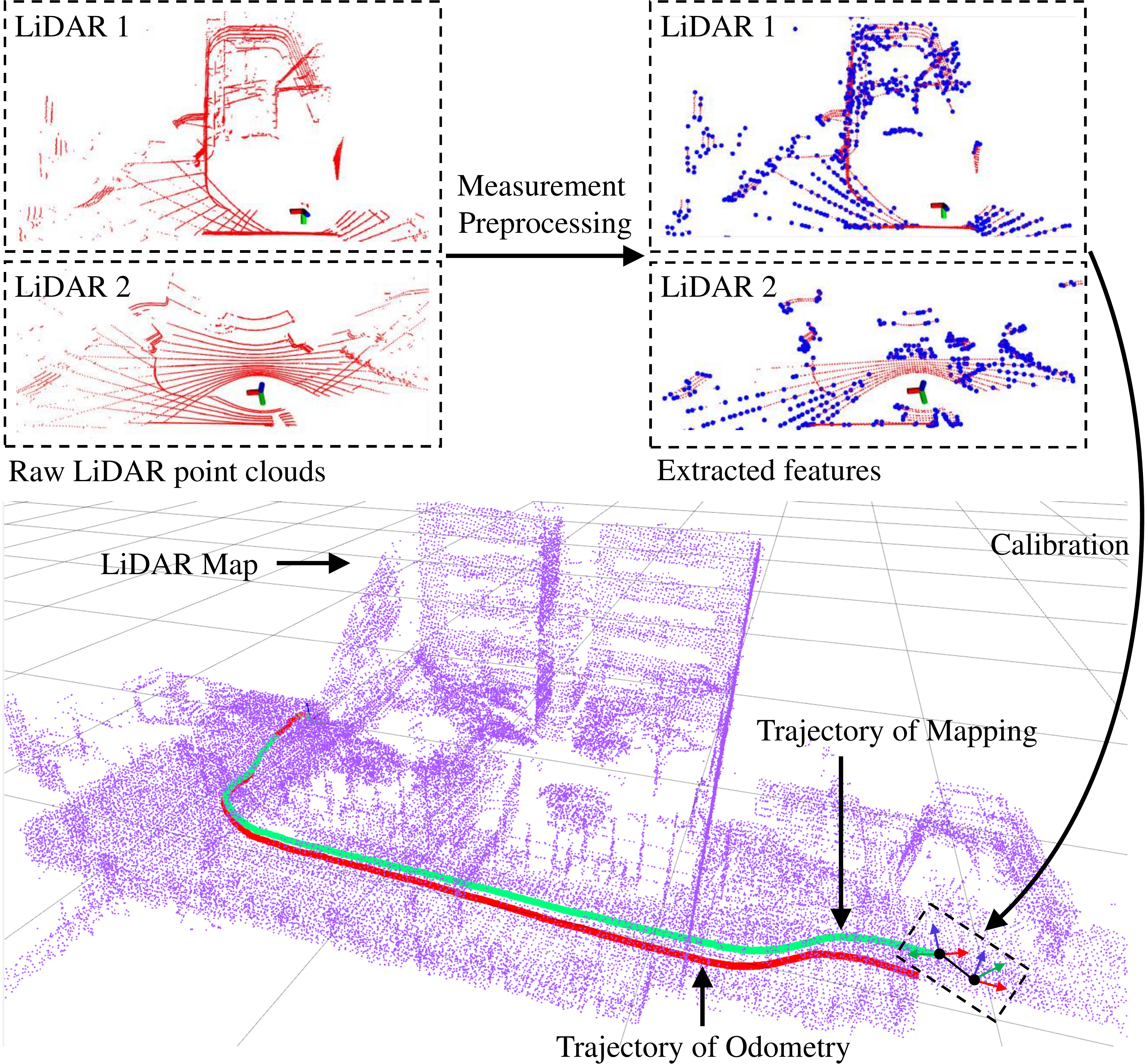}       
	\caption{We visualize the immediate results of M-LOAM. 
			The raw point clouds perceived by different LiDARs are denoised and extracted with edge (blue dots) and planar (red dots) features, 
			which are shown at the top-right position. 
			The proposed online calibration is performed to obtain accurate extrinsics.
			After that, the odometry and mapping algorithms use these features to estimate poses.
			The trajectory of the mapping (green) is more accurate than that of the odometry (red).}
	\label{fig:demo_visualization}
\end{figure}

To the best of our knowledge, M-LOAM is the first complete solution to multi-LiDAR calibration and SLAM. 
The system is evaluated under extensive experiments on both handheld devices and autonomous vehicles, 
covering various scenarios from indoor offices to outdoor urban roads, and outperforms the SOTA LiDAR-based methods.
Regarding the calibration on diverse platforms, our approach achieves an extrinsic accuracy of centimeters in translation and deci-degrees in rotation. 
For the SLAM in different scales, M-LOAM has been successfully applied to provide accurate poses and map results.
Fig. \ref{fig:demo_visualization} visualizes M-LOAM's output at each stage.
To benefit the research community, we publicly release our code, implementation details, and the multi-LiDAR datasets.

\subsection{Organization}
The rest of the paper is organized as follows. 
Section \ref{sec:related_work} reviews the relevant literature. 
Section \ref{sec:problem_statement} formulates the problem and provides basic concepts.
Section \ref{sec:overview} gives the overview of the system.
Section \ref{sec:preprocessing} describes the preprocessing module on LiDAR measurements.
Section \ref{sec:initialization} introduces the motion and extrinsic initialization procedure.
Section \ref{sec:multilo} presents the tightly coupled, sliding window-based multi-LiDAR odometry (M-LO) with online calibration refinement. 
The uncertainty-aware multi-LiDAR mapping algorithm is introduced in Section \ref{sec:mapping}, 
followed by experimental results in Section \ref{sec:experiment}.
In Section \ref{sec:discussion}, we provide a discussion about the proposed system. 
Finally, Section \ref{sec:conclusion} concludes this paper with possible future research directions. 

%% file: related_work.tex
\section{Related Work}
\label{sec:related_work}
Scholarly works on SLAM and extrinsic calibration are extensive. 
In this section, we briefly review relevant results on LiDAR-based SLAM and online calibration methods for multi-sensor systems.

\subsection{LiDAR-Based SLAM}
Generally, many SOTA are developed from the iterative closest point (ICP) algorithm \cite{pomerleau2013comparing,segal2009generalized,serafin2015nicp,magnusson2007scan}, into which the methods including 
measurement preprocessing, degeneracy prediction, and sensor fusion are incorporated.
These works have pushed the current LiDAR-based systems to become fast, robust, and feasible to large-scale environments.

\subsubsection{Measurement Preprocessing}
\label{sec:related_work_preprocessing}
As the front-end of a system, the measurement preprocessing encodes point clouds into a compact representation. 
We categorize related algorithms into either dense or sparse methods. 
As a typical dense method, SuMa \cite{behley2018efficient} demonstrates the advantages of utilizing surfel-based maps for registration and loop detection. 
Its extended version \cite{chen2019suma++} incorporates semantic constraints into the original cost function. 
The process of matching dense pixel-to-pixel correspondences make the method computationally intensive, thus requiring dedicated hardware (GPU)
for real-time operation.
These approaches are not applicable to our cases since we have to frequently perform the registration on a CPU.

Sparse methods prefer to extract geometric features from raw measurements and are thus supposed to have real-time performance.
Grant et al. \cite{grant2013finding} proposed a plane-based registration, while 
Velas et al. \cite{velas2016collar} represented point clouds as collar line segments.
Compared to them, LOAM \cite{zhang2014loam} has fewer assumptions about sensors and surroundings. 
It selects distinct points from both edge lines and local planar patches.
Recently, several following methods have employed ground planes \cite{pandey2017alignment,shan2018lego}, visual detection \cite{chen2020sloam}, 
probabilistic grid maps \cite{ji2018cpfg}, good features \cite{jiao2021greedy}, or directly used dense scanners \cite{bosse2012zebedee,pfrunder2017real,lin2020loam} 
to improve the performance of sparse approaches in noisy or structure-less environments.
To limit the computation time when more LiDARs are involved, we extract the sparse edge and planar features like LOAM. 
But differently, we represent their residuals in a more concise and unified way.

\subsubsection{Degeneracy in State Estimation}
\label{sec:related_work_degeneracy}
The geometric constraints of features are formulated as a state estimation problem. 
Different methods have been proposed to tackle the degeneracy issue.
Zhang et al. \cite{zhang2016degeneracy} defined a \textit{factor}, which is equal to the minimum eigenvalue of information matrices, to determine the system degeneracy.
They also proposed a technique called \textit{solution remapping} to update variables in well-conditioned directions.
This technique was further applied to tasks including localization, registration, pose graph optimization, and inspection of sensor failure \cite{hinduja2019degeneracy,zhang2017enabling,zhen2019estimating}.
To quantify a nonlinear system's observability, Rong et al. \cite{rong2016detection} computed the condition number of the empirical observability Gramian matrix.
Since our problem linearizes the objective function as done in Zhang et al. \cite{zhang2016degeneracy}, 
we also introduce \textit{solution remapping} to update states.
Additionally, our online calibration method employs the \textit{degeneracy factor} as a quantitative metric of the extrinsic calibration.

\subsubsection{Multi-Sensor Fusion}
\label{sec:related_work_sensor_fusion}
Utilizing multiple sensors to improve the motion-tracking performance of single-LiDAR odometry is promising.
Most existing works on LiDAR-based fusion combine visual or inertial measurements.
The simplest way to deal with multi-modal measurements is loosely coupled fusion, where each sensor's pose is estimated independently. 
For example, LiDAR-IMU fusion is usually done by the extended Kalman filter (EKF) \cite{lynen2013robust,wan2018robust,zhao2019a}. 
Tightly coupled algorithms that optimize sensors' poses by jointly exploiting all measurements have become increasingly prevalent in the community. 
They are usually done by either the EKF \cite{zuo2019lic,qin2020lins,hemann2016long} or batch optimization \cite{ye2019tightly,lowe2018complementary,le2019in2lama,le2020in2laama,zheng2019low}.
Besides multi-modal sensor fusion, another approach is to explore the fusion of multiple LiDARs (uni-model),
which is still an open problem.
    
Multiple LiDARs improve a system by maximizing the sensing coverage against extreme occlusion.
Inspired by the success of tightly coupled algorithms, 
we achieve a sliding window estimator to optimize states of multiple LiDARs.
This mode has great advantages for multi-LiDAR fusion.

\subsection{Multi-Sensor Calibration}
\label{sec:related_work_calibration}
Precise extrinsic calibration is important to any multi-sensor system. 
Traditional methods \cite{he2013pairwise,choi2015extrinsic,jiao2019a,xue2019automatic} have to run an ad-hoc calibration procedure before a mission. 
This tedious process needs to be repeated whenever there is slight perturbation on the structure.
A more flexible solution to estimate these parameters is combined SLAM-based techniques. 
Here, extrinsics are treated as one of the state variables and optimized along with sensors' poses.
This scheme is also applicable to some non-stationary parameters, such as robot kinematics, IMU biases, and time offsets.

Kummerle et al. \cite{kummerle2011simultaneous} pioneered a hyper-graph optimization framework to calibrate an onboard laser scanner with wheel encoders. 
Their experiments reveal that online correction of parameters leads to consistent and accurate results.
Teichman et al. \cite{teichman2013unsupervised}, meanwhile, proposed an iterative SLAM-fitting pipeline to resolve the distortion of two RGB-D cameras. 
To recover spatial offsets of a multi-camera system, Heng et al. \cite{heng2015self} formulated the problem as a bundle adjustment, 
while Ouyang et al. \cite{ou2020online} employed the Ackermann steering model of vehicles to constrain extrinsics.
As presented in \cite{qin2018online,le20183d,le2020in2laama,qiu2020real}, 
the online estimation of time offsets among sensors is crucial to IMU-centric systems.
Qin et al. \cite{qin2018online} utilized reprojection errors of visual features to formulate the temporal calibration problem, 
while Qiu et al. \cite{qiu2020real} proposed a more general method to calibrate heterogeneous sensors by analyzing sensors' motion correlation.

This paper implicitly synchronizes the time of multiple LiDARs with an hardware-based external clock
and explicitly focuses on the extrinsic calibration. 
Our approach consists of an online procedure to achieve flexible multi-LiDAR extrinsic calibration. 
To monitor the convergence of estimated extrinsics, we propose a general criterion.
In addition, we model the extrinsic perturbation to reduce its negative effect for long-term navigation tasks.

%% file: problem_statement.tex
\section{Problem Statement}
\label{sec:problem_statement}
We formulate M-LOAM in terms of the Maximum Likelihood Estimation (MLE). 
The MLE leads to a nonlinear optimization problem where the inverse of the Gaussian covariances weights the residual functions.
Before delving into details of M-LOAM, we first introduce some basic concepts.
Section \ref{sec:notation} presents notations.
Section \ref{sec:problem} introduces the MLE,
and Section \ref{sec:uncertainty} describes suitable models to represent uncertain measurements in $\mathbb{R}^{3}$ and transformations in $SE(3)$.
Finally, Section \ref{sec:implementation} briefly presents the implementation of the MLE with approximate Gaussian noise in M-LOAM.

\subsection{Notations and Definitions}
\label{sec:notation}
The nomenclature is shown in Table \ref{tab:nomenclature}.
We consider a system that consists of one primary LiDAR and multiple auxiliary LiDARs. 
The primary LiDAR defines the base frame, and we use $()^{l^{1}}/()^{b}$ to indicate it. 
The frames of the auxiliary LiDARs are denoted by $()^{l^{i,i>1}}$.
We denote $\mathcal{F}$ as the set of available features extracted from the LiDARs' raw measurements. 
Each feature is represented as a point in 3D space: $\mathbf{p}=[x,y,z]^{\top}$.
The state vector, composed of translational and rotational parts, 
is denoted by $\mathbf{x}=[\mathbf{t},\mathbf{q}]$ where $\mathbf{t}$ is a $3\times1$ vector, and $\mathbf{q}$ is the Hamilton quaternion. 
But in the case that we need to rotate a vector, we use the $3\times3$ rotation matrix $\mathbf{R}$ in the \textit{Lie group} \textit{SO(3)}.
We can convert $\mathbf{q}$ into $\mathbf{R}$ with the Rodrigues formula \cite{sola2017quaternion}.
Section \ref{sec:mapping} associates uncertainty with poses on the vector space, 
where we use the $4\times4$ transformation matrix $\mathbf{T}$ in the \textit{Lie group} \textit{SE(3)} to represent a pose.
A rotation matrix and translation vector can be associated with a transformation matrix as
$
\mathbf{T}
=
\begin{bmatrix}
    \mathbf{R} & \mathbf{p}\\
    \mathbf{0}^{\top} & 1
\end{bmatrix}
$.

\subsection{Maximum Likelihood Estimation}
\label{sec:problem}
We formulate the pose and extrinsic estimation of a multi-LiDAR system as an MLE problem \cite{ila2015fast}:
\begin{equation}
\label{equ:mle}
\begin{split}
    \hat{\mathbf{x}}_{k}
    =
    \underset{\mathbf{x}_{k}}{\arg\max}\ 
    p(\mathcal{F}_{k}|\mathbf{x}_{k})
    =
    \underset{\mathbf{x}_{k}}{\arg\min}\ 
    f(\mathbf{x_{k}}, \mathcal{F}_{k}),
\end{split}
\end{equation}
where $\mathcal{F}_{k}$ represents the available features at the $k^{th}$ frame, $\mathbf{x}_{k}$ is the state to be optimized, 
and $f(\cdot)$ is the objective function.
Assuming the measurement model to be subjected to Gaussian noise \cite{barfoot2017state}, problem \eqref{equ:mle} becomes a 
nonlinear least-squares (NLS) problem:
\begin{equation}
\label{equ:nls}
\begin{split}
    \hat{\mathbf{x}}_{k}
    =
    \underset{\mathbf{x}_{k}}{\arg\min}\ 
    \sum_{i=1}^{m}
    \rho
    \Big(\big|\big|\mathbf{r}(\mathbf{x}_{k}, \mathbf{p}_{ki})\big|\big|^{2}_{\bm{\Sigma}_{i}}
    \Big),
\end{split}
\end{equation}
where $\rho(\cdot)$ is the robust Huber loss to handle outliers\cite{bosse2016robust}, 
$\mathbf{r(\cdot)}$ is the residual function, 
and $\bm{\Sigma}_{i}$ is the covariance matrix.  
Iterative methods such as Gauss-Newton or Levenberg-Marquardt can offen be used to solve this NLS problem.
These methods locally linearize the objective function by computing the Jacobian w.r.t. $\mathbf{x}_{k}$ as $\mathbf{J}=\partial f/\partial\mathbf{x}_{k}$.
Given an initial guess, $\mathbf{x}_{k}$ is iteratively optimized by usage of $\mathbf{J}$ until converging to a local minima.
At the final iteration, the least-squares covariance of the state is calculated as $\bm{\Xi}=\bm{\Lambda}^{-1}$ \cite{censi2007accurate}, 
where $\bm{\Lambda}=\mathbf{J}^{\top}\mathbf{J}$ is called the \textit{information matrix}.

\subsection{Uncertainty Representation}
\label{sec:uncertainty}
We employ the tools in \cite{barfoot2014associating} to represent data uncertainty.
We first represent a noisy LiDAR point as
\begin{equation}
\label{equ:point_perturbation_3_1}
    \mathbf{p}
    =
    \bar{\mathbf{p}} + \bm{\zeta},\ \ \ \bm{\zeta}\sim\mathcal{N}(\mathbf{0},\mathbf{Z}),
\end{equation}
where $\bar{\mathbf{p}}$ is a noise-free vector, $\bm{\zeta}\in\mathbb{R}^{3}$ is a small Gaussian perturbation 
variable with zero mean, 
and $\mathbf{Z}$ is a noise covariance of LiDAR measurements.
To make \eqref{equ:point_perturbation_3_1} compatible with transformation matrices (i.e., $\mathbf{p}_{h}'=\mathbf{T}\mathbf{p}_{h}$), 
we also express it with $4\times 1$ homogeneous coordinates:
\begin{equation}
    \label{equ:point_perturbation_4_1}  
    \mathbf{p}_{h}
    =
    \begin{bmatrix}
        \bar{\mathbf{p}}\\
        1
    \end{bmatrix}
    + \mathbf{D}\bm{\zeta}
    =
    \bar{\mathbf{p}}_{h}+ \mathbf{D}\bm{\zeta},
    \ \ \ \bm{\zeta}\sim\mathcal{N}(\mathbf{0},\mathbf{Z}),
\end{equation}
where $\mathbf{D}$ is a matrix that converts a $3\times 1$ vector into homogeneous coordinates.
As investigated in \cite{pomerleau2012noise}, the LiDARs' depth measurement error (also called sensor noise) is primarily affected by the target distances.
$\mathbf{Z}$ is simply set as a constant matrix.\footnote{$\mathbf{Z}=\text{diag}(\sigma^2_{x},\sigma^2_{y},\sigma^2_{z})$, where $\sigma_{x}, \sigma_{y}, \sigma_{z}$ are standard deviations of LiDARs' depth noise along different axes. We refer to manuals to obtain them for a specific LiDAR.}
We then define a random variable in $SE(3)$ with a small perturbation according to\footnote{The $(\cdot)^{\wedge}$ operator turns $\bm{\xi}$ into a member of the \textit{Lie algebra} $\mathfrak{se}(3)$. 
The \textit{exponential map} associates an element of $\mathfrak{se}(3)$ to a transformation matrix in $SE(3)$. 
Similarly, we can also use $(\cdot)^{\wedge}$ and $\exp(\cdot)$ to associate $3\times 1$ vector $\bm{\phi}$ with a rotation matrix in $SO(3)$.
\cite{barfoot2014associating} provides detailed expressions.}
\begin{equation}
    \label{equ:se3_perturbation}
    \mathbf{T} 
    = 
    \exp(\bm{\xi^{\wedge}})\bar{\mathbf{T}},
    \ \ \ \bm{\xi}\sim\mathcal{N}(\mathbf{0}, \bm{\Xi}),
\end{equation}
where $\bar{\mathbf{T}}$ is a noise-free transformation and $\bm{\xi}\in\mathbb{R}^{6}$ is the small perturbation variable with covariance $\bm{\Xi}$. 
This representation allows us to store the mean transformation as $\bar{\mathbf{T}}$ and use $\bm{\xi}$ for perturbation on the vector space. 
We consider that $\bm{\Xi}$ indicates two practical error sources:
\begin{itemize}
    \item \textbf{Degenerate Pose Estimation} arises from cases such as lack of geometrical structures in poorly constrained environments \cite{zhang2016degeneracy}. 
    It typically makes poses uncertain in their degenerate directions \cite{censi2007accurate,brossard2020new}.
    Existing works resort to model-based and learning-based \cite{landry2019cello} methods to estimate pose covariances in the context of ICP.
    or vibration, impact, and temperature drift during long-term operation. 
    \item \textbf{Extrinsic Perturbation} always exists due to calibration errors \cite{jiao2019a}, 
    Wide baseline sensors such as stereo cameras, especially, suffer even more extrinsic deviations than standard sensors. 
    Such perturbation is detrimental to the measurement accuracy \cite{maddern20171,jiao2020mlod} of multi-sensor systems but is hard to measure.
\end{itemize}

The computation of $\bm{\Xi}$ is detailed in Section \ref{sec:mapping}.

\begin{table}[]
    \centering
    \caption{Nomenclature}
    \renewcommand\arraystretch{1}
    \renewcommand\tabcolsep{2pt}        
    \begin{tabularx}{0.47\textwidth}{cX}
        \toprule
        Notation & Explanation \\ 
        \midrule
        & \textbf{Coordinate System and Pose Representation}\\
        \midrule
        $()^{w}, ()^{b}, ()^{l^{i}}$ & Frame of the world, primary sensor, and $i^{th}$ LiDAR\\
        $()^{l^{i}_{k}}$ & Frame of the $i^{th}$ LiDAR while taking the $k^{th}$ point cloud\\                
        $I$ & Number of LiDARs in a multi-LiDAR system\\
        $\mathbf{p}$, $\mathbf{p}_{h}$ & 3D Point in $\mathbb{R}^{3}$ and homogeneous coordinates\\
        $\mathbf{x}$ & State vector\\       
        $\mathbf{t}$ & Translation vector in $\mathbb{R}^{3}$\\     
        $\mathbf{q}$ & Hamilton quaternion\\                
        $\mathbf{R}$ & Rotation matrix in the \textit{Lie group} $SO(3)$\\        
        $\mathbf{T}$ & Transformation matrix in the \textit{Lie group} $SE(3)$\\            
        $\bm{\xi}, \bm{\zeta},\bm{\theta}$ & Gaussian noise variable\\
        $\bm{\Sigma}$ & Covariance matrix of residuals \\
        $\mathbf{Z}$ & Covariance matrix of LiDAR depth measurement noise \\
        $\bm{\Xi}$ & Covariance matrix of pose noise \\
        \midrule
        & \textbf{Odometry and Mapping}\\
        \midrule        
        $\mathcal{F}$ & Features extracted from raw point clouds\\
        $\mathcal{E}, \mathcal{H}$ & Edge and planar subset of extracted features\\ 
        $L, \Pi$ & Edge line and planar patch\\     
        $[\mathbf{w}, d]$ & Coefficient vector of a planar patch\\      
        $\mathcal{M}$ & Local map\\
        $\mathcal{G}$ & Global map\\        
        $\lambda$ & Degeneracy factor\\ 
        \bottomrule
    \end{tabularx}
    \label{tab:nomenclature}
\end{table}

\begin{figure*}
    \centering
    \includegraphics[width=0.98\textwidth]{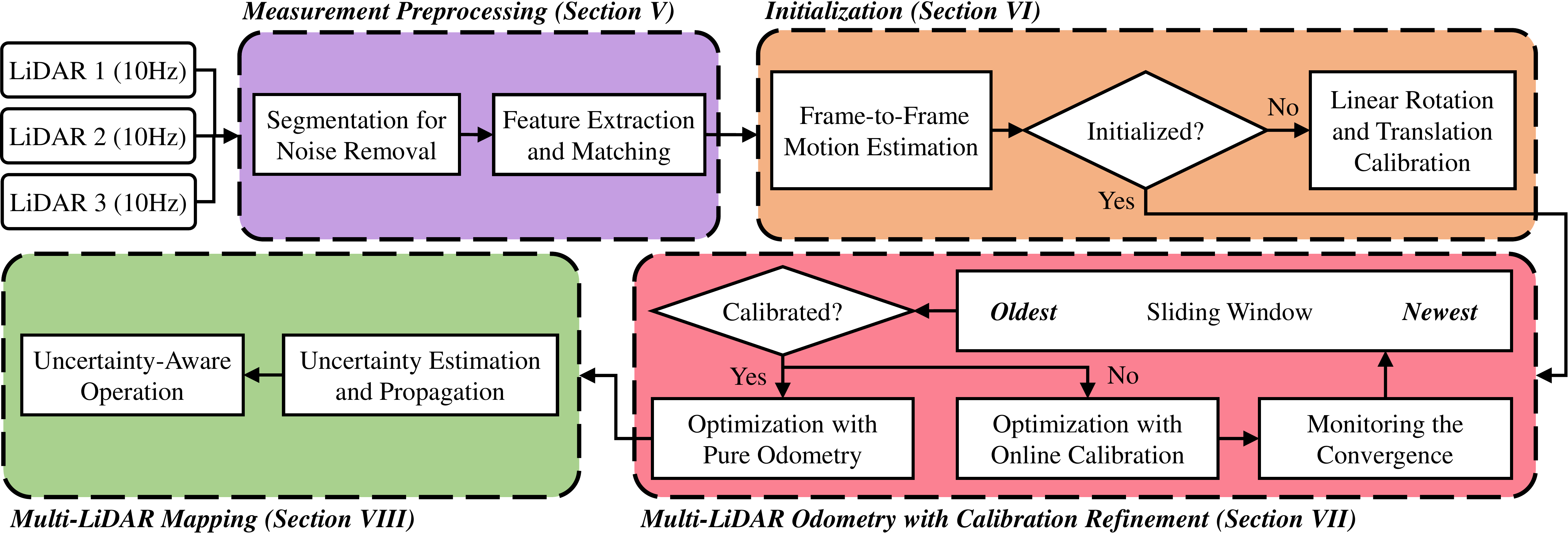}       
    \caption{The block diagram illustrating the full pipeline of the proposed M-LOAM system. The system starts with measurement preprocessing (Section \ref{sec:preprocessing}). 
             The initialization module (Section \ref{sec:initialization}) initializes values for the subsequent nonlinear optimization-based multi-LiDAR odometry with calibration refinement (Section \ref{sec:multilo}). 
             According to the convergence of calibration, the optimization is divided into two subtasks: online calibration and pure odometry. 
             If the calibration converges, we can skip the extrinsic initialization and refinement steps, and 
             enter the pure odometry and mapping phase.
             The uncertainty-aware multi-LiDAR mapping (Section \ref{sec:mapping}) maintains a globally consistent map 
             to decrease the odometry drift and remove noisy points.}
    \label{fig:system_pipeline}     
\end{figure*} 

\subsection{MLE with Approximate Gaussian Noise in M-LOAM}
\label{sec:implementation}
We extend the MLE to design multiple M-LOAM estimators to solve the robot poses and extrinsics in a coarse-to-fine fashion. 
The most important step is to approximate the Gaussian noise covariance $\bm{\Sigma}$ to realistic measurement models.
Based on the discussion in Section \ref{sec:uncertainty}, 
we identify that three sources of error may make landmarks uncertain: sensor noise, degenerate pose estimation, and extrinsic perturbation.
The frame-to-frame motion estimation (Section \ref{sec:single_frame_estimation}) is approximately subjected to the sensor noise.
The tightly coupled odometry (Section \ref{sec:multilo_with_pure_odometry}) establishes a local map for the pose optimization. 
We should propagate pose uncertainties onto each map point.
Nevertheless, this operation is often time-consuming (around $10ms-20ms$) if more LiDARs and sliding windows are involved. 
To guarantee the real-time odometry, we do not compute the pose uncertainty here.
We simply set $\bm{\Sigma}=\mathbf{Z}$ as the covariance of residuals.
In mapping, we are given sufficient time for an accurate pose and a global map.
Therefore, we consider all uncertainty sources.
Section \ref{sec:mapping} explains how the pose uncertainty affects the mapping precision and $\bm{\Sigma}$ is propagated.

%% file: system_overview.tex
\section{System Overview}
\label{sec:overview}
We make three assumptions to simplify the system design:
\begin{itemize}
	\item LiDARs are synchronized, meaning that the temporal latency among different LiDARs is almost zero.
	\item The platform undergoes sufficient rotational and translational motion in the period of calibration initialization.
	\item The local map of the primary LiDAR should share an overlapping FOV with auxiliary LiDARs for feature matching in refinement to shorten the calibration phase. This can be achieved by moving the robot.
\end{itemize}

Fig. \ref{fig:system_pipeline} presents the pipeline of M-LOAM.
The system starts with measurement preprocessing (Section \ref{sec:preprocessing}), in which edge and planar features are extracted and tracked from denoised point clouds. 
The initialization module (Section \ref{sec:initialization}) provides all necessary values, including poses and extrinsics, for bootstrapping the subsequent nonlinear optimization-based M-LO.
The M-LO fuses multi-LiDAR measurements to optimize the odometry and extrinsics within a sliding window.
If the extrinsics already converge, 
we skip the extrinsic initialization as well as refinement steps 
and enter the pure odometry and mapping phase.
The probabilistic mapping module (Section \ref{sec:mapping}) constructs a global map with sufficient features to eliminate the odometry drift.
The odometry and mapping run concurrently in two separated threads.
 

%% file: measurement_preprocessing.tex
\section{Measurement Preprocessing}
\label{sec:preprocessing}
We implement three sequential steps to process LiDARs' raw measurements.
We first segment point clouds into many clusters to remove noisy objects, and then extract edge and planar features. 
To associate features between consecutive frames, we match a series of correspondences.
In this section, each LiDAR is handled independently.

\subsection{Segmentation for Noise Removal}
\label{sec:segmentation}
With knowing the vertical scanning angles of a LiDAR, 
we can project the raw point cloud onto a range image without data loss.
In the image, each valid point is represented by  a pixel. The pixel value records the Euclidean distance from a point to the origin. 
We apply the segmentation method proposed in \cite{bogoslavskyi2016fast} to group pixels into many clusters.
We assume that two neighboring points in the horizontal or vertical direction belong to the same object 
if their connected line is roughly perpendicular ($> 60deg$) to the laser beam.
We employ the breadth-first search algorithm to traverse all pixels, ensuring a constant time complexity.
We discard small clusters since they may offer unreliable constraints in optimization.

\subsection{Feature Extraction and Matching}
\label{sec:feature_extraction_matching}
We are interested in extracting the general edge and planar features.
We follow \cite{zhang2014loam} to select a set of feature points from measurements according to their curvatures.
The set of extracted features $\mathcal{F}$ consists of two subsets:
edge subset (high curvature) $\mathcal{E}$ and planar subset (low curvature) $\mathcal{H}$. 
Both $\mathcal{E}$ and $\mathcal{H}$ consist of a portion of features which are the most representative.
We further collect edge points from $\mathcal{E}$ with the highest curvature and planar points from $\mathcal{H}$ with the lowest curvature. 
These points form two new sets: $\hat{\mathcal{E}}$ and $\hat{\mathcal{H}}$.
The next step is to determine feature correspondences between two consecutive frames, $()^{l^{i}_{k-1}}\rightarrow()^{l^{i}_{k}}$, to construct geometric constraints.
For each point in $\hat{\mathcal{E}}^{l^{i}_{k}}$, two closest neighbors from $\mathcal{E}^{l^{i}_{k-1}}$ are selected as the edge correspondences. 
For a point in $\hat{\mathcal{H}}^{l^{i}_{k}}$, the three closest points to $\mathcal{H}^{l^{i}_{k-1}}$ that form a plane are selected as the planar correspondences.

%% file: initialization.tex
\begin{figure}
	\centering
	\includegraphics[width=0.4\textwidth]{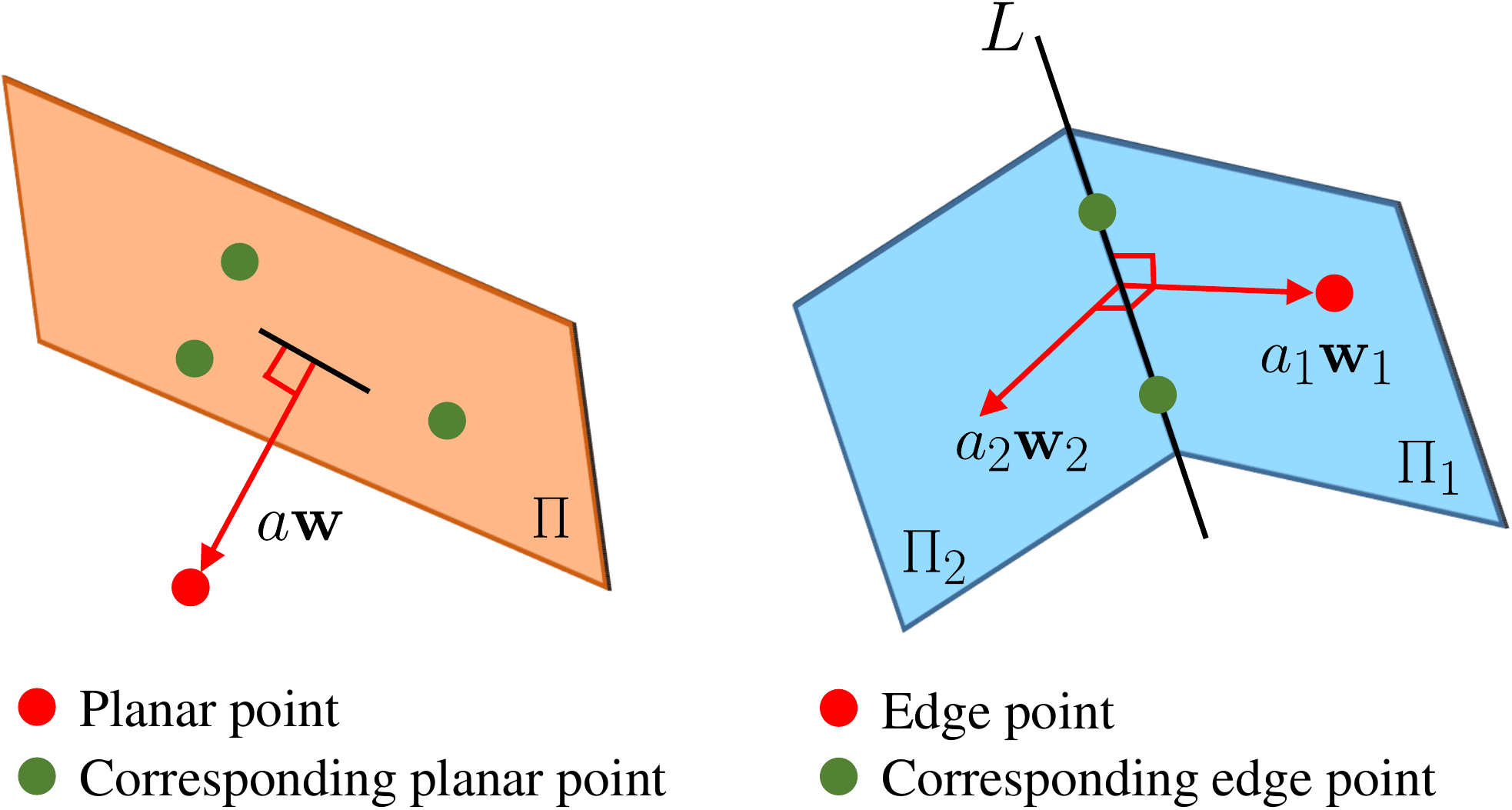}       
	\caption{The planar and edge residuals. The red dot indicates the reference point and the green dots are its corresponding points.}
	\label{fig:measurement_residual}     
\end{figure} 

\section{Initialization}
\label{sec:initialization}
Optimizing states of multiple LiDARs is highly nonlinear and needs initial guesses.
This section introduces our motion and extrinsic initialization approach, 
which does not require any prior mechanical configuration of the sensor suite.
It also does not involve any manual effort, 
making it particularly useful for autonomous robots.

\subsection{Scan-Based Motion Estimation}
\label{sec:single_frame_estimation}
With found correspondences between two 
successive frames of each LiDAR, we estimate the frame-to-frame transformation by minimizing residual errors of all features. 
As illustrated in Fig. \ref{fig:measurement_residual}, the residuals are formulated by both edge and planar correspondences. 
Let $\mathbf{x}_{k}$ be the relative transformation between two scans of a LiDAR at the $k^{th}$ frame.
Regarding planar features, 
for a point $\mathbf{p}\in\hat{\mathcal{H}}^{l_{k}^{i}}$, if $\Pi$ is the corresponding planar patch, the planar residual is computed as
\begin{equation}
\begin{aligned}
	\mathbf{r}_{\mathcal{H}}(\mathbf{x}_{k},\mathbf{p},\Pi)
	=
	a\mathbf{w},\ \ \ 
	a
	=
	\mathbf{w}^{\top}(\mathbf{R}_{k}\mathbf{p} + \mathbf{t}_{k})+d,
\label{equ:planar_residual}
\end{aligned}
\end{equation}
where $a$ is the point-to-plane Euclidean distance and $[\mathbf{w}, d]$ is the coefficient vector of $\Pi$. 
Then, for an edge point $\mathbf{p}\in\hat{\mathcal{E}}^{l^{i}_{k}}$, if $L$ is the corresponding edge line, we define the edge residual as a combination of two planar residuals using $\eqref{equ:planar_residual}$ as
\begin{equation}
	\mathbf{r}_{\mathcal{E}}(\mathbf{x}_{k},\mathbf{p},L)
	=
	\big[
	\mathbf{r}_{\mathcal{H}}(\mathbf{x}_{k},\mathbf{p},\Pi_{1}),\ 
	\mathbf{r}_{\mathcal{H}}(\mathbf{x}_{k},\mathbf{p},\Pi_{2})
	\big],
\label{equ:edge_residual}
\end{equation}
where $[\mathbf{w}_{1},d_1]$ and $[\mathbf{w}_{2},d_2]$ are the coefficients of $\Pi_{1}$ and $\Pi_{2}$,
$\mathbf{w}_{1}$ coincides with the projection direction from $L$ to $\mathbf{p}$,
and $\Pi_{2}$ is perpendicular to $\Pi_{1}$ s.t. $\mathbf{w}_{2}\bot\mathbf{w}_1$, and $\mathbf{w}_{2}\bot L$.
The above definitions are different from that of LOAM \cite{zhang2014loam}, and show two benefits.
Firstly, the edge residuals offer additional constraints to the states.
Furthermore, the residuals are represented as vectors, not scalars, allowing us to multiply a $3\times3$ covariance matrix. 
We minimize the sum of all residual errors to obtain the MLE as
\begin{equation}
\begin{aligned}
	&\hat{\mathbf{x}}_{k}
	=
	\underset{\mathbf{x}_{k}}{\arg \min}
	\sum_{\mathbf{p}\in \hat{\mathcal{F}}^{l^{i}_{k}} }^{}
	\rho\Big(
	\big\|
	\mathbf{r}_{\mathcal{F}}(\mathbf{x}_{k},\mathbf{p})
	\big\|^{2}_{\bm{\Sigma}_{\mathbf{p}}}
	\Big),\\
	&\mathbf{r}_{\mathcal{F}}(\mathbf{x}_{k},\mathbf{p})
	=
	\begin{cases}
		\mathbf{r}_{\mathcal{E}}(\mathbf{x}_{k},\mathbf{p},L)
		& \text{if } \mathbf{p}\in\hat{\mathcal{E}}^{l^{i}_{k}}\\
		\mathbf{r}_{\mathcal{H}}(\mathbf{x}_{k},\mathbf{p},\Pi)
		& \text{if } \mathbf{p}\in\hat{\mathcal{H}}^{l^{i}_{k}}
	\end{cases},
	\label{equ:objective_initialization}
\end{aligned}
\end{equation}
where the Jacobians of $\mathbf{r}_\mathcal{F}(\cdot)$ are detailed in Appendix \ref{app.jacobian_initialization}.

In practice, points are skewed after a movement on LiDARs with a rolling-shutter scan.
After solving the incremental motion $\mathbf{x}_{k}$, we correct points' positions by transforming them into the last frame $()^{l^{i}_{k-1}}$.
Let $t_{k-1}$ and $t_{k}$ be the start and end time of a LiDAR scan respectively. 
For a point $\mathbf{p}$ captured at $t\in(t_{k-1}, t_{k}]$, 
it is transformed as\footnote{The timestampe of each point can be obtained from raw data.}
\begin{equation}
	\mathbf{p}^{l^{i}_{k-1}}
	=
	\mathbf{R}_{k}^{\tau}\mathbf{p}
	+
	\mathbf{t}_{k}^{\tau},\ \ \ 
	\tau = \frac{t-t_{k-1}}{t_{k}-t_{k-1}}, 
\end{equation}
where the rotation and translation are linearly interpolated \cite{ye2019tightly}:
\begin{equation}
	\mathbf{R}^{\tau}_{k}
	=
	\exp(\bm{\phi}^{\wedge}_{k})^{\tau}
	=
	\exp(\tau\bm{\phi}^{\wedge}_{k}),
	\ \ \ 
	\mathbf{t}^{\tau}_{k}
	=
	\tau\mathbf{t}_{k}.
\end{equation}

\subsection{Calibration of Multi-LiDAR System}
\label{sec:initialization_calibration}
The initial extrinsics are obtained by aligning motion sequences of two sensors. 
This is known as solving the hand-eye calibration problem $\mathbf{A}\mathbf{X}=\mathbf{X}\mathbf{B}$, 
where $\mathbf{A}$ and $\mathbf{B}$ are the historical transformations of two sensors and $\mathbf{X}$ is their extrinsics. 
As the robot moves, the following equations of the $i^{th}$ LiDAR should hold for any $k$
\begin{align}
	\mathbf{R}^{l^{i}_{k-1}}_{l^{i}_{k}}
	\mathbf{R}^{b}_{l^{i}}
	&=
	\mathbf{R}^{b}_{l^{i}}    
	\mathbf{R}^{b_{k-1}}_{b_{k}},
	\label{equ:hand_eye_calibration_1}\\
	(\mathbf{R}^{l^{i}_{k-1}}_{l^{i}_{k}} - \mathbf{I}_{3})
	\mathbf{t}^{b}_{l^{i}}
	&=
	\mathbf{R}^{b}_{l^{i}}\mathbf{t}^{b_{k-1}}_{b_{k}}
	- \mathbf{t}^{l^{i}_{k-1}}_{l^{i}_{k}},
\label{equ:hand_eye_calibration_2}
\end{align}    
where the original problem is decomposed into the rotational and translational components according to \cite{yang2015monocular}.
We implement this method to initialize the extrinsics online.

\begin{figure*}
	\centering
	\includegraphics[width=0.93\textwidth]{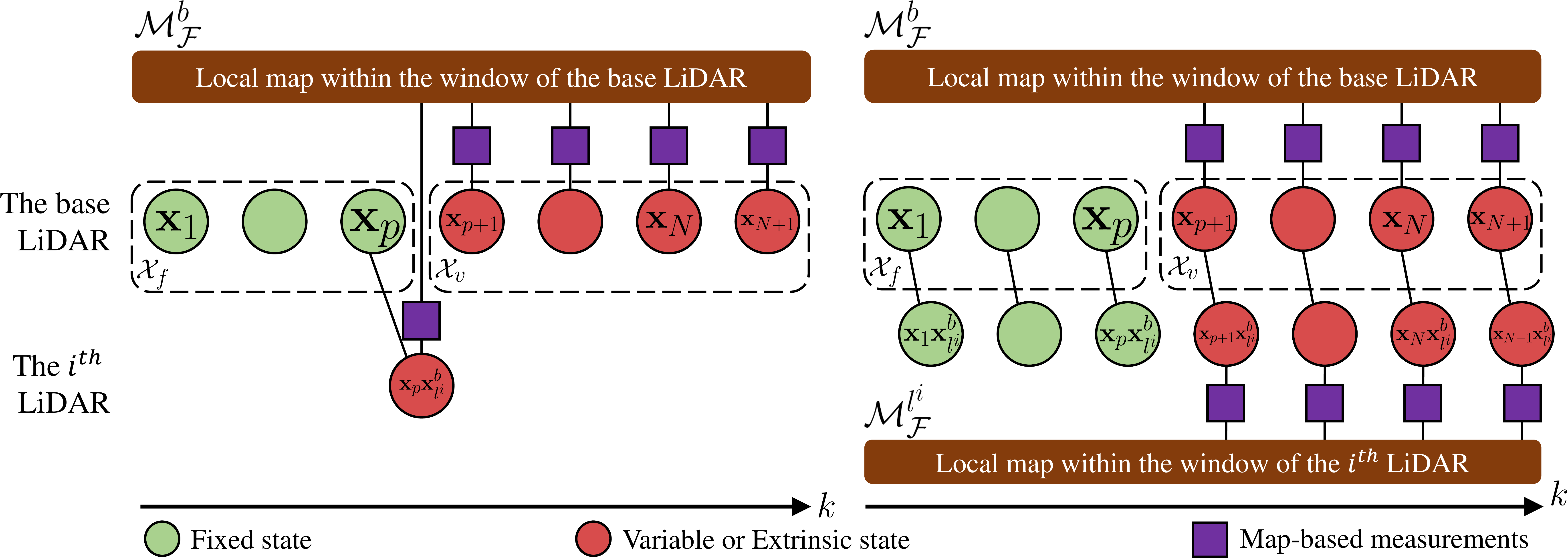}
	\caption{Illustration of a graphical model for a sliding window estimator ($p=3, N=6$) with online calibration (left) and pure odometry (right). 
			$\mathcal{M}^{b}_{\mathcal{F}}$ ($\mathcal{M}^{l^{i}}_{\mathcal{F}}$) is the local map of the base ($i^{th}$) LiDAR. It consists of transformed and merged feature points captured by the base ($i^{th}$) LiDAR from the first $N$ frames in the window. 
			Note that the extrinsics are optimized in the online calibration, while they are fixed in the pure odometry.}       
	\label{fig:graph_optimization}     
\end{figure*}

\subsubsection{Rotation Initialization}
We rewrite $\eqref{equ:hand_eye_calibration_1}$ as a linear equation by employing the quaternion:
\begin{equation}
\begin{aligned}
	&\mathbf{q}^{l^{i}_{k-1}}_{l^{i}_{k}}\otimes
	\mathbf{q}^{b}_{l^{i}}
	=
	\mathbf{q}^{b}_{l^{i}}\otimes    
	\mathbf{q}^{b_{k-1}}_{b_{k}} \\
	\Rightarrow \ 
	&\left[\mathbf{Q}_{1}(\mathbf{q}_{l^{i}_{k-1}}^{l^{i}_{k}}) 
	-
	\mathbf{Q}_{2}(\mathbf{q}_{b_{k}}^{b_{k-1}}) \right] 
	\mathbf{q}^{b}_{l^{i}}
	=
	\mathbf{Q}^{k-1}_{k}\mathbf{q}^{b}_{l^{i}},
\label{equ:rotation_residual}
\end{aligned}    
\end{equation}
where $\otimes$ is the quaternion multiplication operator and $\mathbf{Q}_{1}(\cdot)$ and $\mathbf{Q}_{2}(\cdot)$ 
are the matrix representations for left and right quaternion multiplication \cite{sola2017quaternion}. 
Stacking $\eqref{equ:rotation_residual}$ from multiple time intervals, we form an overdetermined linear system as
\begin{equation}
\begin{aligned}
	\begin{bmatrix}
		w^{0}_{1}\cdot\mathbf{Q}_{1}^{0} \\
		\vdots \\
		w^{K-1}_{K}\cdot\mathbf{Q}_{K}^{K-1} 
	\end{bmatrix}_{4K\times 4}
	\mathbf{q}_{l^{i}}^{b} = \mathbf{Q}_{K} \mathbf{q}_{l^{i}}^{b} = \mathbf{0}_{4K\times 4},
\label{equ:rotation_linear_system}
\end{aligned}  
\end{equation}
where $K$ is the number of constraints, and $w^{k-1}_{k}$ are robust weights defined as the angle in the angle-axis representation of the residual quaternion:
\begin{equation}
\begin{split}
	w^{k-1}_{k}
	&=
	\rho'(\phi), \ \ \ 
	\phi
	=
	2\arctan
	\big(||\mathbf{q}_{xyz}||,q_{w}\big)
	,\\
	\mathbf{q}
	&=
	(\check{\mathbf{q}}^{b^{}}_{l^{i}})^{*} \otimes
	(\mathbf{q}^{l^{i}_{k-1}}_{l^{i}_{k}})^{*} \otimes
	\check{\mathbf{q}}^{b}_{l^{i}} \otimes
	\mathbf{q}^{b_{k-1}}_{b_{k}},
\end{split}
\label{equ:weight_linear_system}
\end{equation}
where $\rho'(\cdot)$ is the derivative of the Huber loss,
$\check{\mathbf{q}}^{b}_{l^{i}}$ is the currently estimated extrinsic rotation,
and $\mathbf{q}^{*}$ is the inverse of $\mathbf{q}$.
Subject to $\|\mathbf{q}^{b}_{l^{i}}\|=1$, we find the closed-form solution of \eqref{equ:rotation_linear_system} using SVD. 
For the full observability of 3-DoF rotation, sufficient motion excitation are required.
Under sufficient constraints, the null space of $\mathbf{Q}_{K}$ should be rank one.
This means that we only have one zero singular value.
Otherwise, the null space of $\mathbf{Q}_{K}$ may be larger than one due to degenerate motions on one or more axes.
Therefore, we need to ensure that the second small singular value $\sigma_{\text{min}2}$ is large enough by checking whether this condition is achieved.
We set a threshold $\sigma_{\mathbf{R}}$, and terminate the rotation calibration if $\sigma_{\text{min}2} > \sigma_{\mathbf{R}}$.
The increasing data grows the row of $\mathbf{Q}_{K}$ rapidly.
To bound the computational time, we use a priority queue \cite{cormen2009introduction} with the length $K=300$ to incrementally store historical constraints. 
Constraints with small rotation are removed.

\subsubsection{Translation Initialization}
Once the rotational calibration is finished, we construct a linear system from \eqref{equ:hand_eye_calibration_2} by incorporating all the available data as
\begin{equation}
\label{equ:trans_initialization}
	\begin{bmatrix}
	\mathbf{R}^{l^{i}_{0}}_{l^{i}_{1}} - \mathbf{I}_3\\
	\vdots\\
	\mathbf{R}^{l^{i}_{K-1}}_{l^{i}_{K}} - \mathbf{I}_3
	\end{bmatrix}_{3K\times 3}
	\mathbf{t}^{b}_{l^{i}}
	=
	\begin{bmatrix}
	\hat{\mathbf{R}}^{b}_{l^{i}}\mathbf{t}^{b_{0}}_{b_{1}}
	- \mathbf{t}^{l^{i}_{0}}_{l^{i}_{1}}\\
	\vdots\\
	\hat{\mathbf{R}}^{b}_{l^{i}}\mathbf{t}^{b_{K-1}}_{b_{K}}
	- \mathbf{t}^{l^{i}_{K-1}}_{l^{i}_{K}}
	\end{bmatrix}_{3K\times 1},
\end{equation}
where $\mathbf{t}^{b}_{l^{i}}$ is obtained by applying the least-squares approach.
However, the translation at the $z-$ axis is unobservable if the robot motion is planar.
In this case, we set $t_z=0$ and rewrite \eqref{equ:trans_initialization} by removing the $z-$ component of $\mathbf{t}^{b}_{l^{i}}$. 
Unlike \cite{jiao2019automatic}, our method cannot initialize $t_{z}$ by leveraging ground planes
and must recover it in the refinement phase (Section \ref{sec:multilo_with_online_calibration}).

%% file: multilo.tex
\section{Tightly Coupled Multi-LiDAR Odometry With Calibration Refinement}
\label{sec:multilo}
Taking the initial guesses as input, we propose a tightly coupled M-LO to optimize all states within a sliding window. 
This procedure is inspired by the recent success of bundle adjustment, graph-based formation, and marginalization in multi-sensor systems \cite{leutenegger2013keyframe,qin2018vins,ye2019tightly}. 

\subsection{Formulation}
\label{sec:multilo_formulation}
The full state vector in the sliding window is defined as 
\begin{equation}
\begin{aligned}
	\mathcal{X}
	&=
	[\mathcal{X}_{f},\ \ \ \ \ \ \ \ \ \mathcal{X}_{v},\ \ \ \ \ \ \ \ \ \ \ \ \ \ \ \ \mathcal{X}_{e}]\\
	&=
	[\mathbf{x}_{1}, \cdots, \mathbf{x}_{p}, \mathbf{x}_{p+1}, \cdots, \mathbf{x}_{N+1},
	\mathbf{x}_{l^{2}}^{b},\cdots,\mathbf{x}_{l^{I}}^{b}],\\
	\mathbf{x}_{k}
	&=
	[\mathbf{t}^{w}_{b_{k}}, \mathbf{q}^{w}_{b_{k}}],
	\ \ k\in[1,N+1],
	\\
	\mathbf{x}_{l^{i}}^{b}
	&=
	[\mathbf{t}_{l^{i}}^{b}, \mathbf{q}_{l^{i}}^{b}],
	\ \ i\in[1, I],    
\end{aligned}    
\end{equation}
where $\mathbf{x}_{k}$ is the state of the primary sensor in the world frame at different timestamps, 
$\mathbf{x}_{l^{i}}^{b}$ represents the extrinsics from the primary LiDAR to auxiliary LiDARs, and $N+1$ is the number of states in the window. 
To establish data association to constrain these states, we build local maps.

Fig. \ref{fig:graph_optimization} visualizes the graph-based formualtion.
We use $p$ to index the pivot state of the window and set $\mathbf{x}_{p}$ as the origin of the local map.
With the relative transformations from the pivot frame to other frames, the map is constructed by concatenating with features at the first $N$ frames i.e. $\mathcal{F}^{l^{i}_{k}}, k\in[1, N]$.
The local feature map of the $i^{th}$ LiDAR, which consists of the local edge map and local planar map, is denoted by $\mathcal{M}^{l^{i}}$. 
We split $\mathcal{X}$ into three groups: $\mathcal{X}_{f}$, $\mathcal{X}_{v}$, and $\mathcal{X}_{e}$.
$\mathcal{X}_{f}=[\mathbf{x}_{1},\cdots,\mathbf{x}_{p}]$ are considered as the fixed, accurate states. 
$\mathcal{X}_{v}=[\mathbf{x}_{p+1},\cdots,\mathbf{x}_{N+1}]$ are considered as the variables which are updated recursively during optimization. 
$\mathcal{X}_{e}=[\mathbf{x}_{l^{2}}^{b},\cdots,\mathbf{x}_{l^{I}}^{b}]$ are the extrinsics. 
Their setting depends on the convergence of the online calibration. 
We minimize the sum of all residual errors within the sliding window to obtain a MAP estimation as
\begin{equation}
\begin{aligned}
	\hat{\mathcal{X}}
	=
	\underset{{\mathcal{X}}}{\arg \min}\ 
	\bigg\{
	\big\|\mathbf{r}_{pri}(\mathcal{X})\big\|^{2}
	+
	f_{\mathcal{M}}(\mathcal{X})
	\bigg\},
\end{aligned}
\label{equ:objective_function_multilo_whole}
\end{equation}
where $\mathbf{r}_{pri}(\mathcal{X})$ is the prior term from the state marginalization defined in Section \ref{sec:marginalization} 
and $f_{\mathcal{M}}(\mathcal{X})$ is the sum of map-based residual errors. 
Its Jacobians \eqref{equ:objective_function_multilo_whole} are given in Appendix \ref{app.jacobian_initialization}.
Different from the frame-to-frame estimation in Section \ref{sec:single_frame_estimation}, 
the presented sliding-window estimator jointly optimizes all states in the window. 
This approach outputs more accurate results since the local map provides dense and reliable correspondences. 
If sensors are precisely calibrated, the constraints from other LiDARs are also used.
According to the convergence of calibration, 
we divide the problem into two subtasks: \textit{online calibration} (variable $\mathcal{X}_{e}$) and \textit{pure odometry} (fixed $\mathcal{X}_{e}$). At each task, the definition of $f_{\mathcal{M}}(\mathcal{X})$ is different, 
and we present the details in Section \ref{sec:multilo_with_online_calibration} and \ref{sec:multilo_with_pure_odometry}.

\subsection{Optimization With Online Calibration}
\label{sec:multilo_with_online_calibration}
We exploit the map-based measurements to refine the coarse initialization results. 
Here, we treat the calibration as a registration problem.
$f_{\mathcal{M}}(\mathcal{X})$ is divided into two functions w.r.t. $\mathcal{X}_{v}$ and $\mathcal{X}_{e}$.
For states in $\mathcal{X}_{v}$, the constraints are constructed from correspondences between features of the primary sensor at the latest frames i.e. $\mathcal{F}^{{b}_{k}},k\in[p+1,N+1]$ and those of the primary local map i.e. $\mathcal{M}^{b}$. 
For states in $\mathcal{X}_{e}$, the constraints are built up from correspondences between features of auxiliary LiDARs at the $p^{th}$ frame i.e. $\mathcal{F}^{l^{i}_{p}}$ and the map $\mathcal{M}^{b}$.

The correspondences between $\mathcal{F}^{b_{k}}$ and $\mathcal{M}^{b}$ are found using the method in \cite{zhang2014loam}.
KD-Tree is used for fast indexing in a map. 
\textit{1)} For each edge point, we find a set of its nearest points in the local edge map within a specific region. 
This set is denoted by $\mathcal{S}$, and its covariance is then computed. The eigenvector associated with the largest value implies the direction of the corresponding edge line. By calculating the mean of $\mathcal{S}$, the position of this line is also determined. 
\textit{2)} For each planar point, the coefficients of the corresponding plane in the local planar map are obtained by solving a linear system such as $\mathbf{w}\mathbf{s}+d=0,\forall\mathbf{s}\in\mathcal{S}$.
Similarly, we find correspondences between  $\mathcal{F}^{l^{i}_{p}}$ and $\mathcal{M}^{b}$.
Finally, we define the objective as the sum of all measurement residuals for the online calibration as
\begin{equation}
\begin{aligned}
	f_\mathcal{M}(\mathcal{X})
	&=
	f_\mathcal{M}(\mathcal{X}_{v}) + f_\mathcal{M}(\mathcal{X}_{e})\\
	&=
	\sum_{k=p+1}^{N+1}
	\sum_{\mathbf{p}\in\mathcal{F}^{b_{k}}}^{}
	\rho\Big(
	\big\|
	\mathbf{r}_{\mathcal{F}}(\mathbf{x}_{p}^{-1}\mathbf{x}_{k},\mathbf{p})
	\big\|^{2}_{\bm{\Sigma}_{\mathbf{p}}}
	\Big)\\
	&\ + \ \ 
	\sum_{i=2}^{I}
	\sum_{\mathbf{p}\in\mathcal{F}^{l^{i}_{p}}}^{}
	\rho\Big(    
	\big\|
	\mathbf{r}_{\mathcal{F}}(\mathbf{x}^{b}_{l^{i}},\mathbf{p})
	\big\|^{2}_{\bm{\Sigma}_{\mathbf{p}}}    
	\Big),
\end{aligned}
\label{equ:objective_online_calibration}
\end{equation}
where $\mathbf{x}_{p}^{-1}\mathbf{x}_{k}$ defines the relative transformation from the pivot frame to the $k^{th}$ frame. 

\subsection{Optimization With Pure Odometry}
\label{sec:multilo_with_pure_odometry}
Once we finish the online calibration by fulfilling the convergence criterion (Section \ref{sec:calibration_termination}), 
the optimization with pure odometry given accurate extrinsics is then performed.
In this case, we do not optimize the extrinsics, and utilize all available map-based measurement to improve the single-LiDAR odometry. 
We incorporate constraints between features of all LiDARs and local maps into the cost function as
\begin{equation}
\begin{aligned}
	f_\mathcal{M}(\mathcal{X})
	&=
	f_\mathcal{M}(\mathcal{X}_{v})\\
	&=
	\sum_{k=p+1}^{N+1}
	\sum_{\mathbf{p}\in\mathcal{F}^{b_{k}}}^{}
	\rho\Big(
	\big\|
	\mathbf{r}_{\mathcal{F}}(\mathbf{x}_{p}^{-1}\mathbf{x}_{k},\mathbf{p})
	\big\|^{2}_{\bm{\Sigma}_{\mathbf{p}}}
	\Big)\\
	&+
	\sum_{i=2}^{I}
	\sum_{k=p+1}^{N+1}
	\sum_{\mathbf{p}\in\mathcal{F}^{l^{i}_{k}}}^{}
	\rho\Big(    
	\big\|
	\mathbf{r}_{\mathcal{F}}(\mathbf{x}_{p}^{-1}\mathbf{x}_{k}\mathbf{x}^{b}_{l^{i}},\mathbf{p})
	\big\|^{2}_{\bm{\Sigma}_{\mathbf{p}}}    
	\Big),
\end{aligned}
\label{equ:objective_pure_odometry}
\end{equation}
where $\mathbf{x}_{p}^{-1}\mathbf{x}_{k}\mathbf{x}^{b}_{l^{i}}$ is the transformation from the primary LiDAR at the pivot frame to auxiliary LiDARs at the $k^{th}$ frame.

\subsection{Monitoring the Convergence of Calibration}
\label{sec:calibration_termination}
While working on the online calibration in an unsupervised way, it is of interest to decide whether calibration converges.
After the convergence, we fix the extrinsics.
This is beneficial to our system since both the odometry and mapping are given more geometric constraints from auxiliary LiDARs for more accurate poses.
As derived in \cite{zhang2016degeneracy}, the \textit{degeneracy factor} $\lambda$, which is the smallest eigenvalue of the \textit{information matrix}, 
reveals the condition of an optimization-based state estimation problem. 
Motivated by this work, we use $\lambda$ to indicate whether our problem contains sufficient constraints or not for accurate extrinsics. 
The detailed pipeline to update extrinsics and monitor the convergence is summarized in Algorithm \ref{alg.calibration_termination}.
The algorithm takes the function $f_{\mathcal{M}}(\cdot)$ defined in \eqref{equ:objective_online_calibration} as well as the current extrinsics as input, 
and returns the optimized extrinsics. 
On line $4$, we compute $\lambda$ from the \textit{information matrix} of the cost function.
On lines $5$--$7$, the extrinsics are updated if $\lambda$ is larger than a threshold. 
On line $8$, we use the number of candidate extrinsics to check the convergence.
On lines $9$--$10$, the convergence criterion is met, and the termination is thus triggered. 
We then compute the sampling mean of $\mathcal{L}$ as the resulting extrinsics and the sampling covariance as the calibration covariance.

\begin{algorithm}[t]
	\caption{Monitoring Calibration Convergence}
	\label{alg.calibration_termination}        
	\LinesNumbered
	\KwIn{objective $f_{\mathcal{M}}(\cdot)$, current extrinsics
		$\check{\mathbf{x}}\triangleq\mathbf{x}^{b}_{l^{i}}$}
	\KwOut{optimal extrinsics $\hat{\mathbf{x}}$, covariance matrix $\bm{\Xi}_{calib}$}
	Denote $\mathcal{L}$ the set of all eligible estimates;\\
	\If{calibration is ongoing}
	{
		Linearize $f_{\mathcal{M}}$ at $\check{\mathbf{x}}$ to obtain $\bm{\Lambda}=(\frac{\partial f_{\mathcal{M}}}{\partial \mathbf{x}})^{\top}
		\frac{\partial f_{\mathcal{M}}}{\partial \mathbf{x}}$;\\		
		Compute the smallest eigenvalue $\lambda$ of $\bm{\Lambda}$;\\
		\If{$\lambda > \lambda_{calib}$} 
		{
			Set $\check{\mathbf{x}}$ as the current extrinsics of the system;\\
			$\mathcal{L} \leftarrow \mathcal{L}\cup\check{\mathbf{x}}$;\\
			\If{$|\mathcal{L}|>\mathcal{L}_{calib}$}{
				$\hat{\mathbf{x}} \leftarrow E[\mathbf{x}]$ as the mean;\\ 
				$\bm{\Xi}_{calib} \leftarrow Cov[\mathbf{x}]$ as the covariance;\\ 				
				Stop the online calibration;\\
			}
		}
	}
	\Return $\hat{\mathbf{x}}$, $\bm{\Xi}_{calib}$\\
\end{algorithm}

\subsection{Marginalization}
\label{sec:marginalization}
We apply the marginalization technique to remove the oldest variable states in the sliding window.
The marginalization is a process to incorporates historical constraints as a prior into the objective, 
which is an essential step to maintain the consistency of odometry and calibration results.
In our system, $\mathbf{x}_{p}$ is the only state to be marginalized after each optimization.
By applying the Schur complement, we obtain the linear information matrix $\bm{\Lambda}_{rr}^{*}$  and residual 
$\mathbf{g}_{r}^{*}$ w.r.t. the remaining states.
The prior residuals are formulated as $\|\mathbf{r}_{pri}\|^{2}=\mathbf{g}_{r}^{*\top}\bm{\Lambda}_{rr}^{*-1}\mathbf{g}_{r}^{*}$.
Appendix \ref{app.marginalization} provides some mathematical foundations.

%% file: mapping.tex
\begin{figure}
	\centering
	\includegraphics[width=0.48\textwidth]{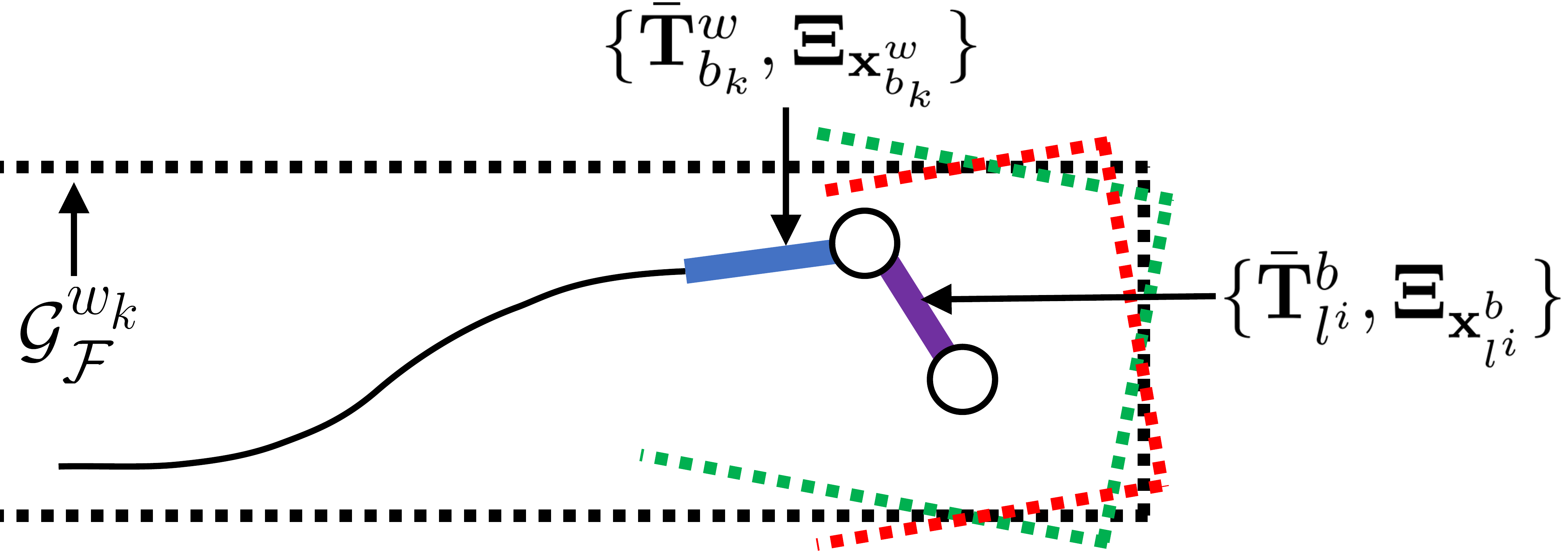}
	\caption{Illustration of the mapping process and occurrence of noisy map points. 
			The black curve represents historical poses. The blue curve indicates the current pose of the primary LiDAR. 
			The purple curve shows the extrinsics from the primary LiDAR to the auxiliary LiDAR.
			With the pose and extrinsics, input features (red and green dots) are transformed and added into the global map (black dots). 
			The noisy poses make new map points uncertain.}
	\label{fig:mapping_error}
\end{figure}

\section{Uncertainty-Aware Multi-LiDAR Mapping}
\label{sec:mapping}
We first review the pipeline of the mapping module of typical LiDAR SLAM systems \cite{zhang2014loam,shan2018lego,lin2020loam}.
Taking the prior odometry as input, the algorithm constructs a global map and refines the poses with enough constraints.
This is done by minimizing the sum of all map-based residual errors as
\begin{equation}
\label{equ:mapping_objective}
\begin{aligned}
	\hat{\mathbf{x}}_{b_{k}}^{w}
	=
	\underset{\mathbf{x}_{b_{k}}^{w}}{\arg \min}\ 
	\sum_{i=1}^{I}
	\sum_{\mathbf{p}\in\mathcal{F}^{l^{i}_{k}}}
	\rho\Big(
	\big\|
	\mathbf{r}_{\mathcal{F}}(\mathbf{x}_{b_{k}}^{w}\mathbf{x}^{b}_{l^{i}},\mathbf{p})
	\big\|^{2}_{\bm{\Sigma}_{\mathbf{p}}}
	\Big),    
\end{aligned}
\end{equation}
where $\mathcal{F}^{l{^{i}_{k}}}$ are the input features while perceiving the $k^{th}$ point cloud, 
$\mathcal{G}^{{w}_{k}}_{\mathcal{F}}$ is the global map, 
and $\mathbf{x}_{b_{k}}^{w}\mathbf{x}^{b}_{l^{i}}$ denotes the state of the $i^{th}$ LiDAR.
We use the method in Section \ref{sec:multilo_with_online_calibration} to find correspondences 
between $\mathcal{F}^{l{^{i}_{k}}}$ and $\mathcal{G}^{{w}_{k}}_{\mathcal{F}}$.
After solving \eqref{equ:mapping_objective}, the resulting pose is used to transform current features into the world frame, and add them into the map.
To reduce the computational and memory complexity, the map is also
downsized using a voxel grid filter \cite{rusu20113d}.
However, the precision of optimization depends on the map quality.
Fig. \ref{fig:mapping_error} visualizes the occurrence of noisy map points (landmarks) transformed by the uncertain pose.
We believe that three sources of uncertainties make map points noisy: sensor noise, degenerate pose estimation, and extrinsic perturbation.

In the next section, we propagate the uncertainties of LiDAR points and poses (represented as transformation matrices) on map points.
As a result, each map point is modeled as an i.i.d Gaussian variable.
We then propose an uncertainty-aware approach to improve the robustness and accuracy of the multi-LiDAR mapping algorithm.

\subsection{Uncertainty Propagation}
\label{sec:uncertainty_propagation}
Continuing the preliminaries in Section \ref{sec:uncertainty}, we now compute $\bm{\Xi}$.
The mapping poses are optimized by solving the NLS problem \eqref{equ:mapping_objective}.
We directly calculate the inverse of the \textit{information matrix}, i.e. $\bm{\Xi}_{\mathbf{x}^{w}_{b_{k}}}=\bm{\Lambda}^{-1}$, as the covariance.
The setting of the extrinsic covariances depends on a specific situation. 
We generally define the extrinsic covariance as
\begin{equation}
\begin{aligned}
	\bm{\Xi}_{\mathbf{x}^{b}_{l^{i}}}
	=
	\alpha
	\cdot
	\bm{\Xi}_{calib}, 
	\ \ \ 
	\bm{\xi}^{b}_{l^{i}}
	\sim
	\mathcal{N}(\mathbf{0},\bm{\Xi}_{\mathbf{x}^{b}_{l^{i}}}),	
\end{aligned}
\label{equ:ext_covariance}
\end{equation}
where $\bm{\xi}^{b}_{l^{i}}$ is the perturbation variable of extrinsics, 
$\bm{\Xi}_{calib}$ is the calibration covariance calculated according to Algorithm \ref{alg.calibration_termination}, 
and $\alpha$ is a scaling parameter allowing us to increase the magnitude of the covariance. 
If a multi-LiDAR system has been recently calibrated, we set $\alpha=1$, 
whereas if the system has been used for a long time and not re-calibrated,
there should be small extrinsic deviations on LiDARs, and $\alpha$ is set to be larger.
It has an implicit relation with time and external impact, and temperature drift.
Given the mean pose of the primary sensor, the mean extrinsics, and their covariances, we then compute the mean poses of other LiDARs and the covariances i.e. $\{\mathbf{T}^{w}_{l^{i}_{k}}, \bm{\Xi}^{w}_{l^{i}_{k}}\}$. 
This is a problem about compounding two noisy poses. 
We follow the fourth-order approximation in \cite{barfoot2014associating} to calculate them.
And then, we need to pass the Gaussian uncertainty of a point through a noisy transformation to 
produce a new landmark $\mathbf{y}\in\mathcal{G}^{w_{k+1}}_{\mathcal{F}}$ 
with the mean and covariance as $\{\bar{\mathbf{y}}, \bm{\Sigma}\}$. 
By transforming a point into the world frame, we have
\begin{equation}
\begin{aligned}
	\mathbf{y}
	\triangleq
	\mathbf{T}^{w}_{l^{i}_{k}}
	\mathbf{p}_{h}
	&=
	\exp(\bm{\xi}^{w^\wedge}_{el^{i}_{k}})
	\bar{\mathbf{T}}^{w}_{l^{i}_{k}}
	(\bar{\mathbf{p}}_{h} + \mathbf{D}\bm{\zeta})
	\\
	&\approx
	\big(\mathbf{I} + \bm{\xi}^{w^\wedge}_{l^{i}_{k}}\big)
	\bar{\mathbf{T}}^{w}_{l^{i}_{k}}
	(\bar{\mathbf{p}}_{h} + \mathbf{D}\bm{\zeta}),    
\end{aligned}
\end{equation}
where we keep the first-order approximation of the exponential map.
If we multiply out the equation and retain only the first-order terms, we have
\begin{equation}
\begin{aligned}
	\mathbf{y}
	&\approx
	\mathbf{h}
	+
	\mathbf{H}\bm{\theta},
\end{aligned}
\end{equation}
where
\begin{equation}
\begin{aligned}
	\mathbf{h}
	&=
	\bar{\mathbf{T}}^{w}_{l^{i}_{k}}\bar{\mathbf{p}}_{h},
	\ \ \ \ \ \ \ \ \ 
	\mathbf{H}
	=
	\big[(\bar{\mathbf{T}}^{w}_{l^{i}_{k}}\bar{\mathbf{p}}_{h})^{\odot}
	\ \ \ \bar{\mathbf{T}}^{w}_{l^{i}_{k}}\mathbf{D}\big],\\
	\bm{\theta}
	&=
	[\bm{\xi}^{w\top}_{l^{i}_{k}},\bm{\zeta}^{\top}]^{\top},
	\ \ \bm{\theta}\sim\mathcal{N}(\mathbf{0}, \bm{\Theta}),
	\ \ \bm{\Theta}=\text{diag}(\bm{\Xi}^{w}_{l^{i}_{k}},\mathbf{Z}),\\
\end{aligned}
\end{equation}
and the operator $\odot$ converts a $4\times1$ column into a $4\times6$ matrix:
\begin{equation}
	\begin{bmatrix}
	\bm{\varepsilon} \\
	\eta\\
	\end{bmatrix}^{\odot}
	=
	\begin{bmatrix}
	\eta\mathbf{I} & -\bm{\varepsilon}^{\wedge}\\
	\mathbf{0}^{\top} & \mathbf{0}^{\top}
	\end{bmatrix},
	\ \ \ 
	\bm{\varepsilon}\in\mathbb{R}^{3},
	\ \ \ 
	\eta=1.
\end{equation}

All perturbation variables are embodied in $\bm{\theta}$ in $\mathbb{R}^{9}$. 
The covariance $\bm{\Theta}=\text{diag}(\bm{\Xi}^{w}_{l^{i}_{k}},\mathbf{Z})$ denotes the combined uncertainties of sensor readings, estimated poses, and extrinsics.
Linearly transformed by $\mathbf{H}$, $\mathbf{y}$ is Gaussian with the mean $\bar{\mathbf{y}}$ and covariance $\bm{\Sigma}$ as
\begin{equation}
\begin{aligned}
	\bar{\mathbf{y}}
	=\mathbf{h},
	\ \ \ 
	\bm{\Sigma}
	=
	\mathbf{H}\bm{\Theta}\mathbf{H}^{\top}.
\end{aligned}
\label{equ:point_covariance}
\end{equation}

We follow \cite{kim2017uncertainty} to use the trace, i.e., $\text{tr}(\bm{\Sigma})$, to quantify the magnitude of a covariance.

\subsection{Uncertainty-Aware Operation}
\label{sec:application_uncertainty_propagation}
The original mapping algorithm is integrated with three additional steps to boost its performance and robustness.
In the last section, we show that the covariances of all map points are propagated by considering three sources of error.
First, problem \eqref{equ:mapping_objective} is integrated with the propagated covariance in \eqref{equ:point_covariance}.
This operation lets the cost function take the pose uncertainty and extrinsic perturbation into account. 
As a result, a point that stays near the origin should have a high weight. 
Moreover, the primary LiDAR tends to have more confidence than auxiliary LiDARs given the extrinsic covariances.

At the last iteration of the optimization, 
$\bm{\Xi}_{\mathbf{x}^{w}_{b_{k}}}$ is equal to the inverse of the new \textit{information matrix}.
Second, the uncertainty of each point after transformation is re-propagated.
We filter out outliers if $\text{tr}(\bm{\Sigma})$ is larger than a threshold.
Finally, we modify the original voxel grid filter to downsize the global map in a probabilistic way.
The modified filter samples points for each cube according to their covariances. Let $\{\mathbf{y}_{i}, \bm{\Sigma}_{i}\}$ be the $i^{th}$ point in a cube, and $M$ be the number of points in the cube. The sampled mean and covariance of a cube are
\begin{equation}
\begin{aligned}
	\bar{\mathbf{y}}
	=
	\sum_{i=1}^{M}w_{i}\mathbf{y}_{i},\ \ \ 
	\bm{\Sigma}
	=
	\sum_{i=1}^{M}w_{i}^{2}\bm{\Sigma}_{i},
\end{aligned}
\end{equation}
where $w$ is the threshold, and $w_i = \frac{w - \text{tr}(\bm{\Sigma}_{i})}{\sum_{i=1}^{m}[w -\text{tr}(\bm{\Sigma}_{i})]}$ is a normalized weight.

%% file: experiment.tex
\section{Experiment}
\label{sec:experiment}

We perform simulated and real-world experiments on three platforms to test the performance of M-LOAM. 
First, we calibrate multi-LiDAR systems on all the presented platforms. 
The proposed algorithm is compared with SOTA methods, and two evaluation metrics are introduced. 
Second, we demonstrate the SLAM performance of M-LOAM in various scenarios covering indoor environments and outdoor urban roads.
Moreover, to evaluate the sensibility of M-LOAM against extrinsic error, we test it on the handheld device and vehicle under different levels of extrinsic perturbation.
Finally, we provide a study to comprehensively evaluate M-LOAM's performance and computation time with different LiDAR combinations.

\begin{figure}[]
	\centering
	\subfigure[The Real handheld device.]
	{\label{fig:rhd_device}\centering\includegraphics[width=0.27\textwidth]{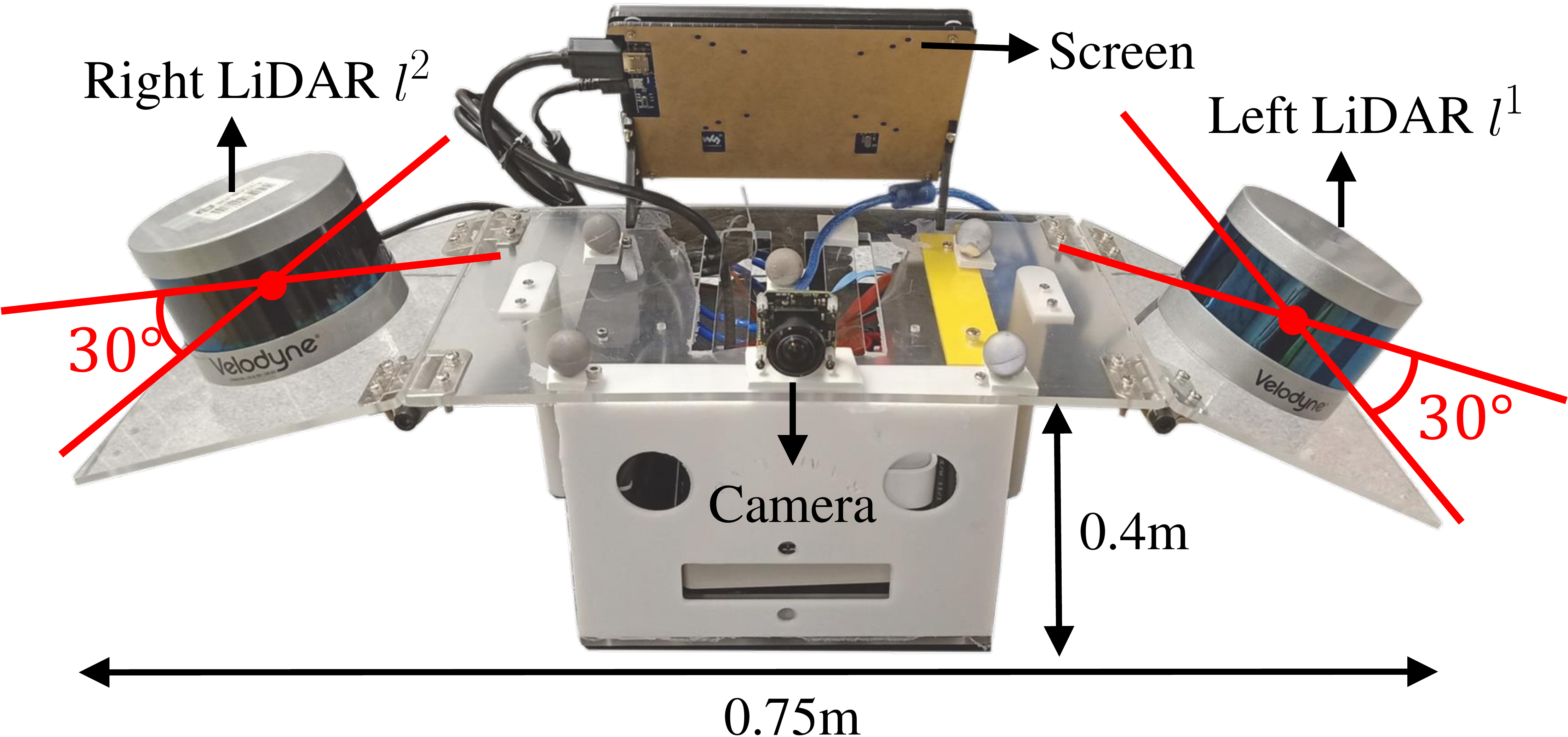}}
	\hfill
	\subfigure[Calibrated point cloud.]
	{\label{fig:rhd_pc}\centering\includegraphics[width=0.2\textwidth]{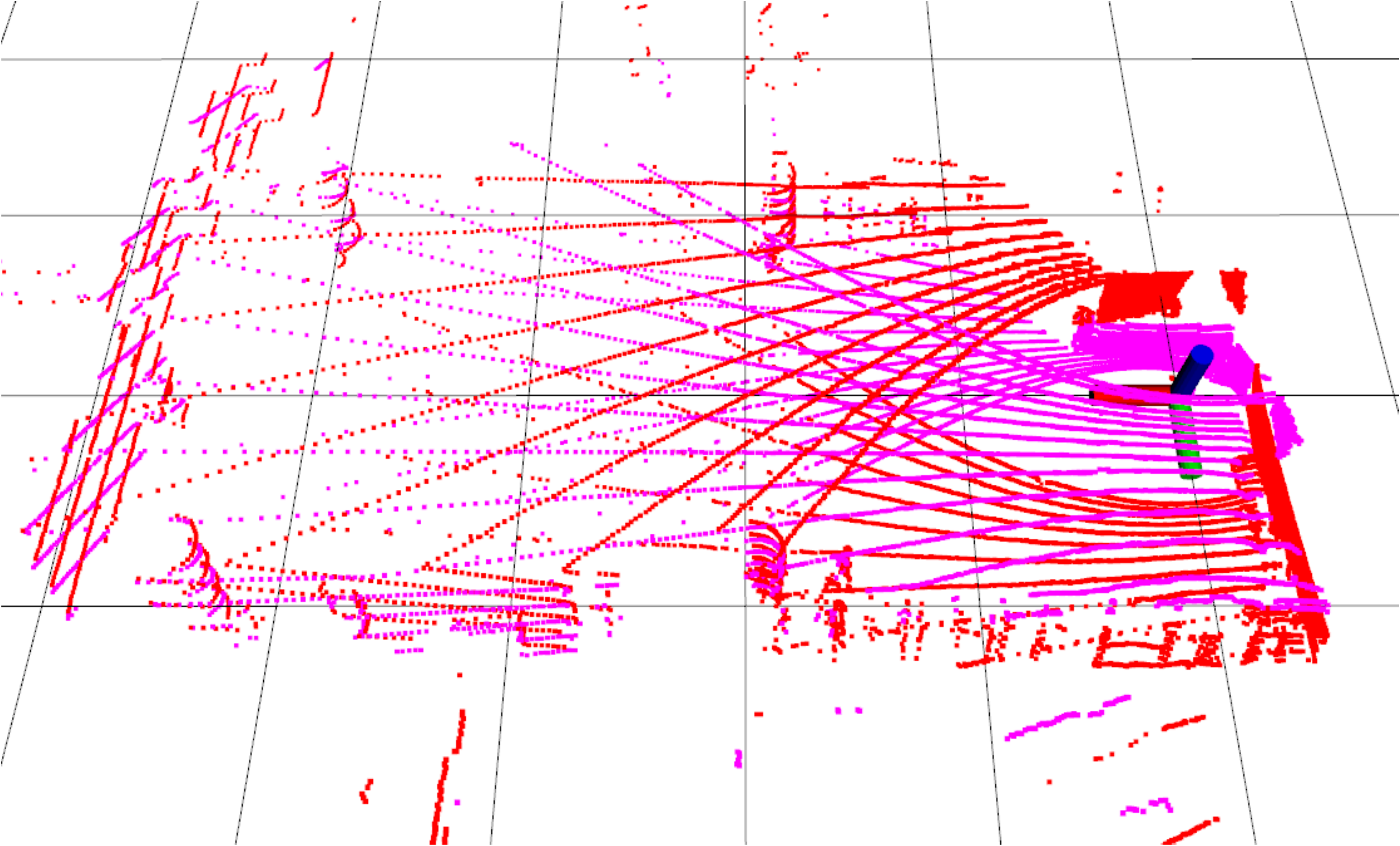}}
	\caption{(a) The real handheld device for indoor tests. Two VLP-16s are mounted at the left and right sides respectively. The attached camera is used to record test scenes. (b) The calibrated point cloud consists of points from the left (red) and right (pink) LiDARs.}
	\label{fig:rhd_device_pc}  
\end{figure}  

\begin{figure}[]
	\centering
	\subfigure[The Real vehicle.]
	{\label{fig:rv_device}\centering\includegraphics[width=0.24\textwidth]{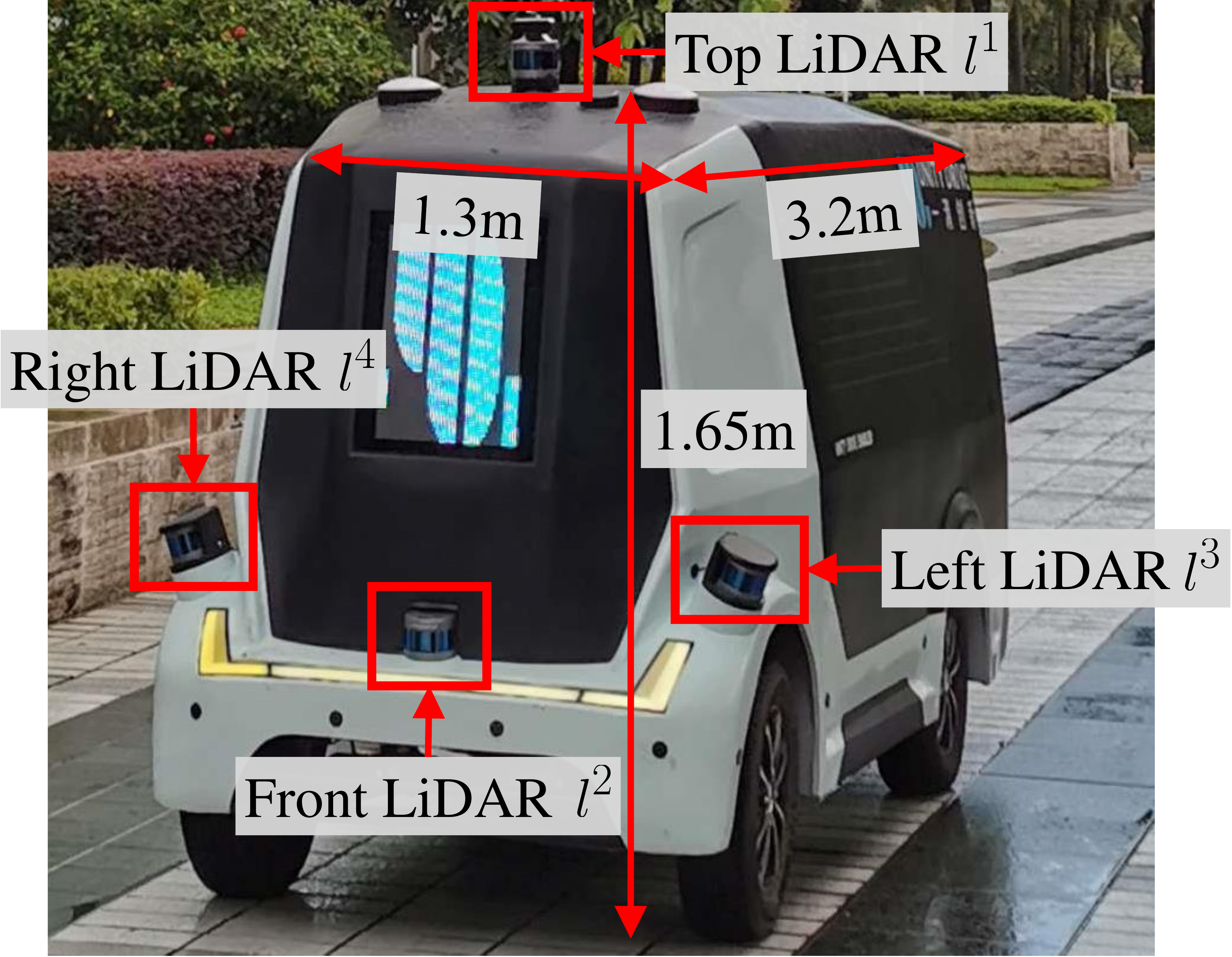}}
	\hfill
	\subfigure[Calibrated point cloud.]
	{\label{fig:rv_pc}\centering\includegraphics[width=0.23\textwidth]{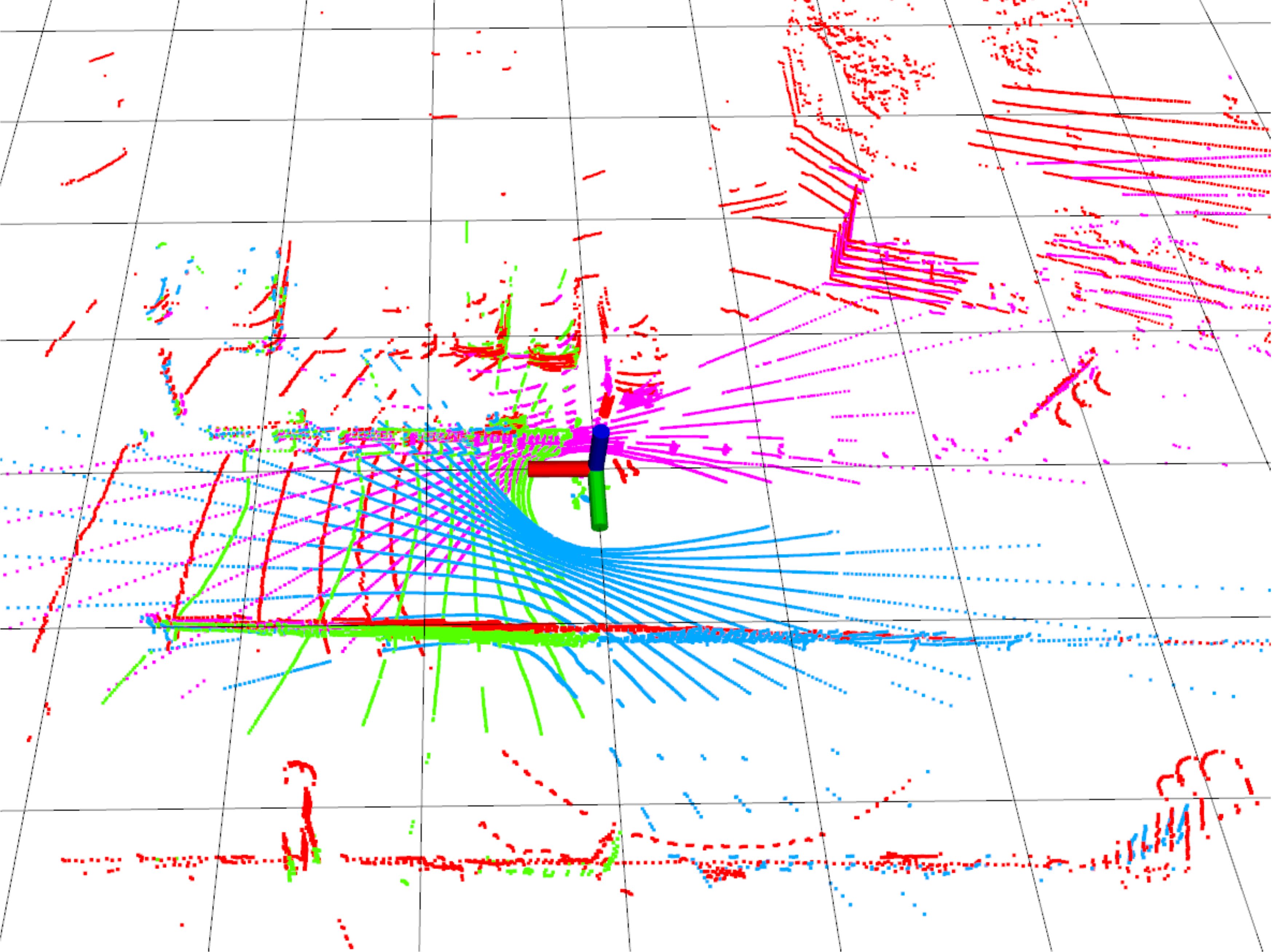}}
	\caption{(a) The real vehicle for large-scale, outdoor tests. Four RS-16s are mounted at the top, front, left, and right position respectively. (b) The calibrated point cloud consists of points from the top (red), front (green), left (blue), and right (pink) LiDARs.}
	\label{fig:rv_device_pc}  
\end{figure}  

\begin{table}[t]
	\centering
	\caption{Parameters for calibration and SLAM.}
	\renewcommand\arraystretch{1.0}
	\renewcommand\tabcolsep{7pt}	
	\begin{tabular}{ccccccc}
		\toprule
		$\sigma_{\mathbf{R}}$ & $\lambda_{calib}$ & $\mathcal{L}_{calib}$ & $N$     & $p$    & $w$  & $\alpha$ \\ 
		\midrule
		$0.25$ & $70$   &  $25$  & $4$   &   $2$    &  $100$ & $\geq1$ \\ 
		\bottomrule
	\end{tabular} 
	\label{tab:parameter}
\end{table}

%% file: exper_implement.tex
\begin{table*}[]
	\centering
	\caption{Calibrated extrinsics. $\downarrow$ indicates that the lower the value, the better the score.}
	\renewcommand\arraystretch{1.25}
	\renewcommand\tabcolsep{3pt}
	\begin{tabular}{ccrrrrrrrrrr}
	\toprule
	\multirow{2}{*}{Case} & \multirow{2}{*}{Method} & \multicolumn{3}{c}{Rotation {$[deg]$}} & \multicolumn{3}{c}{Translation {$[m]$}} &
	\multirow{2}{*}{$EGT_{\mathbf{R}}\ [deg, \downarrow]$} & 
	\multirow{2}{*}{$EGT_{\mathbf{t}}\ [m, \downarrow]$}   &
	\multicolumn{2}{c}{$\overline{MME}\ [\downarrow]$} \\ \cline{3-8} \cline{11-12} 
	& &  \multicolumn{1}{c}{$x$} & \multicolumn{1}{c}{$y$} & \multicolumn{1}{c}{$z$} & \multicolumn{1}{c}{$x$} & \multicolumn{1}{c}{$y$} & \multicolumn{1}{c}{$z$} & & & \multicolumn{1}{c}{$r=0.3m$} & \multicolumn{1}{c}{$r=0.4m$} \\ 
	\midrule
	\multirow{6}{*}{\rotatebox[]{90}{SR1 (Left-Right)}}       
	& Auto-Calib              
	& $6.134$   & $1.669$     & $0.767$    & $0.001$      & $-0.635$    & $-0.083$     & $33.911$ & $0.209$ & $-2.016$ & $-2.463$ \\ 
	& Proposed (Ini.)         
	& $44.154$  & $7.062$     & $1.024$    & $-0.027$     & $-0.719$    & $0.000$      & $8.229$ & $0.328$ & $-2.240$ & $-2.685$ \\ 
	& Proposed (Ini.+Ref.)
	& $40.870$  & $0.397$     & $0.237$  & $-0.012$   & $-0.475$  & $-0.206$   & $\bm{0.997}$ & $\bm{0.018}$ & $\bm{-2.690}$ & $\bm{-3.073}$\\ 
	\cline{2-12}
	& PS-Calib                
	& $\darkgraytext{40.021}$     & $\darkgraytext{-0.005}$     & $\darkgraytext{-0.010}$   & $\darkgraytext{0.001}$       & $\darkgraytext{-0.476}$    & $\darkgraytext{-0.218}$    & $\darkgraytext{0.037}$ & $\darkgraytext{0.003}$ & $\darkgraytext{-2.730}$ & $\darkgraytext{-3.115}$ \\ 
	& W/O Calib               
	& $\darkgraytext{0.000}$     & $\darkgraytext{0.000}$        & $\darkgraytext{0.000}$     & $\darkgraytext{0.000}$     & $\darkgraytext{0.000}$     & $\darkgraytext{0.000}$     & $\darkgraytext{40.000}$ & $\darkgraytext{0.525}$ & $\darkgraytext{-2.358}$ & $\darkgraytext{-2.704}$ \\                           
	& GT                      
	& $\darkgraytext{40.000}$     & $\darkgraytext{0.000}$        & $\darkgraytext{0.000}$     & $\darkgraytext{0.000}$     & $\darkgraytext{-0.477}$    & $\darkgraytext{-0.220}$    & $\darkgraytext{-}$ & $\darkgraytext{-}$ & $\darkgraytext{-2.733}$ & $\darkgraytext{-3.111}$ \\ 
	\midrule
	
	\multirow{6}{*}{\rotatebox[]{90}{SR2 (Left-Right)}}
	& Auto-Calib              
	& $4.680$     & $-1.563$    & $0.647$  & $0.032$      & $-0.751$    & $-0.022$     & $35.337$            & $0.339$          & $-2.336$          & $-2.447$          \\  
	& Proposed (Ini.)         
	& $40.854$     & $3.517$     & $0.285$  & $-0.019$     & $-0.667$    & $0.000$     & $3.632$                 & $0.291$          & $-2.607$      & $-2.804$               \\  
	& Proposed (Ini.+Ref.)
	& $38.442$ & $0.111$ & $-0.037$ & $0.000$ & $-0.504$ & $-0.205$ & $\bm{1.549}$ & $\bm{0.030}$ & $\bm{-3.016}$ & $\bm{-3.192}$ \\ 

	\cline{2-12}
	& PS-Calib                
	& $\darkgraytext{40.021}$     & $\darkgraytext{-0.005}$       & $\darkgraytext{-0.010}$   &$ \darkgraytext{0.001}$    & $\darkgraytext{-0.476}$    & $\darkgraytext{-0.218}$    & $\darkgraytext{0.0365}$ & $\darkgraytext{0.003}$ & $\darkgraytext{-3.113}$ & $\darkgraytext{-3.306}$ \\                           
	& W/O Calib               
	& $\darkgraytext{0.000}$     & $\darkgraytext{0.000}$        & $\darkgraytext{0.000}$     & $\darkgraytext{0.000}$       & $\darkgraytext{0.000}$      & $\darkgraytext{0.000}$        &  $\darkgraytext{40.000}$   & $\darkgraytext{0.525}$ & $\darkgraytext{-2.875}$ & $\darkgraytext{-2.878}$ \\ 
	& GT                      
	& $\darkgraytext{40.000}$     & $\darkgraytext{0.000}$        & $\darkgraytext{0.000}$     & $\darkgraytext{0.000}$       & $\darkgraytext{-0.477}$      & $\darkgraytext{-0.220}$       & $-$ & $-$ & $\darkgraytext{-3.117}$ & $\darkgraytext{-3.313}$ \\ 
	\midrule
	
	\multirow{6}{*}{\rotatebox[]{90}{RHD (Left-Right)}}
	& Auto-Calib              
	& $7.183$     & $-3.735$      & $33.329$     & $0.653$  & $-2.006$    & $-0.400$   & $44.312$ & $1.612$ & $-3.612$          & $-2.711$ \\
	& Proposed (Ini.)         
	& $36.300$     & $0.069$      & $-3.999$     & $0.113$  & $-0.472$    & $-0.103$   & $6.443$ & $0.112$ & $-3.664$                & $-2.839$                \\ 
	& Proposed (Ini.+Ref.) 
	& $37.545$     & $-0.376$      & $0.773$     & $0.066$  & $-0.494$    & $-0.113$   & $\bm{2.491}$ & $\bm{0.064}$ & $\bm{-3.681}$ & $-\bm{2.862}$ \\  
	\cline{2-12}
	& CAD Model               
	& $\darkgraytext{40.000}$     & $\darkgraytext{0.000}$        & $\darkgraytext{0.000}$     & $\darkgraytext{0.000}$       & $\darkgraytext{-0.456}$     & $\darkgraytext{-0.122}$     & $\darkgraytext{2.077}$ & $\darkgraytext{0.092}$ & $\darkgraytext{-3.662}$        & $\darkgraytext{-2.833}$         \\                          		
	& W/O Calib               
	& $\darkgraytext{0.000}$     & $\darkgraytext{0.000}$        & $\darkgraytext{0.000}$     & $\darkgraytext{0.000}$       & $\darkgraytext{0.000}$      & $\darkgraytext{0.000}$      & $\darkgraytext{40.000}$ & $\darkgraytext{0.560}$ & $\darkgraytext{-3.696}$  & $\darkgraytext{-2.868}$   \\ 
	& PS-Calib
	& $\darkgraytext{39.629}$     & $\darkgraytext{-1.664}$       & $\darkgraytext{1.193}$      & $\darkgraytext{0.033}$     & $\darkgraytext{-0.540}$    & $\darkgraytext{-0.142}$    & $\darkgraytext{-}$ & $\darkgraytext{-}$ & $\darkgraytext{-3.696}$   & $\darkgraytext{-2.868}$   \\                           		
	\midrule
	
	\multirow{6}{*}{\rotatebox[]{90}{RV (Top-Front)}}
	& Auto-Calib              
	& $-19.634$    & $21.610$       & $-3.481$    & $-0.130$      & $-0.282$    & $-0.850$     & $22.852$ & $0.791$   & $-2.705$ & $-2.282$                    \\ 
	& Proposed (Ini.)         
	& $1.320$     & $7.264$        & $3.011$     & $-0.324$      & $0.227$    & $0.000$      & $3.217$ & $1.433$   & $-2.721$       & $-2.332$                            \\ 
	& Proposed (Ini.+Ref.) 
	& $-2.057$& $6.495$    & $2.133$ & $0.528$  & $-0.036$ & $-1.102$  & $\bm{0.274}$ & $\bm{0.081}$   & $\bm{-2.885}$      & $\bm{-2.370}$  \\  
	\cline{2-12}
	& CAD Model               
	& $\darkgraytext{0.000}$     & $\darkgraytext{10.000}$        & $\darkgraytext{0.000}$     & $\darkgraytext{0.795}$       & $\darkgraytext{0.000}$      & $\darkgraytext{-1.364}$      & $\darkgraytext{4.505}$ & $\darkgraytext{0.351}$  & $\darkgraytext{-2.771}$          &  $\darkgraytext{-2.312}$  \\                          		
	& W/O Calib               
	& $\darkgraytext{0.000}$     & $\darkgraytext{0.000}$        & $\darkgraytext{0.000}$     & $\darkgraytext{0.000}$       & $\darkgraytext{0.000}$      & $\darkgraytext{0.000}$       & $\darkgraytext{7.227}$ & $\darkgraytext{1.252}$  & $\darkgraytext{-2.785}$     & $\darkgraytext{-2.306}$  \\  
	& PS-Calib
	& $\darkgraytext{-1.817}$    & $\darkgraytext{6.629}$        & $\darkgraytext{2.134}$     & $\darkgraytext{0.536}$       & $\darkgraytext{0.039}$      & $\darkgraytext{-1.131}$      & $\darkgraytext{-}$ & $\darkgraytext{-}$  & $\darkgraytext{-2.902}$    &  $\darkgraytext{-2.416}$  \\                            		
	\bottomrule
	\end{tabular}
	\label{tab:calibration_results}	
\end{table*}

\subsection{Implementation Details}
\label{sec:exp_implement}
We use the PCL library \cite{rusu20113d} to process point clouds and the Ceres Solver \cite{ceres-solver} to solve nonlinear least-squares problems.
In experiments which are not specified, our algorithm is executed on a desktop with an i7 CPU@4.20 GHz and 32 GB RAM.
Three platforms with different multi-LiDAR systems are tested: a simulated robot, a handheld device, and a vehicle.
The LiDARs on real platforms are synchronized with the external GPS clock triggered at an ns-level accuracy.
\begin{itemize}
	\item \textbf{The Simulated Robot (SR)} 
	is built on the Gazebo \cite{koenig2004design}.
	Two 16-beam LiDARs are mounted on a mobile robot for testing.
	We build a closed simulated rectangular room.
	We use the approach from \cite{mou2018optimal} to set the LiDAR configuration for maximizing the sensing coverage. 
	We moved the robot in the room at an average speed of $0.5m/s$.
	The ground-truth extrinsics and poses are provided. 

	\item \textbf{The Real Handheld Device (RHD)} is made for handheld tests and shown in Fig. \ref{fig:rhd_device_pc}. Its configuration is similar to that of the SR.
	Besides two VLP-16s\footnote{\url{https://velodynelidar.com/products/puck}}, we also install a mini computer (Intel NUC) for data collection and a camera (mvBlueFOX-MLC200w) for recording test scenes.
	We used this device to collect data on the campus with an average speed of $2m/s$.
	
	\item \textbf{The Real Vehicle (RV)}
	is a vehicle for autonomous logistic transportation \cite{liu2021role}.
	We conduct experiments on this platform to demonstrate that our system also performs well in large-scale, challenging outdoor environments.     
	As shown in Fig. \ref{fig:rv_device_pc}, four RS-LiDAR-16s\footnote{\url{https://www.robosense.ai/rslidar/rs-lidar-16}} are rigidly mounted at the top, front, left, and right positions respectively.
	We drove the vehicle through urban roads at an average speed of $3m/s$.
	Ground-truth poses are obtained from a coupled LiDAR-GPS-encoder localization system that was proposed in \cite{zheng2019low,zhu2019real}.
\end{itemize}

Table \ref{tab:parameter} shows the parameters which are empirically set.
$\sigma_{\mathbf{R}}$, $\lambda_{calib}$, and $\mathcal{L}_{calib}$ are the convergence thresholds in calibration. 
Setting the last two parameters requires a preliminary training process, which is detailed in the supplementary material \cite{jiao2020supplementary}.
$p$ and $N$ are the size of the local map and the sliding window in the odometry respectively.
$w$ is the threshold of filtering uncertain points in mapping, and
$\alpha$ is the scale of the extrinsic covariance. 
We set $\alpha=10$ for the case of injecting large perturbation in Section \ref{sec:inject_calibration_error}. Otherwise, $\alpha=1$.

%% file: exper_calibration.tex
\subsection{Performance of Calibration}
\label{sec:exp_calibration}

\subsubsection{Evaluation Metrics} 
We introduce two metrics to assess the LiDAR calibration results from different aspects:
\begin{itemize}
	\item \textbf{Error Compared With Ground truth (EGT)} computes the distance between the ground truth and the estimated values in terms of rotation and translation as
	\begin{equation}
	\begin{aligned}
	EGT_{\mathbf{R}}
	&= 
	\big\|
	\ln(
	\mathbf{R}_{gt}
	\mathbf{R}_{est}^{-1}	
	)^{\vee}
	\big\|,
	\\
	EGT_\mathbf{t}
	&= 
	\big\|
	\mathbf{t}_{gt}-\mathbf{t}_{est}
	\big\|,
	\end{aligned}
	\end{equation}    
	
	\item \textbf{Mean Map Entropy (MME)} is proposed to measure the compactness of a point cloud \cite{droeschel2014local}. 
	It has been explored as a standard metric to assess the quality of registration if ground truth is unavailable \cite{razlaw2015evaluation}.
	Given a calibrated point cloud, the normalized mean map entropy is 
	\begin{equation}
	MME
	=
	\frac{1}{m}
	\sum_{i=1}^{m}
	\ln
	\big[
	\det (2\pi e\cdot\mathbf{C}_{\mathbf{p}_{i}})
	\big],
	\end{equation}
	where $m$ is the size of the point cloud and $\mathbf{C}_{\mathbf{p}_{i}}$ is the sampling covariance within a local radius $r$ around $\mathbf{p}_{i}$.
	For each calibration case, we select $10$ consecutive frames of point clouds that contain many planes and compute their average MME values for evaluation.
\end{itemize}

Since the perfect ground truth is unknown in real-world applications, 
we use the results of ``\textbf{PS-Calib}'' \cite{jiao2019a} as the ``relative ground truth'' to compute the \textit{EGT}.
PS-Calib is a well-understood, target-based calibration approach, which should have similar or superior accuracy to our method \cite{huang2018geometric}. 
Another metric is the MME, which computes the score in an unsupervised way. 
It can be interpreted as an information-theoretic measure of the compactness of a point cloud.

\begin{figure}[t]
	\centering
	\includegraphics[width=0.46\textwidth]{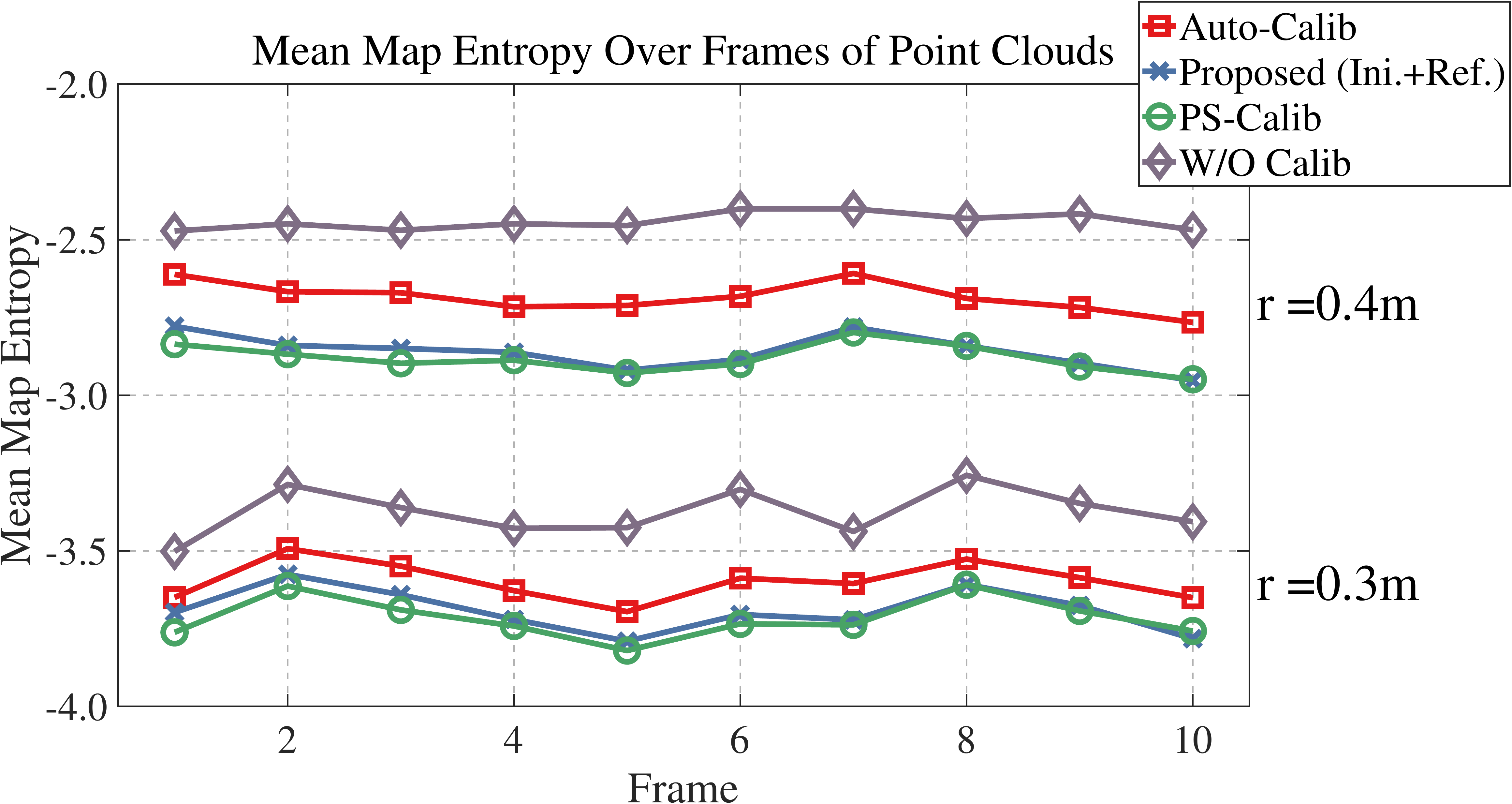}  
	\caption{The MME values over $10$ consecutive frames of point clouds which are calibrated by different approaches on the RHD platform. 
			 The lower the value, the better the score for a method.}
	\label{fig:experiment_mme}
\end{figure}

\subsubsection{Calibration Results}
\label{sec:calibration_result}
The multi-LiDAR systems of all the presented platforms are calibrated by our methods.
To initialize the extrinsics, we manually move these platforms with sufficient rotations and translations.
Table \ref{tab:calibration_results} reports the resulting extrinsics, where two simulated cases (same extrinsics, different motions) and two real-world cases are tested.
Due to limited space, we only demonstrate the calibration between the top LiDAR and front LiDAR on the vehicle.
Our method is denoted by ``\textbf{Proposed (Ini.+Ref.)}'', 
which is compared with an offline multi-LiDAR calibration approach \cite{jiao2019automatic} (``\textbf{Auto-Calib}'').
Although Auto-Calib follows a similar initialization-refinement procedure to obtain the extrinsics, it is different from our algorithm in several aspects.
For example, Auto-Calib only uses planar features in refinement.
And it assumes that LiDARs' views should have large overlapping regions. 
The hand-eye-based initial (``\textbf{Proposed (Ini.)}''),
uncalibrated (``\textbf{W/O Calib}''), CAD (``\textbf{CAD model}'' for real platforms), 
and ground-truth (``\textbf{GT}'') extrinsics are also provided for reference. 

Our hand-eye-based method successfully initializes the rotation offset ($<9deg$) for all cases, but fails to recover the translation offset ($>0.3m$) on the SR and RV.
Both the simulated robot and vehicle have to perform planar movement with a long distance for initialization, making the recovery of the $x-, y-$ translation poor due to the drift of motion estimation.
The planar movement also causes the $z-$ translation to be unobservable.
But we can move the RHD in 6-DoF and rapidly gather rich constraints. Its initialization results are thus good.
Regarding the online refinement, our algorithm outperforms Auto-Calib and demonstrates comparable performance with PS-Calib in terms of the EGT ($<3deg$ and $<0.07m$) and MME metrics.
Based on these results, we conclude that the initialization phase can provide coarse rotational estimates, and the refinement for precise extrinsics is required.

\begin{figure}[t]
	\centering
	\includegraphics[width=0.48\textwidth]{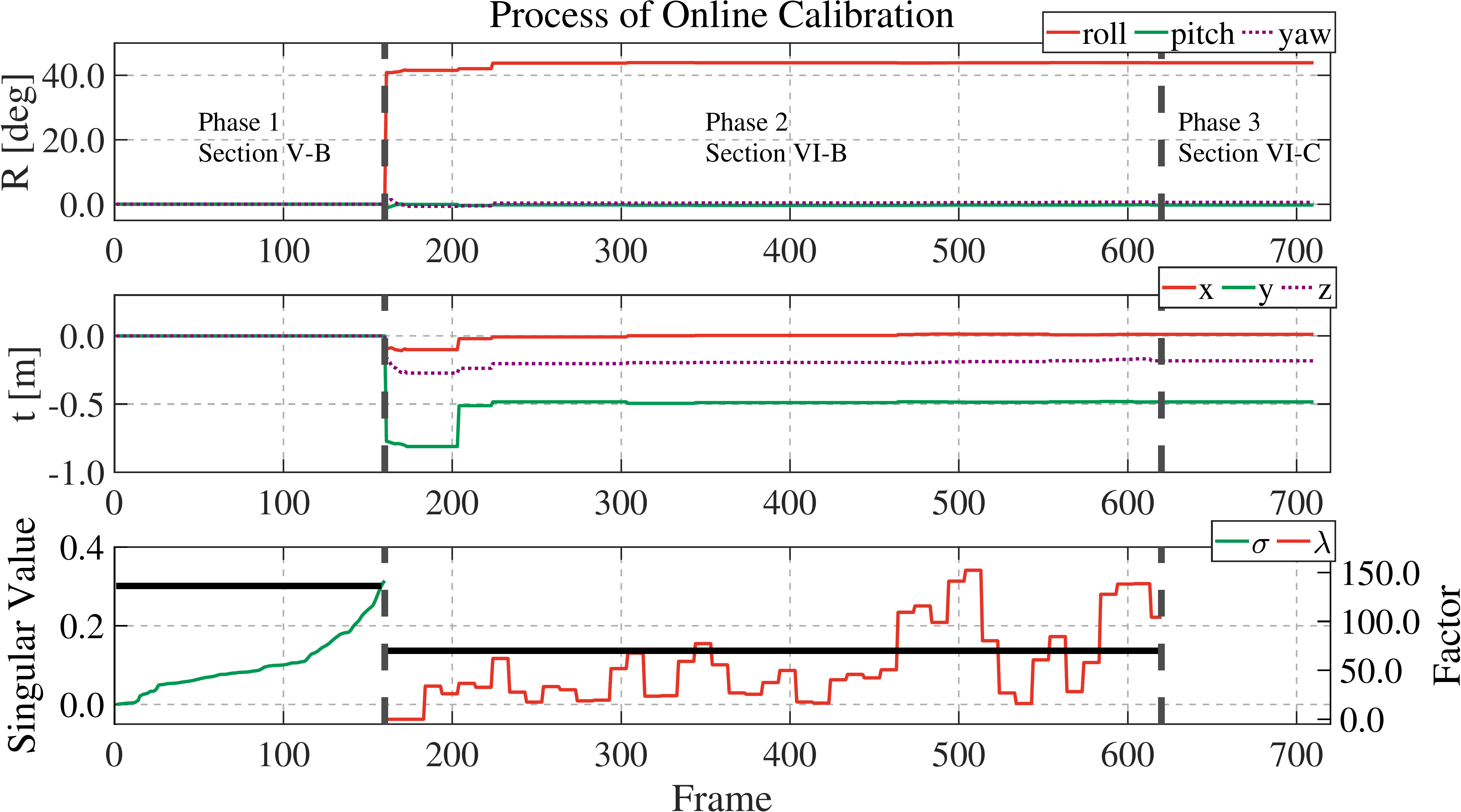} 
	\caption{Detailed illustration of the whole calibration process, including the initialization and optimization with online calibration on the RHD. Different phases are separated by bold dashed lines. In Phase 1, the initial rotation and translation are estimated with the singular value-based exit criteria (Section \ref{sec:initialization_calibration}). In Phase 2, the nonlinear optimization-based calibration refinement process is performed. 
	The convergence is determined by the \textit{degeneracy factor} (Section \ref{sec:calibration_termination}).  
	Phase 3 only optimizes the LiDAR odometry with fixed extrinsics. 
	The black lines in the bottom plot indicate the setting thresholds $\sigma_{\mathbf{R}}$ and $\lambda_{calib}$, which are defined in Section \ref{sec:calibration_termination}.}
	\label{fig:experiment_calibration_process}
\end{figure}

\begin{figure}[t]
	\centering
	\includegraphics[width=0.38\textwidth]{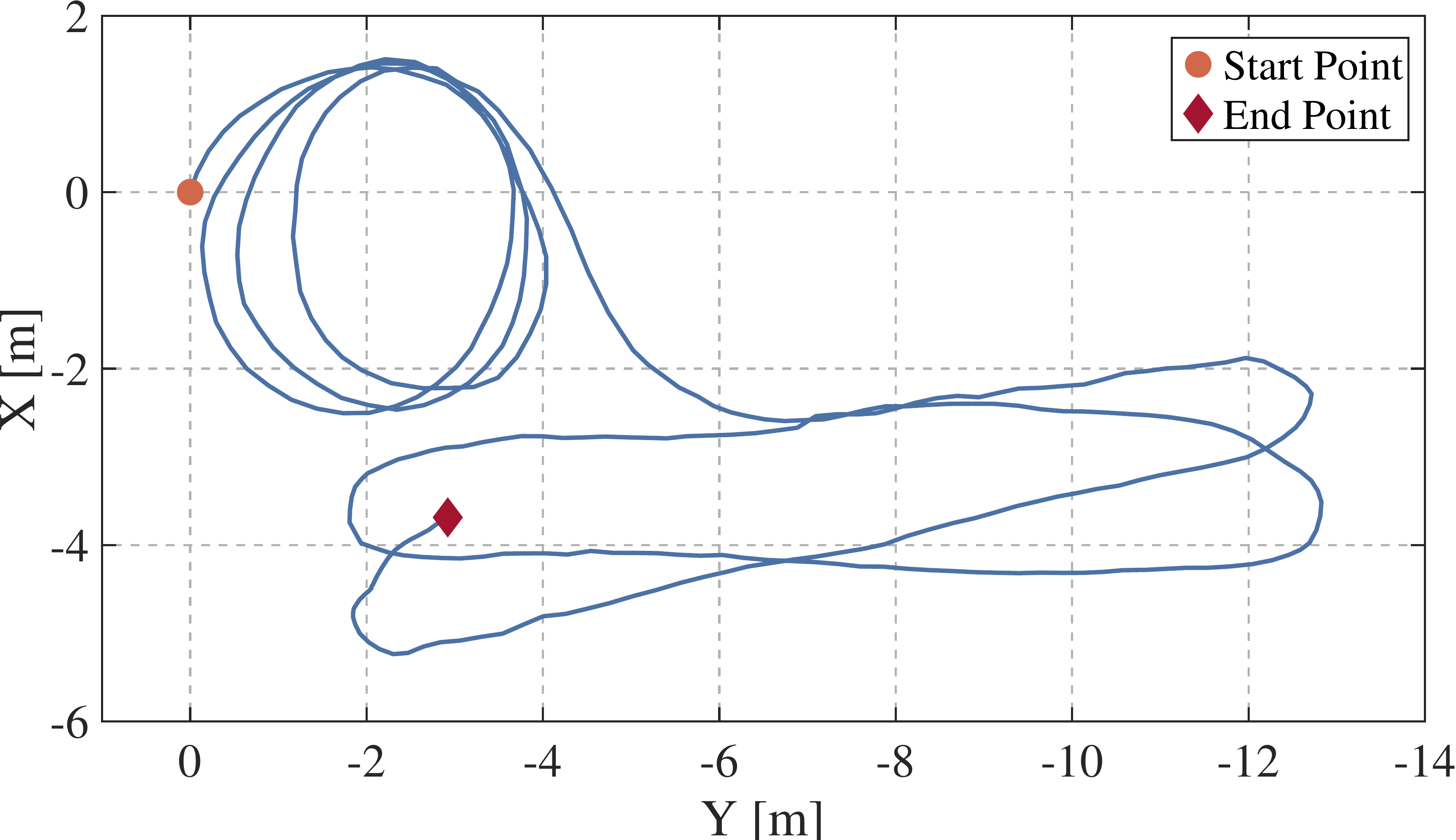}  
	\caption{Calibration trajectory of the sensor suite estimated by M-LO on the RHD. The dot and diamond indicate the start and end point respectively.}
	\label{fig:experiment_calibration_traj}
\end{figure}

We explicitly show the test on the RHD in detail.
In Fig. \ref{fig:experiment_mme}, we plot all MME values over different frames of calibrated point clouds, where the results are consistent with Table \ref{tab:calibration_results}.
Whether $r$ is set to $0.3m$ or $0.4m$, our method always has a better score than Auto-Calib.
Fig. \ref{fig:experiment_calibration_process} illustrates the calibration process, 
with the trajectory of the sensor suite shown in Fig. \ref{fig:experiment_calibration_traj}. 
This process is divided into three sequential phases: 
\textbf{Phase 1} (extrinsic initialization, Section \ref{sec:initialization_calibration}),
\textbf{Phase 2} (odometry with extrinsic refinement, Section \ref{sec:multilo_with_online_calibration}),
and \textbf{Phase 3} (pure odometry and mapping, Section \ref{sec:multilo_with_pure_odometry}).

Phase 1 starts with recovering the rotational offsets without prior knowledge about the mechanical configuration. 
It exits when the second small singular value of $\mathbf{Q}_{K}$ is larger than a threshold. 
The translational components are then computed. 
Phase 2 performs a nonlinear optimization to jointly refine the extrinsics. 
This process may last for a prolonged period if there are not sufficient environmental constraints. 
However, our sliding window-based marginalization scheme can ensure a bounded-complexity program to 
consistently update the extrinsics. 
The convergence condition is monitored with the \textit{degeneracy factor} (Section \ref{sec:calibration_termination})
and number of candidates. 
After convergence, we turn off the calibration and enter Phase 3 that is evaluated in Section \ref{sec:exp_slam}.

We also evaluate the sensitivity of our refinement method to different levels of initial guesses: 
the CAD model as well as rough rotational and translational initialization. Quantitative results can be found in the supplementary material \cite{jiao2020supplementary}.

\begin{figure}[t]
	\centering
	\includegraphics[width=0.48\textwidth]{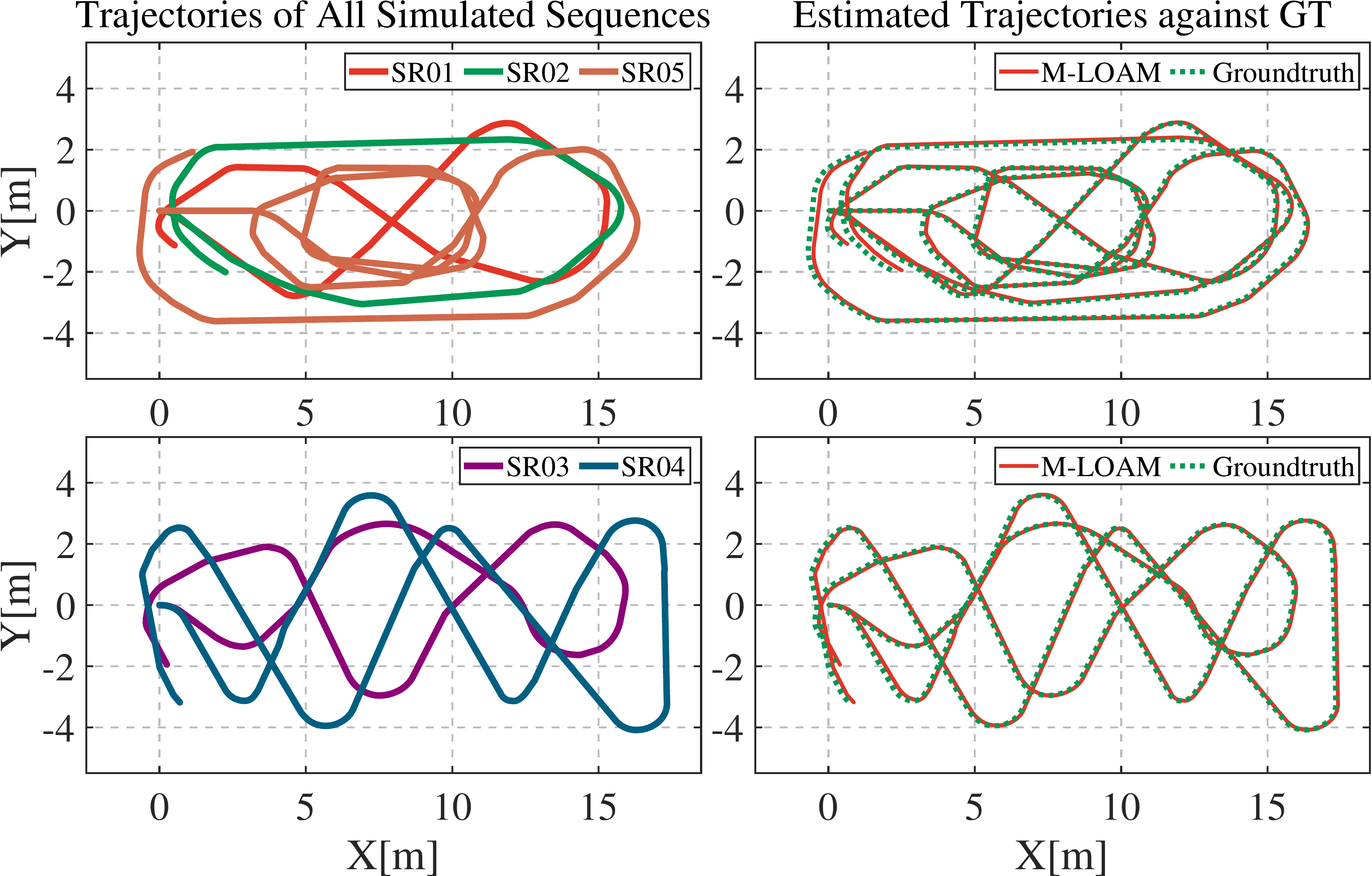}
	\caption{(Left) Trajectories of the \textit{SR01}-\textit{SR05} sequences with different lengths. 
			 (Right) M-LOAM's trajectories compared against the ground truth.}
	\label{fig:sr_trajectory}     
\end{figure}

\begin{table}[t]
	\centering
	\caption{ATE \cite{zhang2018tutorial} on simulated sequences}
	\renewcommand\arraystretch{1.25}
	\renewcommand\tabcolsep{2pt}        
	\begin{tabular}{cccccccc}
		\toprule
		Metric                              & Sequence & Length     & M-LO 
		& \begin{tabular}[c]{@{}c@{}}M-LOAM\\-wo-ua\end{tabular} &M-LOAM  & A-LO & A-LOAM \\ 
		\midrule
		\multirow{5}{*}{\rotatebox[]{90}{$\textbf{RMSE}_{\mathbf{t}}[m]$}}
		& \textit{SR01easy} & $40.6m$ & $0.482$ & $\bm{0.041}$ & $\bm{0.041}$ & $2.504$  & $0.060$ \\  
		& \textit{SR02easy} & $39.1m$ & $0.884$ & $\bm{0.034}$ & $\bm{0.034}$ & $3.721$  & $0.060$ \\  
		& \textit{SR03hard} & $49.2m$ & $0.838$ & $\bm{0.032}$ & $\bm{0.032}$ & $4.738$  & $0.059$ \\  
		& \textit{SR04hard} & $74.2m$ & $0.757$ & $\bm{0.032}$ & $\bm{0.032}$ & $2.083$  & $0.388$ \\  
		& \textit{SR05hard} & $81.2m$ & $0.598$ & $\bm{0.033}$ & $\bm{0.033}$ & $4.841$  & $0.208$ \\  
		\midrule
		
		\multirow{5}{*}{\rotatebox[]{90}{$\textbf{RMSE}_{\mathbf{R}}[deg]$}} 
		& \textit{SR01easy} & $40.6m$ & $3.368$ & $0.824$ & $\bm{0.676}$ & $26.484$ & $0.751$ \\
		& \textit{SR02easy} & $39.1m$ & $7.971$ & $1.070$ & $0.882$ & $37.903$ & $\bm{0.576}$ \\
		& \textit{SR03hard} & $49.2m$ & $6.431$ & $0.994$ & $0.865$ & $38.923$ & $\bm{0.750}$ \\
		& \textit{SR04hard} & $74.2m$ & $5.728$ & $0.919$ & $0.772$ & $21.027$ & $\bm{0.711}$ \\
		& \textit{SR05hard} & $81.2m$ & $6.509$ & $0.754$ & $\bm{0.554}$ & $87.999$ & $2.250$ \\
		\bottomrule
	\end{tabular}
	\label{tab:sr_rmse_result}
\end{table}

\begin{figure}[t]
	\centering
	\includegraphics[width=0.45\textwidth]{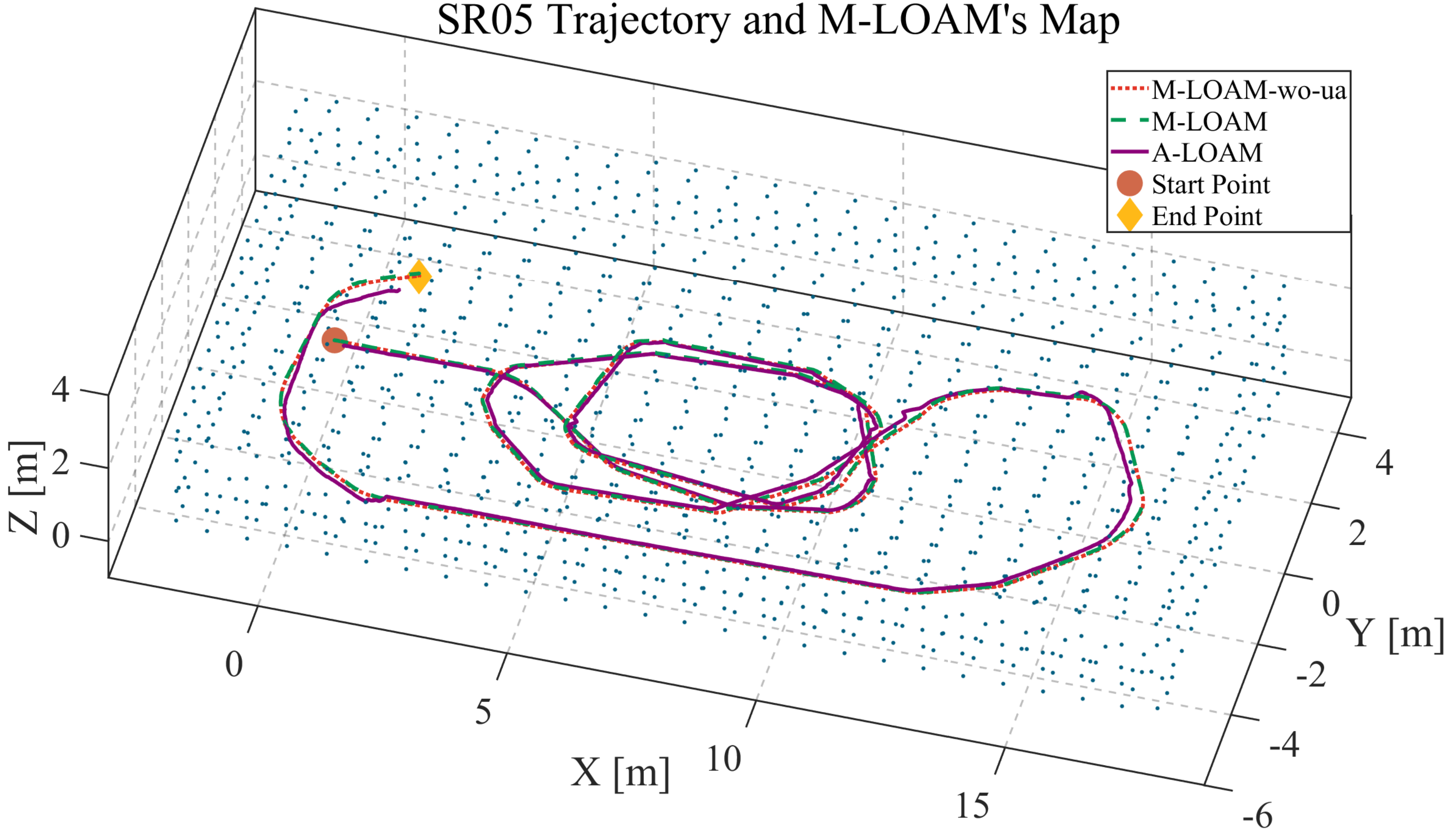}
	\caption{Trajectories on \textit{SR05} of M-LOAM-wo-ua, M-LOAM, and A-LOAM and the map constructed by M-LOAM. 
			 A-LOAM has a few defects, while M-LOAM-wo-ua's trajectory nearly overlaps with that of M-LOAM.}
	\label{fig:sr05_compare}     
\end{figure}

\begin{figure}[t]
	\centering
	\includegraphics[width=0.45\textwidth]{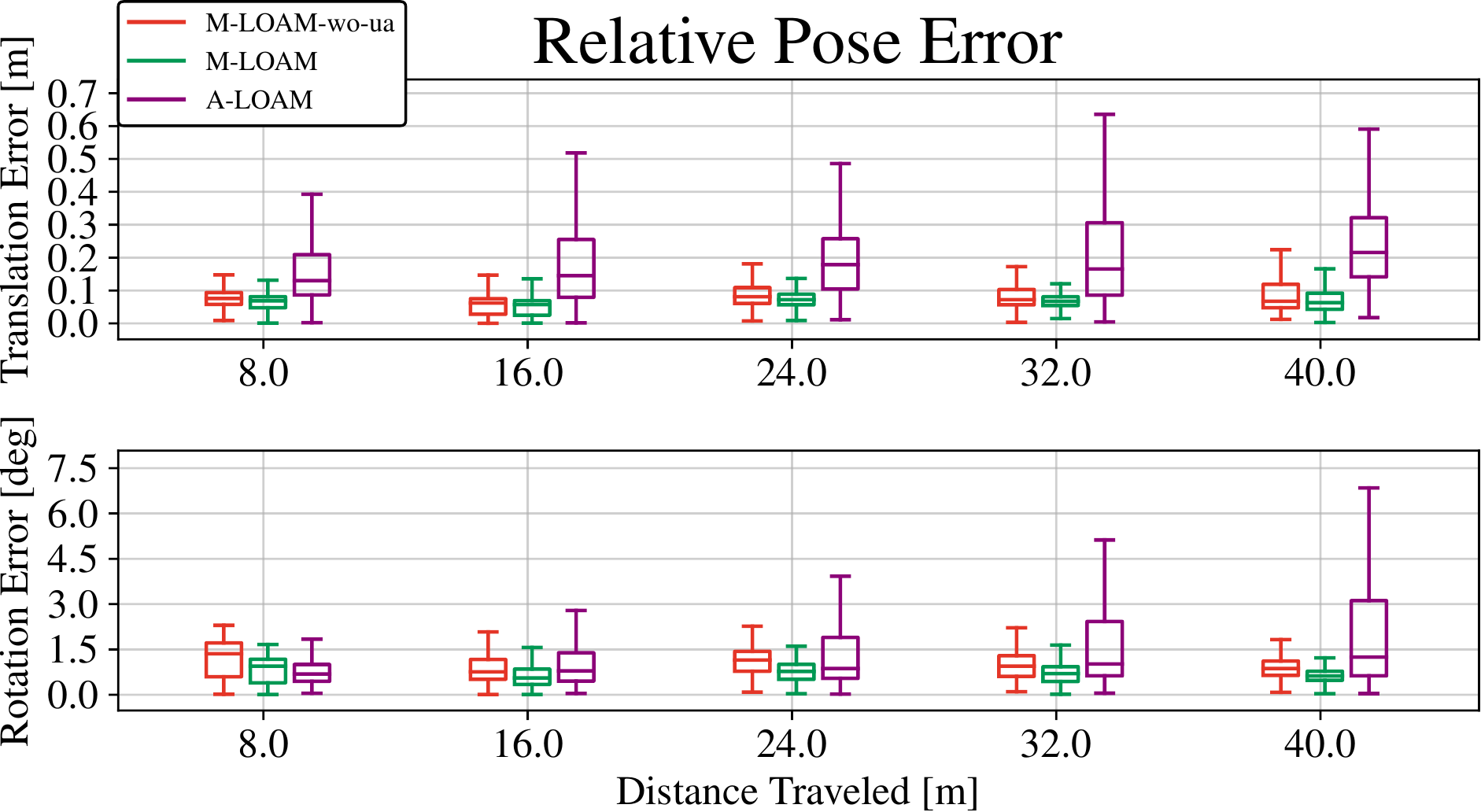}
	\caption{The mean RPE on \textit{SR05} with $10$ trials. For the distance $40m$, 
			 the median values of the relative translation and rotation error of M-LOAM-wo-ua, M-LOAM, and A-LOAM are 
		 	 ($0.87$deg, $0.07$m),
		 	 ($\textbf{0.62deg}$, $\textbf{0.06m}$), and
		 	 ($1.26$deg, $0.22$m) respectively.}
	\label{fig:sr05_rpe}     
\end{figure}

%% file: exper_overview_localization.tex
\begin{figure}[t]
	\centering
	\subfigure[Poses and the map with covariance visualization]    
	{\label{fig:rhd01_pc}\centering\includegraphics[width=0.35\textwidth]{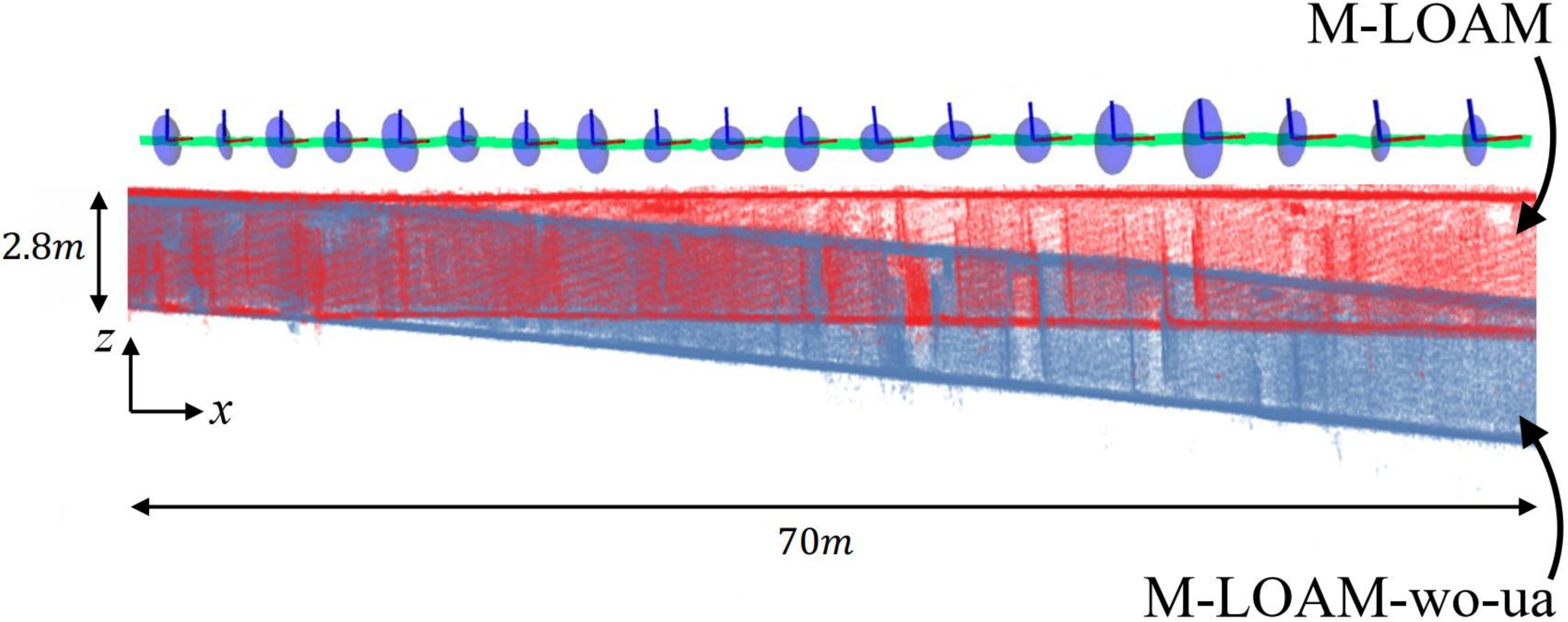}}
	\hfill
	\subfigure[Scene image]
	{\label{fig:rhd01_pitcture}\centering\includegraphics[width=0.13\textwidth]{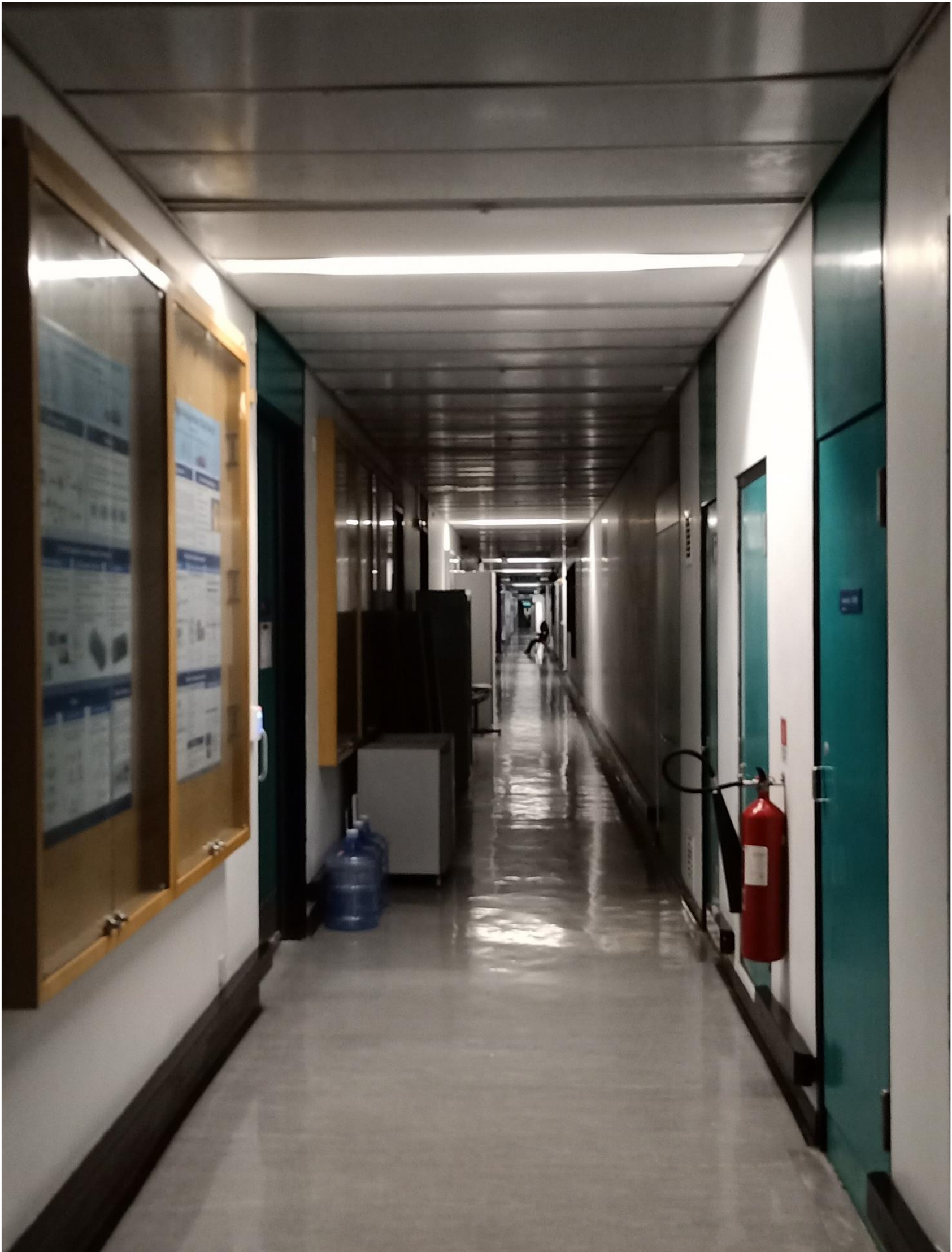}}
	\caption{(a) Side view of sample poses with covariances estimated by M-LOAM and generated map on \textit{RHD01corridor}. 
		     The below blue map is created by M-LOAM-wo-ua. 
		     The upper red map is created with M-LOAM. 
		     The covariances of pose calculated by M-LOAM are visualized as blue ellipses. 
		     A large radius represents a high uncertainty of a pose. The pose estimates in the $x\textendash, z\textendash$ direction are degenerate and uncertain, 
		     making the map points on the ceiling and ground noisy. M-LOAM is able to maintain the map quality by smoothing the noisy points. 
		     (b) The scene image.}
	\label{fig:rhd01}
\end{figure}

\begin{figure}[t]
	\centering
	\subfigure[Map and trajectories]
	{\label{fig:rhd02_xz_pc_traj}\centering\includegraphics[width=0.260\textwidth]{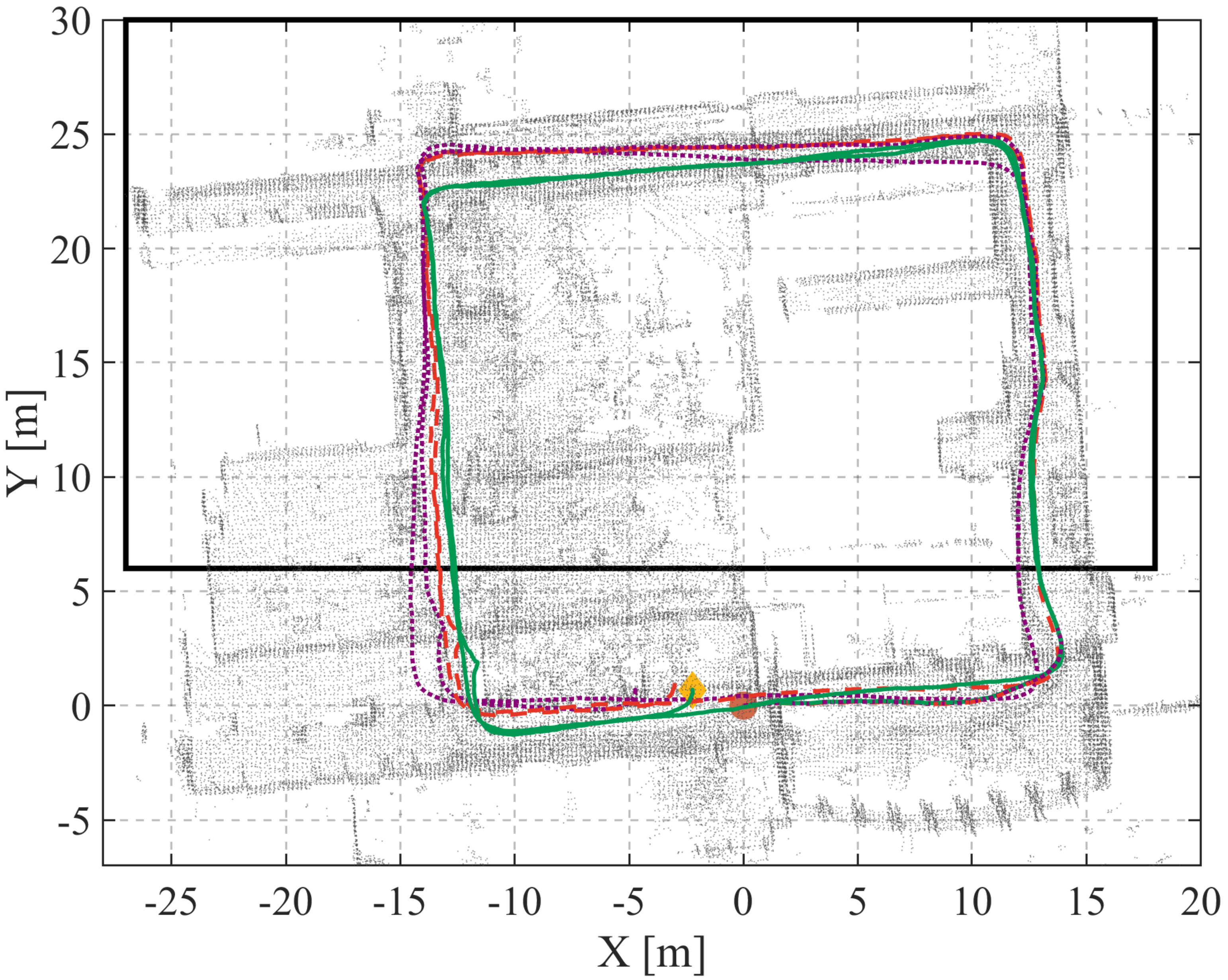}}
	\hfill	
	\subfigure[Trajectories]
	{\label{fig:rhd02_xz_traj}\centering\includegraphics[width=0.22\textwidth]{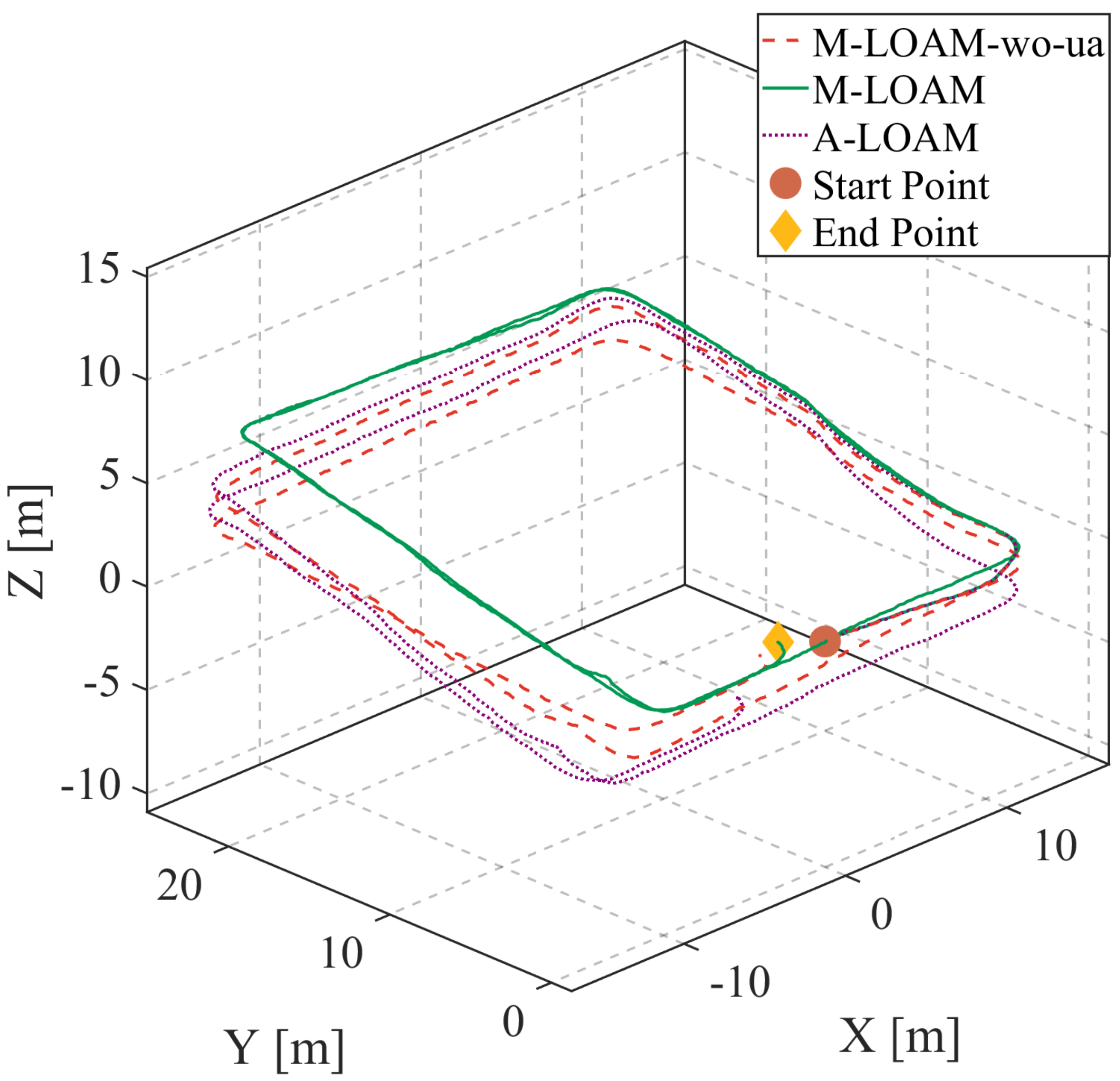}}
	\hfill
	\subfigure[Pose and map points with uncertainties]
	{\label{fig:rhd02_uct}\centering\includegraphics[width=0.281\textwidth]{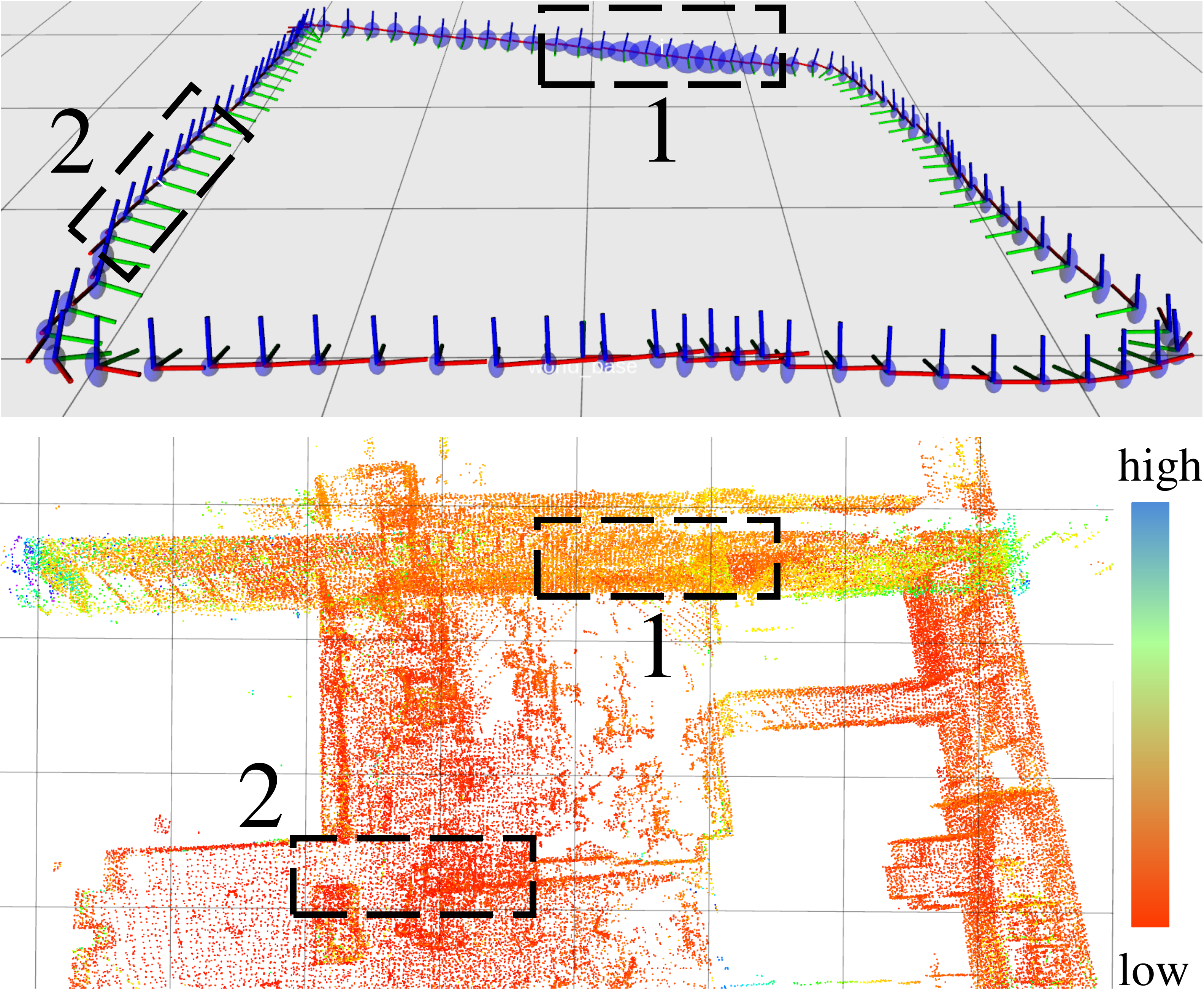}}  
	\hfill
	\subfigure[Scene images]
	{\label{fig:rhd02_pict}\centering\includegraphics[width=0.198\textwidth]{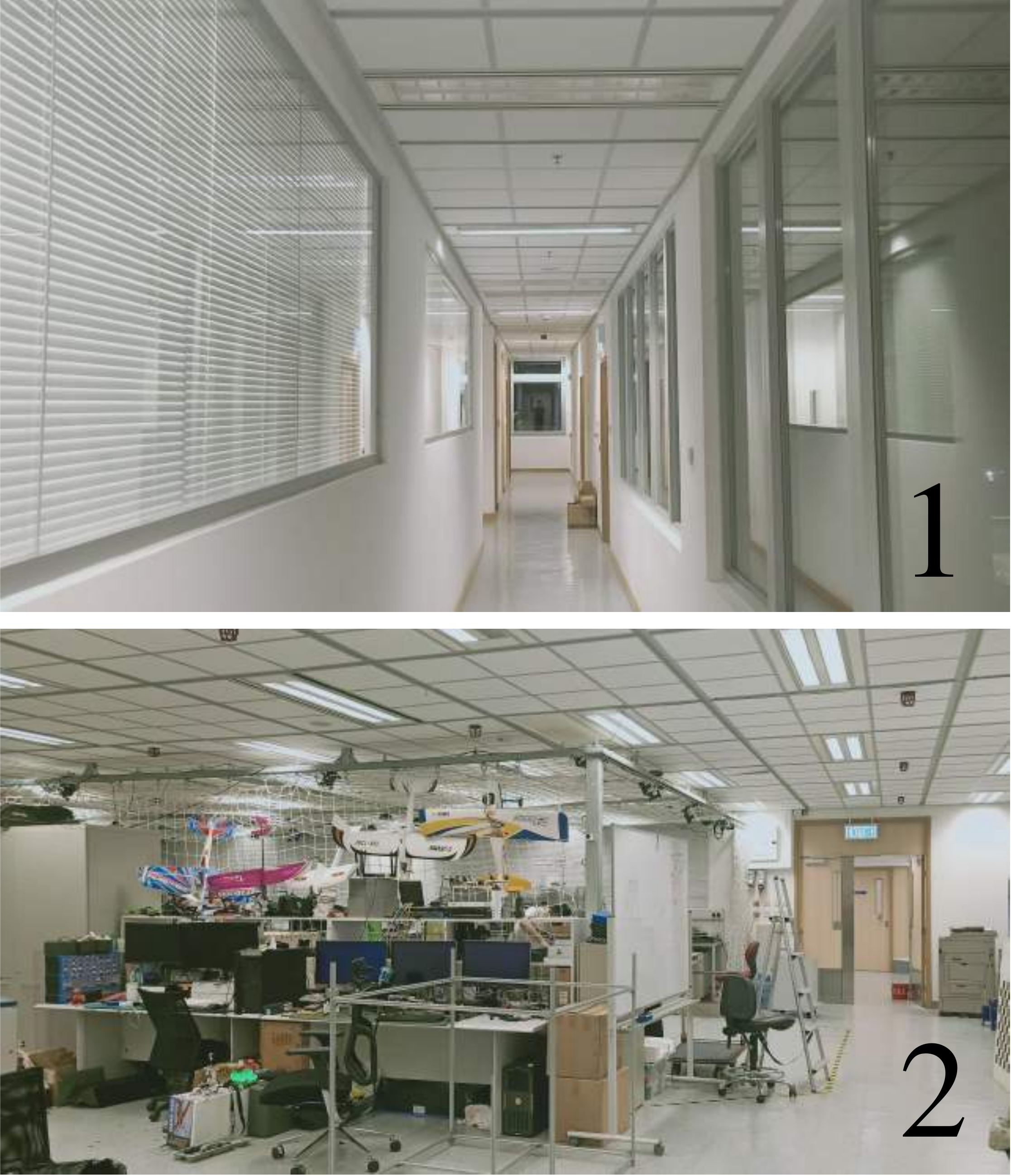}}
	\caption{Results on \textit{RHD02lab}. 
		     (a) Map generated in a laboratory and estimated trajectories (from right to left). 
		     The black box is the region shown in the bottom figures. Two loops are in this sequence. 
		     (b) Trajectories in another viewpoint. 
		     (c) Visualization of poses and map points uncertainty. The grid size is $5$m. Covariances of poses are represented as blue ellipses.
		     The larger the radius, the higher the uncertainty. 
		     The uncertainty of a point is measured by the trace of its covariance. 
		     The larger the trace, the higher the uncertainty. 
			 The marked regions indicate the degenerate (scene 1) and well-conditioned (scene 2) pose estimation respectively. 
			 With compounded uncertainty propagation, the map points in scene 1 become uncertain. 
		     (d) Scene images.}
	\label{fig:rhd02}  
\end{figure} 

\subsection{Performance of SLAM}
\label{sec:exp_slam}
We evaluate M-LOAM on both simulated and real-world sequences which are collected by the SR, RHD, and RV platforms.
The multi-LiDAR systems are calibrated with our online approach (Section \ref{sec:multilo_with_online_calibration}). 
The detailed extrinsics can be found on the first, third, and fourth row in Table \ref{tab:calibration_results}.
We compare M-LOAM with two SOTA, open-source LiDAR-based algorithms: 
\textbf{A-LOAM}\footnote{\url{https://github.com/HKUST-Aerial-Robotics/A-LOAM}} (the advanced implementation of LOAM \cite{zhang2014loam}) 
and \textbf{LEGO-LOAM}\footnote{\url{https://github.com/RobustFieldAutonomyLab/LeGO-LOAM}} \cite{shan2018lego}. 
Both of them directly take the calibrated and merged point clouds as input.
In contrast, our method formualtes the sliding-window estimator to fuse point clouds.
There are many differences in detail among these methods, as presented in the technical sections.
Overall, our system is more complete with online calibration, uncertainty estimation, and probabilistic mapping.
LEGO-LOAM is a ground-optimized system and requires LiDARs to be horizontally installed.
It easily fails on the SR and RV. 
We thus provide its results only on the RV sequences for a fair comparison. 
The results estimated by parts of M-LOAM are also provided. 
These are denoted by \textbf{M-LO} and \textbf{M-LOAM-wo-ua}, indicating our proposed odometry (Section \ref{sec:multilo_with_pure_odometry}) and the complete M-LOAM without the awareness of uncertainty (Section \ref{sec:application_uncertainty_propagation}), respectively.
To fulfill the real-time requirement, we run the odometry at $10$Hz and the mapping at $5$Hz.

%% file: exper_sr_localization.tex
\subsubsection{Simulated Experiment}
\label{sec:exp_slam_sr}

We move the SR to follow $5$ paths with the same start point to verify our method. 
Each sequence is performed with $10$-trial SLAM tests, and at each trial, zero-mean Gaussian noises with an SD of $0.05m$ are added onto the point clouds.
The ground-truth and the estimated trajectories of M-LOAM are plotted in Fig. \ref{fig:sr_trajectory}.
The absolute trajectory error (ATE) on all sequences is shown in Table \ref{tab:sr_rmse_result}, as evaluated in terms of root-mean-square error (RMSE) \cite{zhang2018tutorial}. 
All sequences are split into either an \textit{easy} or \textit{hard} level according to their length.
First, M-LO outperforms A-LO around $4-10$ orders of magnitudes, which shows that the sliding-window estimator can refine the frame-to-frame odometry.
Second, we observe that the mapping module greatly refines the odometry module.
Third, M-LOAM outperforms other methods in most cases.
This is due to two main reasons.
\textit{1)} The small calibration error may potentially affect the map quality and degrade M-LOAM-wo-ua's and A-LOAM's performance.
\textit{2)} The estimates from A-LO do not provide a good pose prior to A-LOAM.
Since the robot has to turn around in the room for exploration, A-LOAM's mapping error accumulates rapidly and thus makes the optimized poses worse.
This explains why A-LOAM has large error in \textit{SR04hard} and \textit{SR05hard}.
One may argue that A-LOAM has less rotational error than other methods on \textit{SR02easy} -- \textit{SR04hard}.
We explain that A-LOAM uses ground points to constrain the roll and pitch angles, while M-LOAM tends to filter them out.

We show the results of \textit{SR05hard} in detail.
The estimated trajectories are shown in Fig. \ref{fig:sr05_compare}.
A-LOAM has a few defects in the marked box region and at the tail of their trajectories, while M-LOAM-wo-ua's trajectory nearly overlaps with that of M-LOAM. 
The relative pose errors (RPE) evaluated by \cite{zhang2018tutorial} are shown in Fig. \ref{fig:sr05_rpe}. 
In this plot, M-LOAM has lower rotation and translation errors than others over a long distance.

%% file: exper_rhd_localization.tex
\begin{figure}[]
	\centering
	\subfigure[Map and M-LOAM's trajectory.]
	{\label{fig:rhd03_pc}\centering\includegraphics[width=0.308\textwidth]{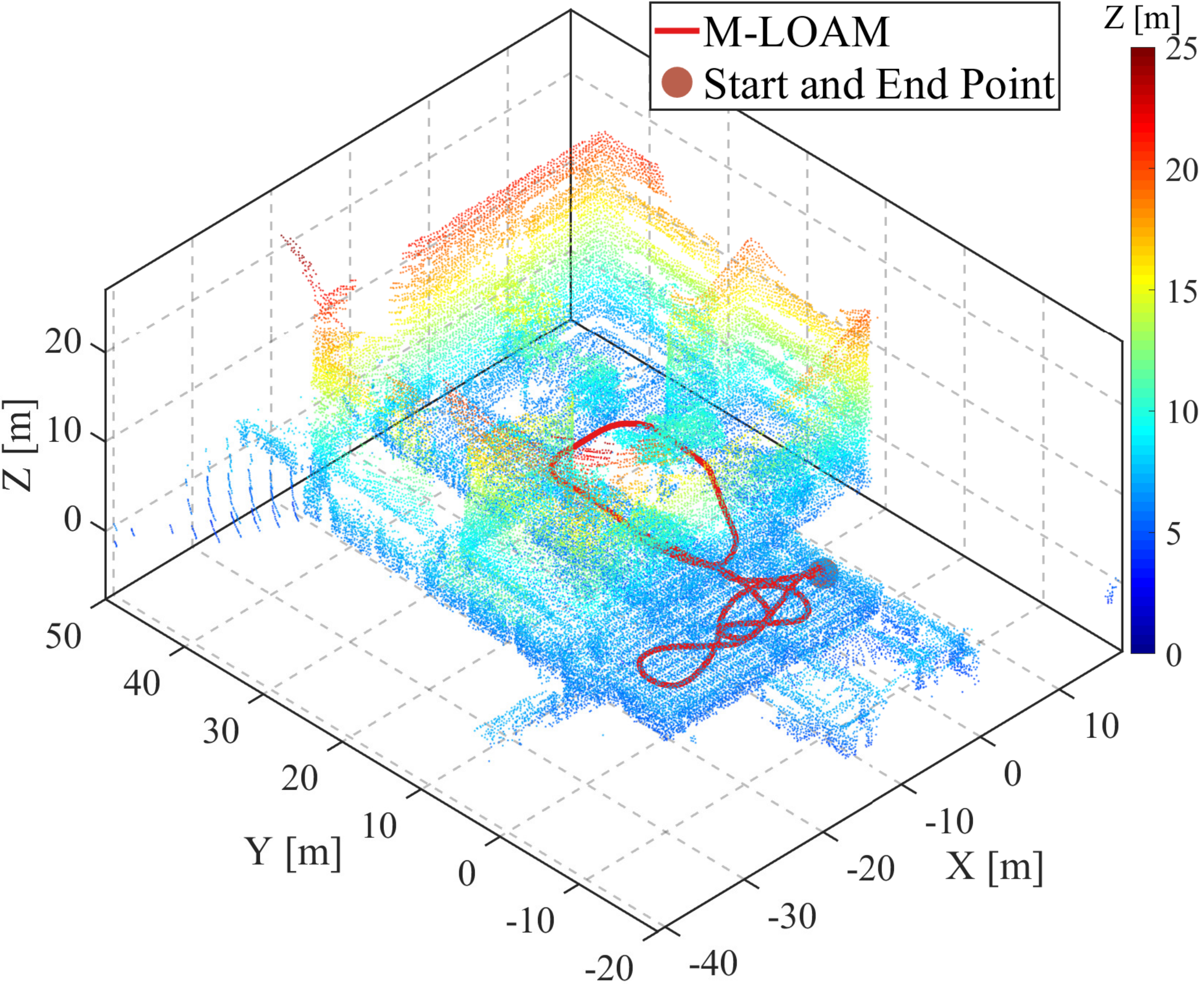}}
	\subfigure[Scene image.]    
	{\label{fig:rhd03_picture}\centering\includegraphics[width=0.17\textwidth]{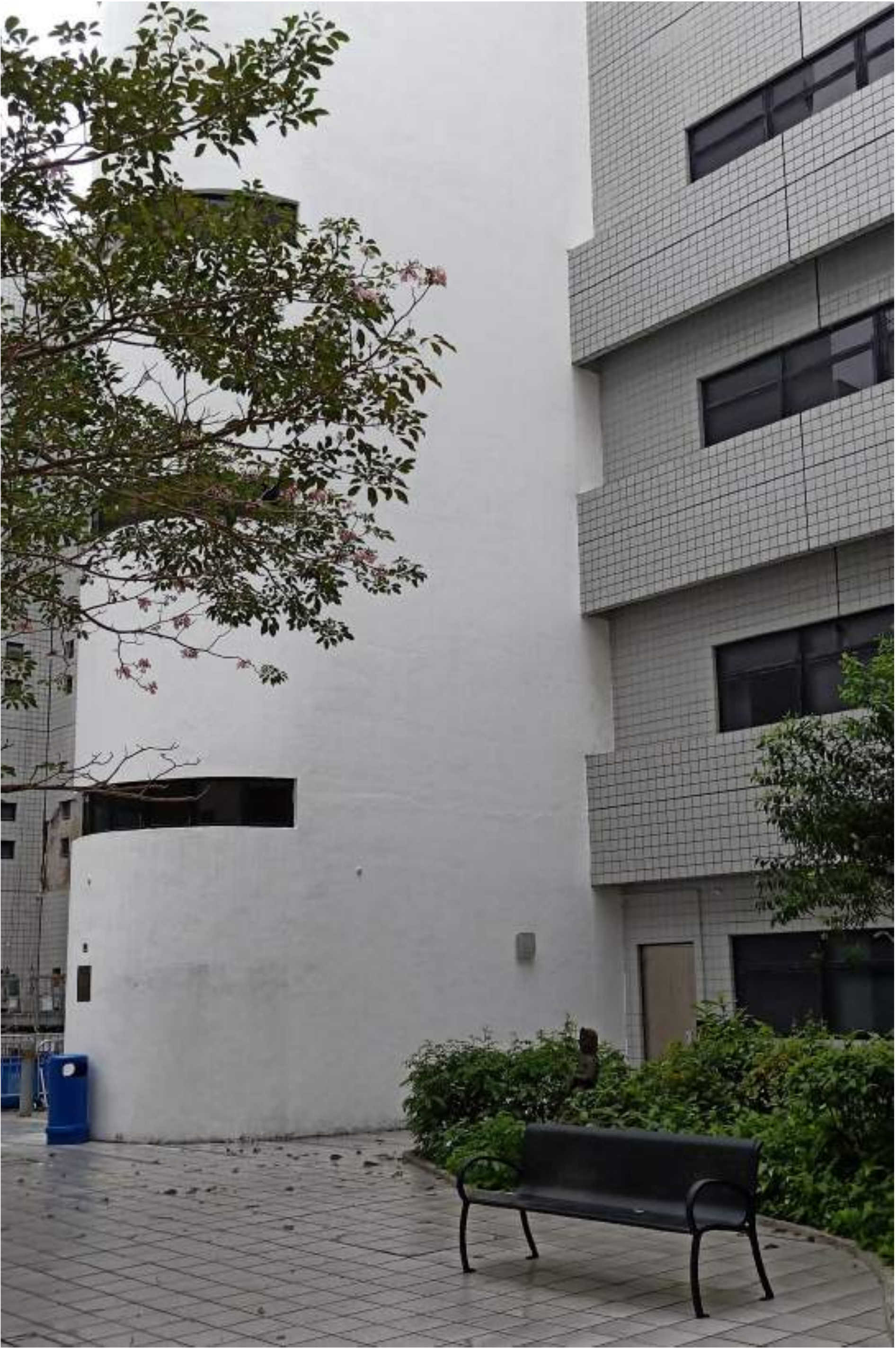}}
	\caption{Results of \textit{RHD03garden}. (a) Map generated in a garden, and the trajectory estimated by M-LOAM. The colors of the points vary from blue to red, indicating the altitude changes ($0m$ to $23m$) (b) Scene images.}
	\label{fig:rhd03}
\end{figure}  

\begin{figure}[t]
	\centering
	\includegraphics[width=0.49\textwidth]{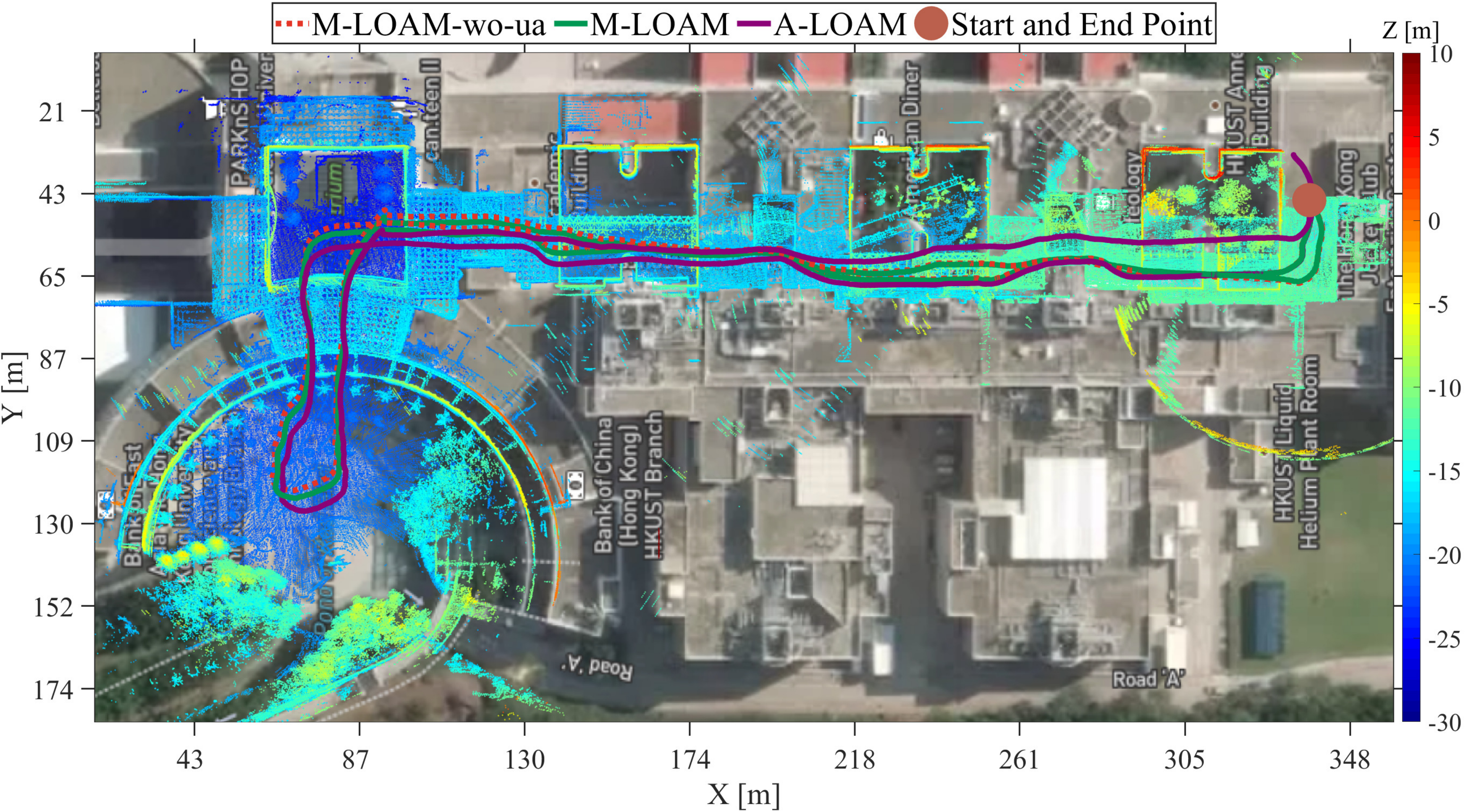}
	\caption{Mapping results of \textit{RHD04building} that goes through the HKUST academic buildings and the trajectories estimated by different methods (total length is $700m$). 
			 The map is aligned with Google Maps. The colors of the points vary from blue to red, indicating the altitude changes ($0m$ to $40m$)}
	\label{fig:rhd04building}    
\end{figure}  

\begin{table}[t]
	\centering
	\caption{Mean Relative Pose Drift}
	\renewcommand\arraystretch{1.25}
	\renewcommand\tabcolsep{3pt}
	\begin{tabular}{crrrrrr}
		\toprule
		\multicolumn{1}{c}{Sequence}    & Length & \multicolumn{1}{c}{M-LO}
		& \multicolumn{1}{c}{\begin{tabular}[c]{@{}c@{}}M-LOAM\\-wo-ua\end{tabular}} & \multicolumn{1}{c}{M-LOAM} & \multicolumn{1}{c}{A-LO} &\multicolumn{1}{c}{A-LOAM} \\ 
		\midrule
		\textit{RHD02}       & $197m$  & $3.82\%$ & $0.29\%$        & $\bm{0.07}\%$  & $14.18\%$ & $1.13\%$ \\ 
		\textit{RHD03}    & $164m$  & $0.88\%$ & $\bm{0.029}\%$  & $0.044\%$      & $5.31\%$  & $0.32\%$ \\ 
		\textit{RHD04}  & $695m$  & $7.30\%$ & $0.007\%$       & $\bm{0.003}\%$ & $34.02\%$ & $6.03\%$ \\ 
		\bottomrule
	\end{tabular}
	\label{tab:rhd_pose_drift}
\end{table}

\subsubsection{Indoor Experiment}
\label{sec:indoor_exp}
We used the handheld device to collect four sequences called \textit{RHD01corridor}, \textit{RHD02lab}, \textit{RHD03garden}, and \textit{RHD04building} to test our approach.

The first experiment is done in a long and narrow corridor. As emphasized in \cite{censi2007accurate}, this is a typical poorly-constrained environment. 
Here we show that the uncertainty-aware operation is beneficial to our system. 
In Fig. \ref{fig:rhd01}, we illustrate the sample poses of M-LOAM and the generated map on \textit{RHD01corridor}. 
These ellipses represent the size of the pose covariances. A large radius indicates that the pose in the $x-, z-$ directions of each mapping step is uncertain.
This is mainly caused by the fact that only a small set of points scan the walls and ceiling, which cannot provide enough constraints.
Map points become uncertain due to noisy transformations.
The uncertainty-aware operation of M-LOAM is able to capture and discard uncertain points.
This leads to a map with a reasonaly good signal-to-noise ratio, which generally improves the precision of optimization.
This is the reason why M-LOAM outperforms M-LOAM-wo-ua.

We conduct more experiments to demonstrate the performance of M-LOAM on other RHD sequences.
For evaluation, these datasets contain at least a closed loop.
Our results of \textit{RHD02lab} are shown in Fig. \ref{fig:rhd02}.
This sequence contains two loops in a lab region.
Fig. \ref{fig:rhd02_xz_pc_traj} shows M-LOAM's map, and Fig. \ref{fig:rhd02_xz_traj} shows the trajectories estimated by different methods.
Both M-LOAM-wo-ua and A-LOAM accumulates significant drift at the $x\textendash, y\textendash, z\textendash$ directions after two loops, while M-LOAM's results are almost drift free.
Fig. \ref{fig:rhd02_uct} shows the estimated poses and map points in the first loop.
The covariances of the poses and points evaluated by M-LOAM are visible as ellipses and colored dots in the figure.
Besides the corridor in scene $1$, we also mark the well-conditioned environment in scene $2$ for comparison.
Fig. \ref{fig:rhd02_pict} shows the scene pictures.
The results fit our previous explanation that the points in scene $1$ are uncertain because of noisy poses.
In contrast, scene $2$ has more constraints for estimating poses, making the map points certain.

Fig. \ref{fig:rhd03} shows the results of \textit{RHD03garden}. 
Since the installation of LiDARs on the RHD has a large roll angle, the areas over a $20$m height are scanned. 
Another experiment is carried out in a longer sequence. 
This dataset lasts for $12$ minutes, and the total length is about $700m$. 
The estimated trajectories and M-LOAM's map are aligned with Google Map in Fig. \ref{fig:rhd04building}. 
Both M-LOAM-wo-ua and M-LOAM provide more accurate and consistent results than A-LOAM.

Finally, we evaluate the pose drift of methods with $10$ repeated trials on \textit{RHD02}--\textit{RHD04}.
We employ the point-to-plane ICP \cite{pomerleau2013comparing} to measure the distance between the start and end point.
This ground truth distance is used to compare with that of the estimates, and the mean relative drift is listed in Table \ref{tab:rhd_pose_drift}.
Both M-LOAM-wo-ua and M-LOAM achieve a similar accuracy on \textit{RHD03} and \textit{RHD04} since the surroundings of these sequences are almost well-conditioned.
We conclude that the uncertainty-aware operation is not really necessary in well-conditioned environments and well-calibrated sensors, 
but maximally reduces the negative effect of uncertainties.

%% file: exper_rv_localization.tex
\subsubsection{Outdoor Experiment}
\label{sec:outdoor_experiment}
The large-scale, outdoor sequence was recorded with the RV platform (Fig. \ref{fig:rv_device_pc}).
This sequence covers an area around $1100m$ in length and $450m$ in width and has $110m$ in height changes. 
The total path length is about $3.23km$. The data lasts for $38$ minutes, and contains the $10$-Hz point clouds from fdour LiDARs and $25$-Hz ground-truth poses. 
This experiment is very significant to test the stability and durability of M-LOAM. 

\begin{figure}[t]
	\centering
	\includegraphics[width=0.49\textwidth]{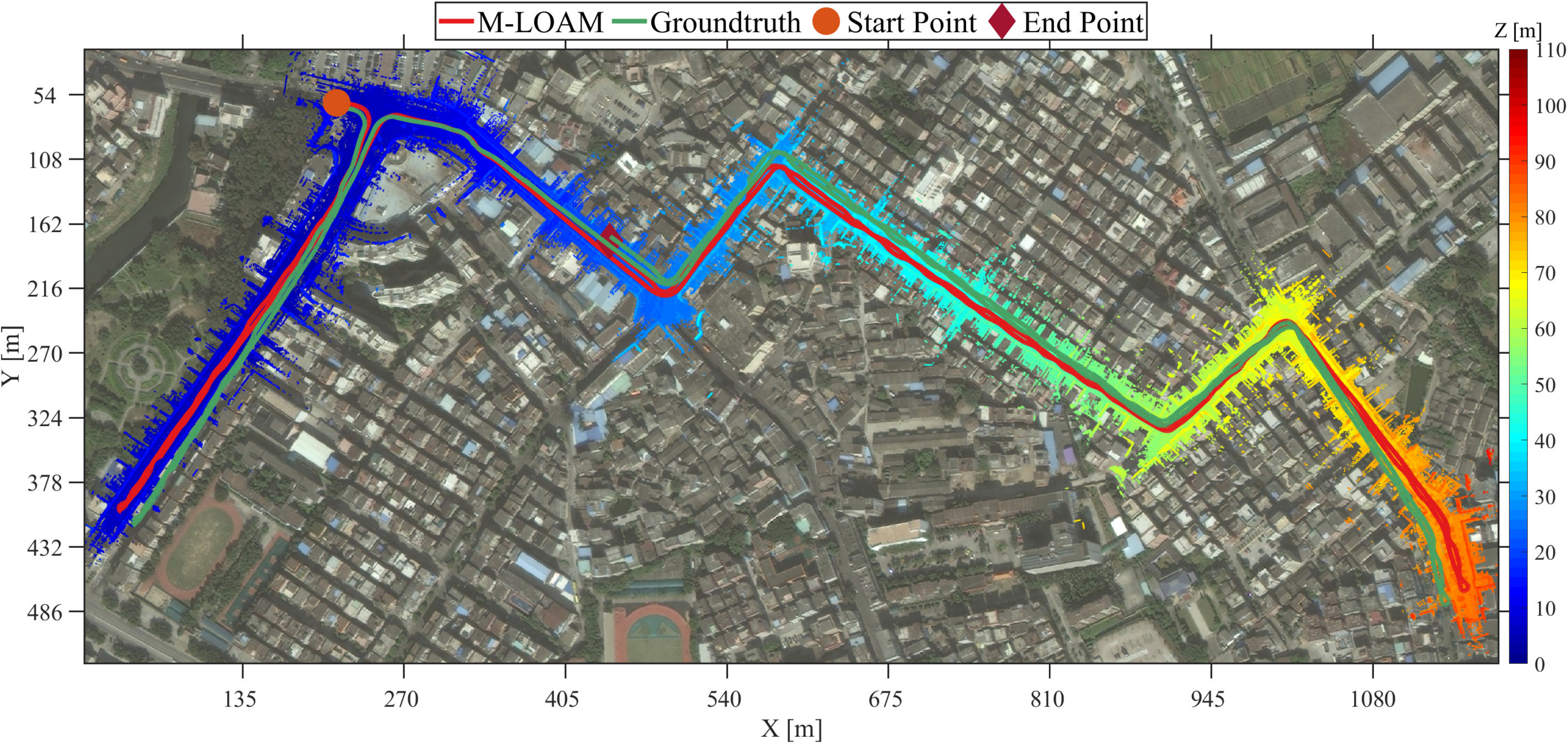}
	\caption{Mapping results of urban road and estimated trajectory against the ground truth on the RV sequence (total length is $3.23km$). 
			 The colors of the points vary from blue to red, indicating the altitude changes ($-5m$ to $105m$).}
	\label{fig:rv01street}    
\end{figure}

\begin{figure}[t]
	\centering
	\includegraphics[width=0.45\textwidth]{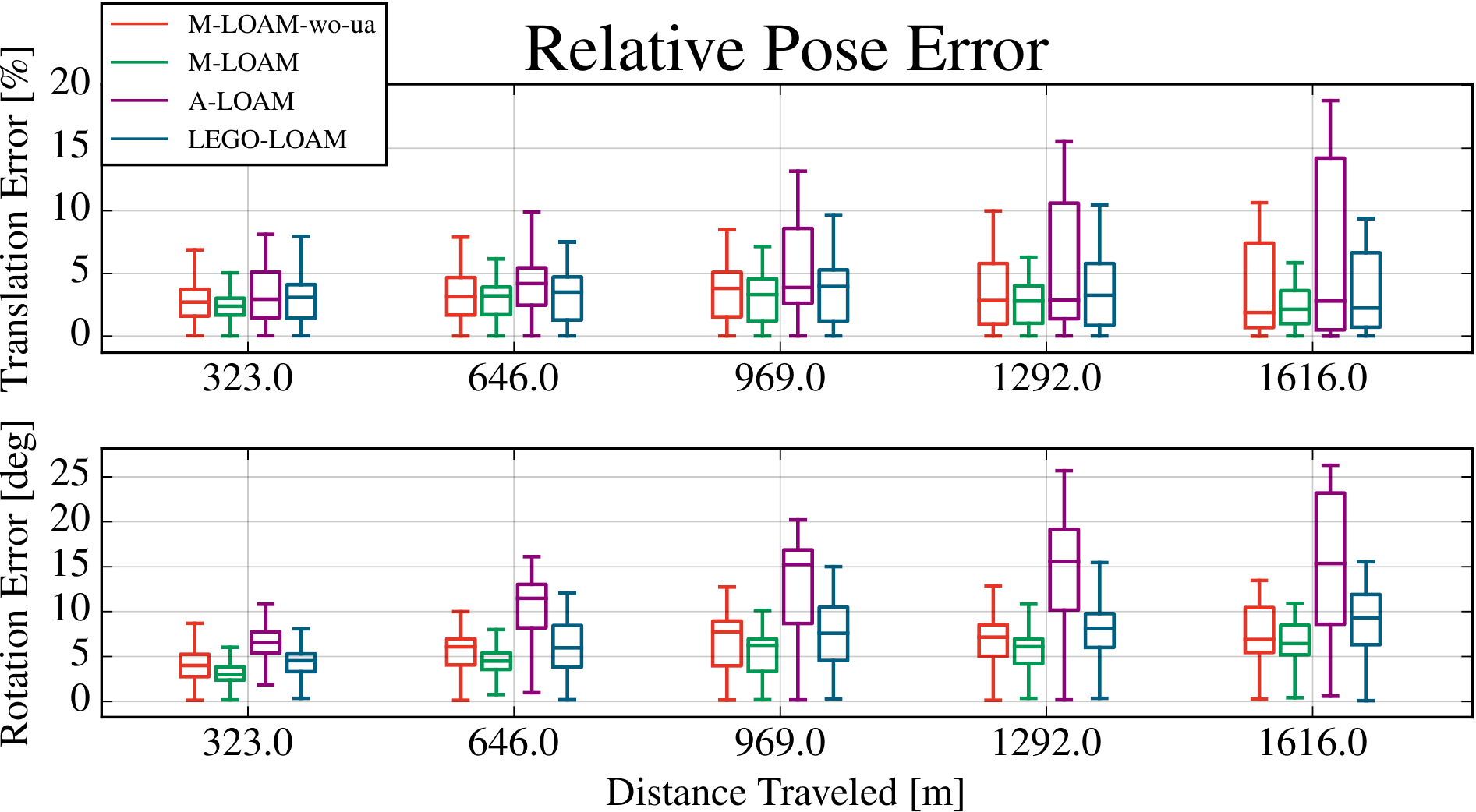}
	\caption{RPE on the RV sequence. For the $1616m$ distance, the median values of the relative translation (in percentage) 
			 and rotation error of M-LOAM-wo-ua, M-LOAM, A-LOAM, and LEGO-LOAM are ($\textbf{6.90}$\textbf{deg}, $\textbf{1.87\%}$), ($\textbf{6.45}$\textbf{deg}, 
			 $\textbf{2.14\%}$), ($15.36$deg, $2.80\%$), ($9.33$deg, $2.23\%$) respectively.}       
	\label{fig:rv_rpe}     
\end{figure}

M-LOAM's trajectory against the ground truth and the built map is aligned with Google Map in Fig. \ref{fig:rv01street}. 
We present the RPE of M-LOAM, M-LOAM-wo-ua, A-LOAM, and LEGO-LOAM in Fig. \ref{fig:rv_rpe}.
A-LOAM has the highest errors among them. 
Both M-LOAM-wo-ua and M-LOAM have competitive results with LEGO-LOAM. In addition, the outlier terms of M-LOAM are fewer than other methods.
We thus extend our previous findings that the uncertainty-aware mapping has the capability to enhance the robustness of the system.

%% file: exper_inject_perturbation.tex
\begin{table*}[t]
	\centering
	\caption{ATE given different extrinsics from bad to good: Inject perturbation, Initialization, and CAD model.}
	\renewcommand\arraystretch{1.25}
	\renewcommand\tabcolsep{3pt}
	\begin{tabular}{ccrrrrrrrrrr}
		\toprule
		\multirow{2}{*}{Case} & \multirow{2}{*}{Extrinsic Source} & \multicolumn{3}{c}{Rotation {$[deg]$}} & \multicolumn{3}{c}{Translation {$[m]$}} &
		\multicolumn{4}{c}{ATE: $\textbf{RMSE}_{\mathbf{t}}[m]\ (\textbf{RMSE}_{\mathbf{R}}[deg])$}
		\\ \cline{3-8} \cline{9-12}
		& & \multicolumn{1}{c}{$x$} & \multicolumn{1}{c}{$y$} & \multicolumn{1}{c}{$z$} & \multicolumn{1}{c}{$x$} & \multicolumn{1}{c}{$y$} & \multicolumn{1}{c}{$z$} & 
		\multirow{1}{*}{M-LOAM-wo-ua} & 
		\multirow{1}{*}{M-LOAM}   &
		\multirow{1}{*}{A-LOAM}   &
		\multirow{1}{*}{LEGO-LOAM} \\ 
		\midrule
		\multirow{3}{*}{\rotatebox[]{0}{\begin{tabular}[c]{@{}c@{}}RHD \\(Left-Right)	\end{tabular}}} 		
		& Inject Perturbation
		& $49.629$ & $8.236$   & $11.193$  & $0.133$ & $-0.440$& $-0.042$   & $7.74(29.63)$ & $\bm{0.88}(\bm{8.10})$ & $4.16(17.79)$   & $-$   \\ 		       
		& Initialization		
		& $36.300$ & $0.069$  & $-3.999$ & $0.113$  & $-0.472$& $-0.103$    & $0.90(\bm{6.46})$ & $\bm{0.79}(6.92)$ & $1.11(6.44)$ & $-$  \\ 		            		
		& CAD Model
		& $40.000$ & $0.000$    & $0.000$ & $0.000$   & $-0.456$ & $-0.122$ & $0.53(3.61)$ & $\bm{0.20}(\bm{2.43})$ & $0.53(3.94)$  & $-$\\ 
		\midrule
		\multirow{3}{*}{\rotatebox[]{0}{\begin{tabular}[c]{@{}c@{}}RV \\(Top-Front)	\end{tabular}}}
		& Inject Perturbation
		& $8.183$& $16.629$    & $12.134$ & $0.636$   & $0.139$  & $-1.031$ & $0.75(4.05)$ & $\bm{0.56}(\bm{3.33})$ & $17.85(15.46)$ & $1.06(6.01)$ \\
		& Initialization
		& $1.320$ & $7.264$    & $3.011$ & $-0.324$  & $0.227$& $0.000$ & $0.67(3.45)$ & $\bm{0.60}(\bm{3.23})$ & $11.72(8.37)$ & $0.90(4.06)$  \\
		& CAD Model               
		& $0.000$ & $10.000$    & $0.000$ & $0.795$   & $0.000$  & $-1.364$ & $0.48(\bm{2.29})$ & $\bm{0.43}(2.56)$   & $12.95(5.40)$  & $0.73(2.44)$ \\
		\bottomrule
	\end{tabular}
	\label{tab:add_calibration_error_slam}
\end{table*}

\begin{figure}[t]
	\centering
	\subfigure[Trajectories in a down view.]
	{\label{fig:rhd_inject_uct_1}\centering\includegraphics[width=0.2112\textwidth]{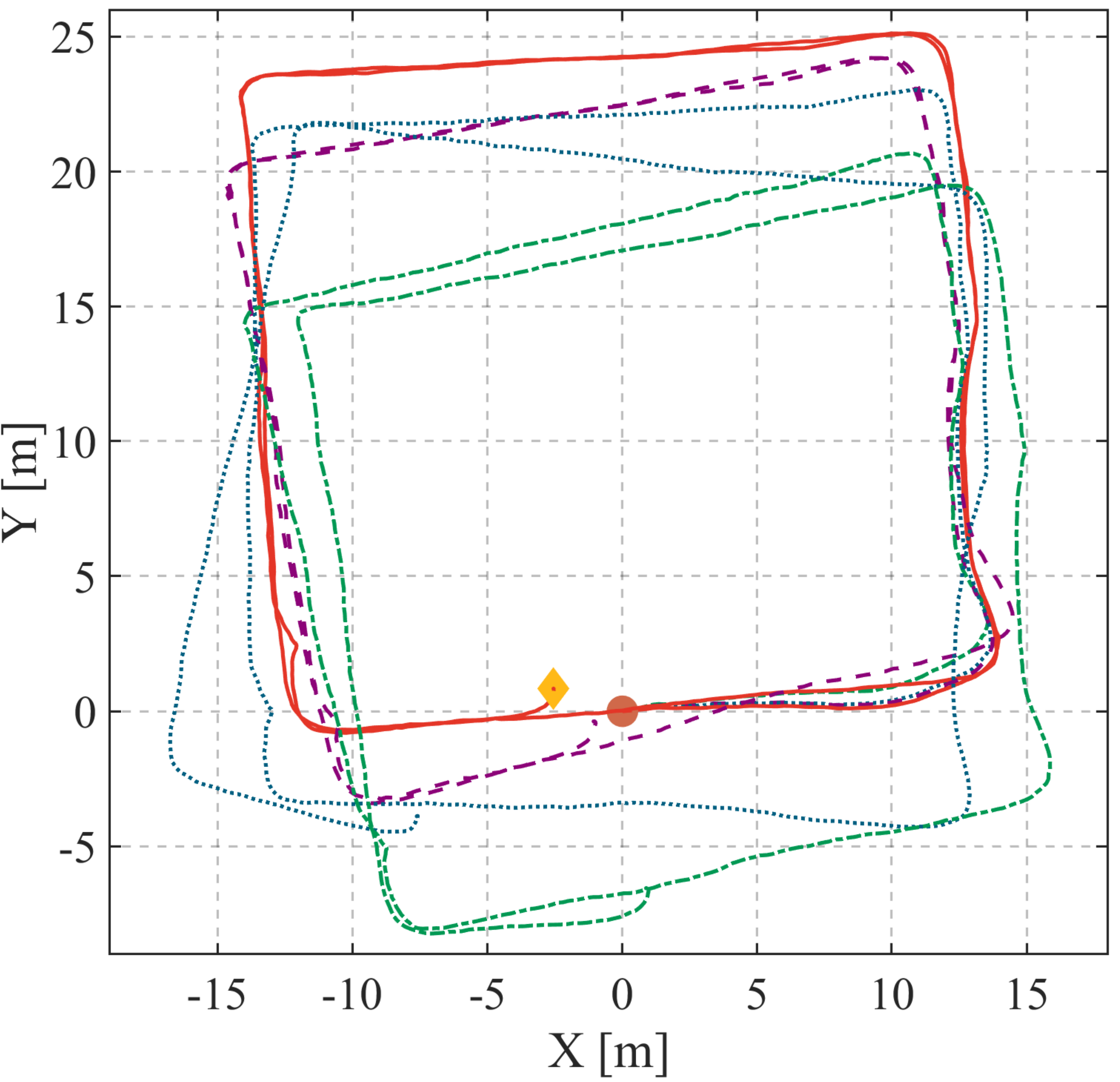}}     
	\hfill
	\subfigure[Trajectories in another view.]
	{\label{fig:rhd_inject_uct_2}\centering\includegraphics[width=0.256\textwidth]{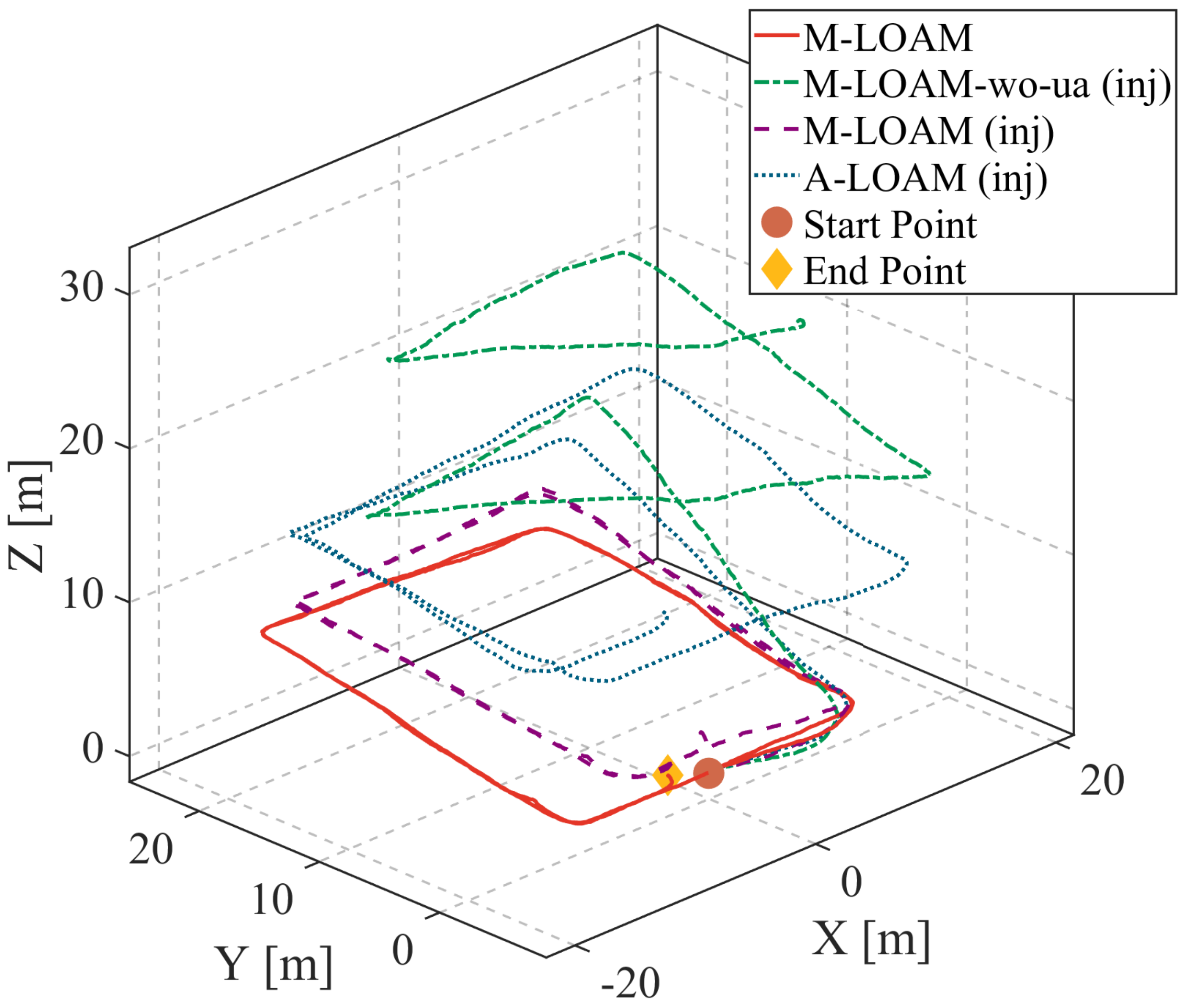}}
	\caption{Trajectories on \textit{RHD02lab} with being injected by a large extrinsic perturbation. The detailed settings are shown in Table \ref{tab:add_calibration_error_slam}.}
	\label{fig:rhd_inject_uct}  
\end{figure} 

\begin{figure}[t]
	\centering
	\includegraphics[width=0.45\textwidth]{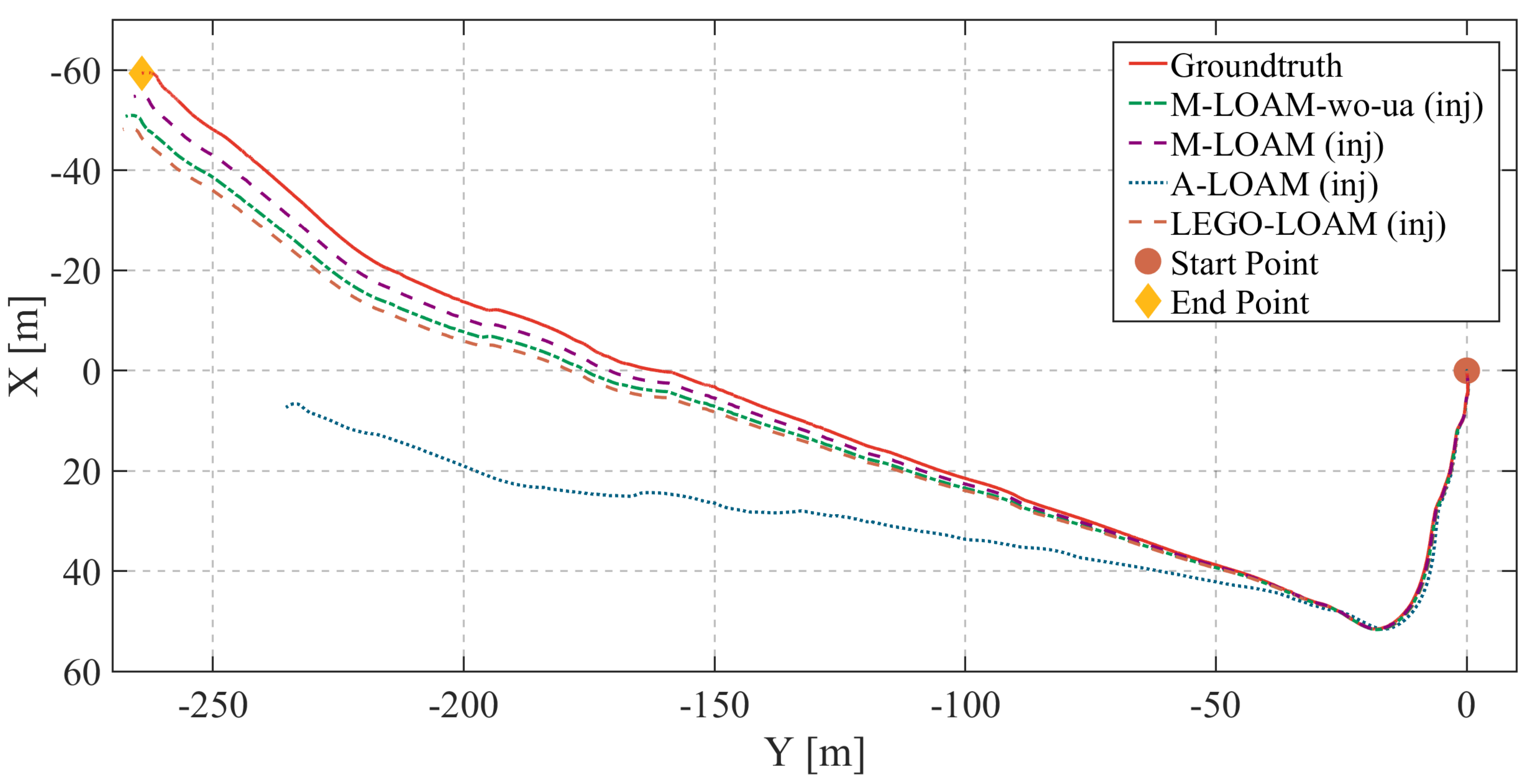}
	\caption{Trajectories on $341m$-length sequence (a part of the RV sequence) injected with a large extrinsic perturbation. The detailed settings are shown in Table \ref{tab:add_calibration_error_slam}.}
	\label{fig:rv_inject_uct}
\end{figure} 

\subsection{Sensitivity to Noisy Extrinsics}
\label{sec:inject_calibration_error}
In this section, we evaluate the sensitivity of M-LOAM to different levels of extrinsic perturbation.
On the RHD and RV platforms, we test our method by setting the extrinsics with different levels of accuracy: CAD model, initialization, and perturbation injection. 
The experiment settings are listed in Table \ref{tab:add_calibration_error_slam}.
The injected perturbation is the simulated shock on the ground truth extrinsics with $[10, 10, 10]deg$ in roll, pitch, and yaw and $[0.1, 0.1, 0.1]m$ along the $x-$, $y-$, and $z-$ axes.
We use \textit{RHD02lab} and a partial sequence on RV to compare M-LOAM with the baseline methods.
It should be noted that extrinsic calibration is turned off, and we only use the top and front LiDAR on the vehicle in experiments.
The estimated trajectories under the largest perturbation are shown in Fig. \ref{fig:rhd_inject_uct} and Fig. \ref{fig:rv_inject_uct} for different platforms.
These methods are marked with `(inj)'.
We calculate the ATE in Table \ref{tab:add_calibration_error_slam}.
Here, M-LOAM's trajectory on \textit{RHD02lab} in Section \ref{sec:indoor_exp} is used to compute the error.
We observe that all methods' performance degrades along with the increasing extrinsic perturbation.
But both M-LOAM-wo-ua and M-LOAM have smaller error. 
In particular, under the largest perturbation, M-LOAM is much more robust since it can track sensors' poses.

%% file: exper_ablation_study.tex
\begin{table}[t]
	\centering
	\caption{Average feature number and running time on a desktop of M-LOAM on the RV sequence with different LiDAR setups.}
	\renewcommand\arraystretch{1.25}
	\renewcommand\tabcolsep{2pt}	
	\begin{tabular}{crrrr}
		\toprule
		Setup    & One-LiDAR & Two-LiDAR & Three-LiDAR & Four-LiDAR\\ 
		\midrule
		Edge features    & $1061$        & $1366$        & $1604$          & $1851$     \\ 
		Planar features  & $4721$        & $5881$        & $7378$          & $8631$    \\ 
		\midrule		
		Measurement $[ms]$ & $4.6\pm0.4$        & $4.8\pm0.5$        & $5.5\pm0.7$          & $6.0\pm0.7$    \\ 
		Odometry $[ms]$ & $27\pm5$        & $57\pm7$        & $69\pm7$          & $73\pm9$         \\ 
		Mapping $[ms]$ & $78\pm11$        & $91\pm13$        & $111\pm16$          & $126\pm15$         \\ 
		\bottomrule
		\multicolumn{5}{l}{*Specifications: Intel i7 CPU@4.20 GHz and 32 GB RAM}
	\end{tabular}
	\label{tab:multi_lidar_test}
	\vspace{-0.4cm}	
\end{table}

\begin{figure*}[t]
	\centering
	\subfigure[RPE in the case of $3m/s$. From one to four LiDARs,
	the median values of the rotation and translation error (in percentage) in odometry are:
	($56.53$deg, $20.19\%$),
	($54.53$deg, $19.29\%$),	
	($48.08$deg, $17.57\%$), 
	($\textbf{42.96}$\textbf{deg}, $\textbf{15.73\%}$) respectively,
	while those in mapping are:
	($6.27$deg, $2.35\%$),
	($6.26$deg, $2.16\%$),		
	($6.77$deg, $2.32\%$),		
	($\textbf{6.45}$\textbf{deg}, $\textbf{2.14\%}$) respectively.
	]	
	{\label{fig:rv05abl_rpe}\centering\includegraphics[width=0.485\textwidth]{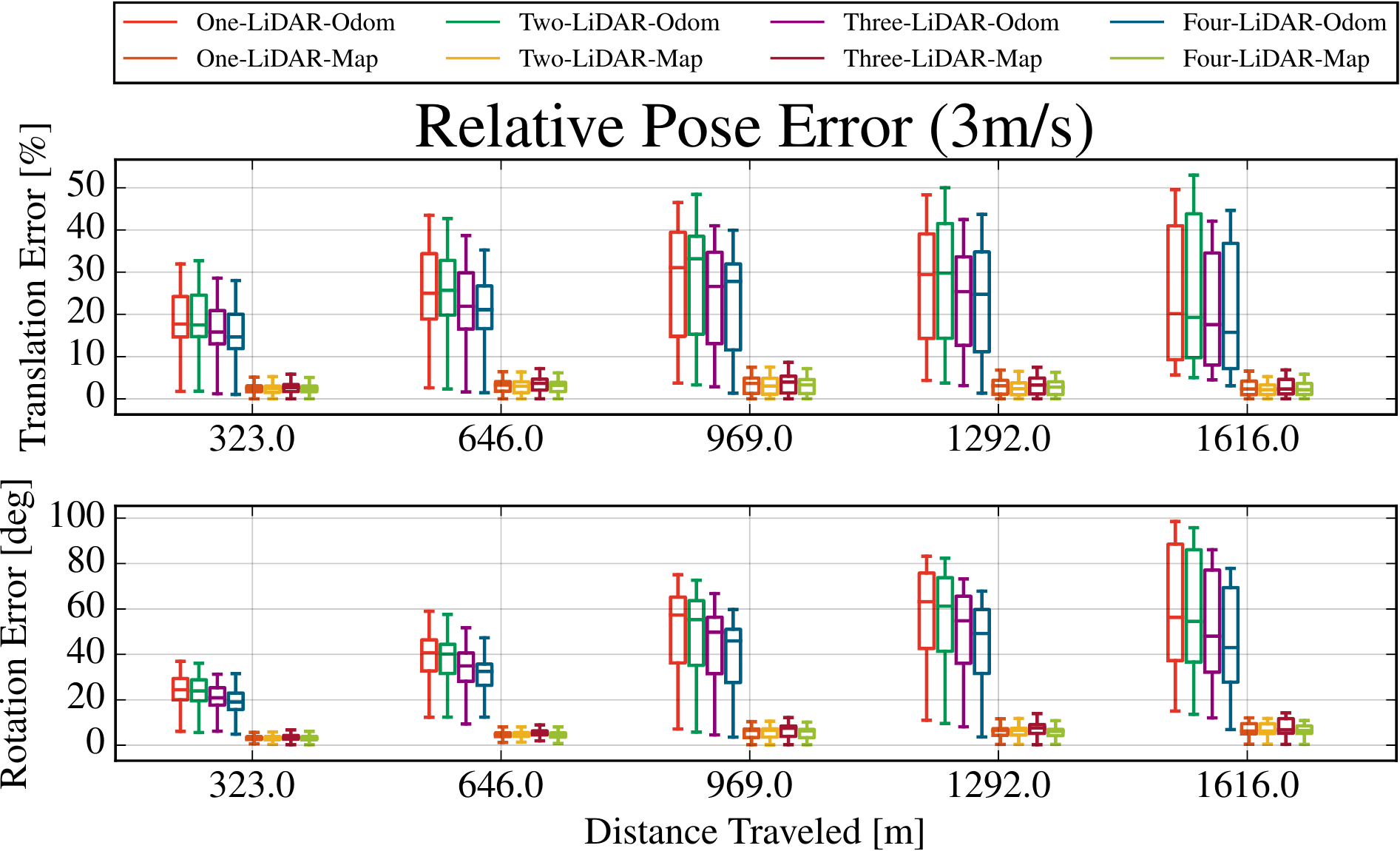}}
	\hfill	
	\subfigure[RPE in the case of  $9m/s$. From one to four LiDARs,
	the median values of the rotation and translation error (in percentage) in odometry are:
	($29.38$deg, $11.56\%$),
	($28.67$deg, $11.00\%$),	
	($24.60$deg, $9.83\%$), 
	($\textbf{21.02}$\textbf{deg}, $\textbf{8.01\%}$) respectively,
	while those in the mapping are:
	($6.86$deg, $2.42\%$),
	($6.43$deg, $2.19\%$),		
	($7.04$deg, $2.24\%$),		
	($\textbf{6.06}$\textbf{deg}, $\textbf{2.11\%}$) respectively.
	]	
	{\label{fig:rv06abl_rpe}\centering\includegraphics[width=0.485\textwidth]{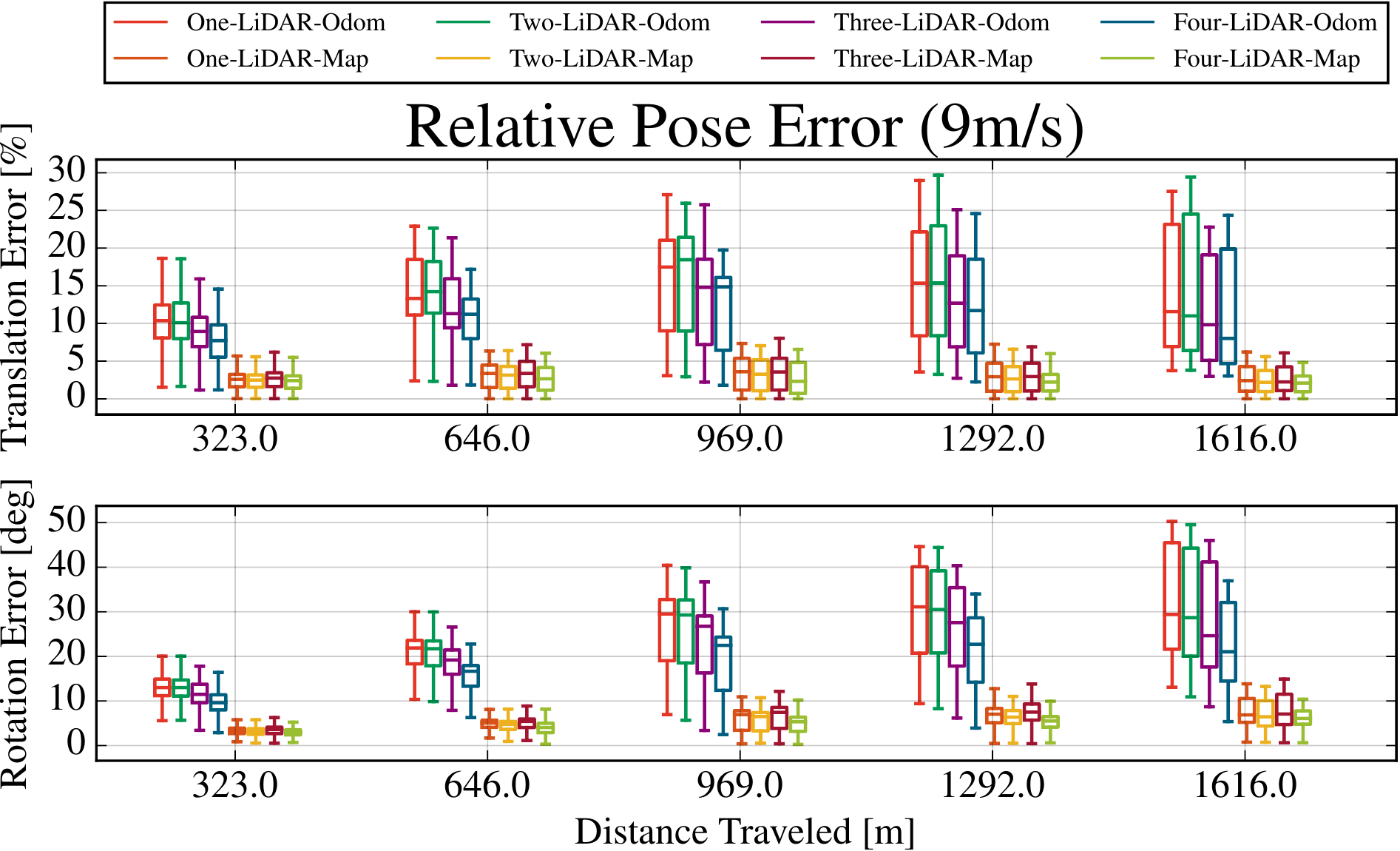}}
	\caption{RPE of M-LOAM on the RV sequence with different numbers of LiDARs in two cases. 
			Better visualization in the colored version.}
	\label{fig:rv_abl_study}
\end{figure*}  

\subsection{Single LiDAR v.s. Multiple LiDARs}
\label{sec:exp_ablation_study}

In this section, we explore the specific improvements in utilizing more LiDARs in M-LOAM. 
The RV platform has four LiDARs. 
We use One-, Two-, Three-, Four-LiDAR to denote the setups of $l^{1}$, $l^{1,2}$, $l^{1,2,3}$, and $l^{1,2,3,4}$ respectively (Fig. \ref{fig:rv_device}).
We also use x-Odom and x-Map to denote results provided by the odometry and mapping using different setups, respectively.  
The tests are carried out on the complete RV sequence.
We first report statistics of the program in Table \ref{tab:multi_lidar_test}, including the average number of edge and planar features as well as average running time of measurement processing, optimization with pure odometry, and uncertainty-aware mapping.
We see that the odometry time increases with more LiDARs because the system needs to handle more geometric constraints.
This phenomenon does not appear in measurement preprocessing since we parallelize this module.
The mapping time does not grow linearly because we use the voxel grid filter to bound the map's complexity.
We also provide the running time on an Intel NUC (i7 CPU@3.1GHz) in the supplementary material \cite{jiao2020supplementary},
where the results are consistent with Table \ref{tab:multi_lidar_test}.

To demonstrate that more features boost the system performance, we conduct experiments in two cases: driving at a normal speed and high speed. 
We use the original RV sequence in the first case. 
To simulate that the vehicle is moving faster in the second case, we extract one frame from every three frames to construct a new dataset.
We evaluate the odometry and mapping of these setups in Fig. \ref{fig:rv_abl_study}. 
The errors of the odometry decrease if more LiDARs are used. 
In the second case, the median values of Four-LiDAR-Odom are smaller than those of One-LiDAR-Odom around $8deg$ relative rotation error and $3.5\%$ translation error.
But this improvement in mapping is small because the map already provides sufficient constraints.
When the vehicle is moving at a higher speed, the global map becomes sparser. 
Consequently, the improvement of mapping on multiple LiDARs is noticeable.
The boxplot of the Four-LiDAR setup has a smaller variance than others.

%% file: discussion.tex
\section{Discussion}
\label{sec:discussion}

\subsection{Main Advantages}
We highlight that M-LOAM is a robust, reliable, and complete system to provide accurate calibration, odometry, and mapping results for different multi-LiDAR systems.
We can extend M-LOAM to many types of LiDAR combinations, as shown in experiments.
A typical application of M-LOAM is autonomous driving, where the multi-LiDAR system is gradually becoming a standard setup on vehicles.
As verified in the experiments, the usage of multiple LiDARs boosts the SLAM performance in both robustness and accuracy. 
For other perception problems such as 3D object detection \cite{jiao2020mlod} and tracking \cite{zhang2020robust}, the multi-LiDAR systems are also beneficial.

As compared with the SOTA, M-LOAM introduces the sliding window-based tightly-coupled odometry to fuse multiple LiDARs 
and the uncertainty-aware mapping to maintain the globally consistent map with good noise-singal ratio. 
Furthermore, rather than operating calibrated and merged point clouds directly, it processes the multi-LiDAR measurements in an separate way. 
This design is advantageous in several aspects: 
\textit{1)} programs (e.g., segmentation and feature extraction) can be easily parallelized, 
\textit{2)} the LiDAR's scan models can be used to generate a range image without data loss, and 
\textit{3)} the extrinsic perturbation on the system can be formulated.
We consider that the above improvements enable M-LOAM to outperform A-LOAM on most sequences.
Compared with LEGO-LOAM, which uses ground features, M-LOAM is more applicable to diverse applications.
Nevertheless, integrating M-LOAM with the ground-optimization pipeline of LEGO-LOAM for mobile robots is encouraging.

The Gaussian distribution is our core hypothesis in modeling data uncertainty.
Based on it, we use a tractable method to estimate covariances of poses (derived from information matrices) and extrinsics (given a sampling covariance after calibration).
Even though these covariances are approximate, as shown in experiments, e.g., Fig. \ref{fig:rhd_inject_uct}, 
the proposed uncertainty-aware operation significantly improves the robustness of M-LOAM against degeneracy and extreme extrinsic perturbation.

\subsection{Limitations}
We recognize that the proposed calibration and SLAM methods have limitations.
First, the calibration process requires some pre-set thresholds which are obtained from experiments.
Its accuracy is not perfect for applications such as HD map construction. 
In practice, the errors should be smaller than $0.01m$ and $1deg$.
Otherwise, the calibration errors are proportionally propagated to the map and deteriorate the map quality.
This effect cannot be entirely eliminated even though the extrinsic perturbation is modeled by our method.

Second, our system utilizes several point cloud registrations in different phases to estimate states. 
As a typical non-convex problem, registration requires a good initial transformation.
But LiDARs only produce a low-frequency data stream, making this problem sometimes challenging.
For example, when a robot moves and turns at a high frequency, our method barely tracks its poses.
Also, we use the linear model to interpolate sensors' poses, which cannot represent smooth or fast motion well. 
In these cases, it would be better to use high-order curves such as B-spline for interpolation \cite{park2020elasticity}.

Finally, we extract the simple edge and planar points from environments. However, these features present drawbacks in real tests.
For instance, they only provide constraints in their perpendicular directions. 
In a long tunnel, where all planes are mostly parallel, M-LOAM may fail.
Another example is that such features do not have enough recognition power to enable robust matching across frames with large viewpoint changes.
As compared with surfel-based or visual features, they are less useful for tasks such as place recognition and relocalization.

%% file: conclusion.tex
\section{Conclusion and Future Work}
\label{sec:conclusion}

In this paper, we propose a complete and robust solution for multi-LiDAR extrinsic calibration and SLAM.
This approach contains several desirable features, including fast segmentation for noise removal, motion and extrinsic initialization, 
online extrinsic calibration with convergence identification, a tightly coupled M-LO, and uncertainty-aware multi-LiDAR mapping. 
We conduct extensive experiments covering scenarios from indoor offices to outdoor urban roads for evaluation.
Our approach calibrates kinds of multi-LiDAR systems for different platforms. It yields accuracies centimeters in translation and deci-degrees in rotation and is comparable to a SOTA target-based method.
For SLAM, the proposed system typically reaches a localization accuracy below $40$cm in medium-scale ($>150m$) scenarios and of a few meters in the large-scale urban roads ($>3.2km$).
For the benefit of the community, we make our implementation open-source. 

There are several directions for future research. 
Adding a loop-closure module into our system is desirable, which helps to correct the accumulated drift and keep the global map \cite{chen2020overlapnet}.
Another research direction concerns object-centric SLAM. 
Two challenges are recently growing in the community.
On the one hand, the widely used low-level geometric features are not representative and sensitive to viewpoint change.
On the other hand, data sparsity and occlusion in LiDAR-based object detectors are the dominant bottlenecks.
A possible solution to them is to develop a SLAM approach which can use object-level features to optimize both ego-motion and motion of dynamic objects.
Trials on cameras or visual-inertial systems have been proposed in \cite{yang2019cubeslam,zhang2020vdo,qin2020avp}, while works on LiDARs are rare.
Finally, extending our approach on calibration and uncertainty modeling to sensors in various modalities, e.g., IMUs \cite{ye2019tightly}, radars \cite{barnes2020oxford} and event-cameras \cite{gallego2020event}, is promising.
For instance, we can propagate the IMU noise model to predict pose uncertainties, or the proposed convergence criteria can be used for the extrinsic calibration of multi-modal sensors.

%% file: appendix.tex
\appendix

\subsection{Jacobians of Residuals}
\label{app.jacobian_initialization}
The state vector is defined as $\mathbf{x}=[\mathbf{t}, \mathbf{q}]$. We convert $\mathbf{q}$ into a rotation matrix $\mathbf{R}$ by the Rodrigues formula \cite{sola2017quaternion}:
\begin{equation}
	\mathbf{R}
	=
	(q_{w}^{2} - \mathbf{q}_{xyz}^{\top}\mathbf{q}_{xyz})\mathbf{I}
	+
	2\mathbf{q}_{xyz}\mathbf{q}_{xyz}^{\top}
	+
	2q_{w}\mathbf{q}_{xyz}^{\wedge},
\end{equation}

\subsubsection{Jacobians of $\mathbf{r}_{\mathcal{H}}$}
The residuals in \eqref{equ:objective_initialization} are rewritten as
\begin{equation}
\begin{aligned}
	\mathbf{r}_{\mathcal{H}}(\mathbf{x},\mathbf{p})
	&=
	\big[
	\mathbf{w}^{\top}(\mathbf{R}\mathbf{p}+\mathbf{t}) + d
	\big]
	\mathbf{w}\\
	&=
	\text{diag}(\mathbf{w})
	\begin{bmatrix}
	\mathbf{w} & \mathbf{w} & \mathbf{w}
	\end{bmatrix}^{\top}	
	(\mathbf{R}\mathbf{p}+\mathbf{t})+d\mathbf{w}\\
	&=
	\mathbf{W}(\mathbf{R}\mathbf{p}+\mathbf{t})+d\mathbf{w},
\end{aligned}	
\end{equation}

Using the left perturbation: $\mathbf{R}\exp(\delta\bm{\phi}^{\wedge})\approx\mathbf{R}(\mathbf{I}+\delta\bm{\phi}^{\wedge})$,
the Jacobians of the rotation and translation are calculated as
\begin{equation}
\label{equ:residual_jacobian}
\begin{aligned}
	\frac{\partial \mathbf{r}_{\mathcal{H}}(\mathbf{x},\mathbf{p})}{\partial\mathbf{x}}
    &= 
    [
   	\frac{\partial \mathbf{r}_{\mathcal{H}}(\mathbf{x},\mathbf{p})}{\partial\mathbf{t}},\ 
	\frac{\partial \mathbf{r}_{\mathcal{H}}(\mathbf{x},\mathbf{p})}{\partial\mathbf{q}}
   	],\\
    &= 
	[\mathbf{W},\ 
 	 -\mathbf{W}\mathbf{R}\mathbf{p}^{\wedge},\  \mathbf{0}_{3\times1}].
\end{aligned}
\end{equation}
where the quaternion is updated according to
$\delta\mathbf{q}\approx[\frac{1}{2}\delta\bm{\phi}, 1]^{\top}$.

\subsubsection{Jacobians of residuals in $f_{\mathcal{M}}$ for Online Calibration}
The objective function in \eqref{equ:objective_online_calibration} has two terms: $f_{\mathcal{M}}(\mathcal{X}_{v})$ and $f_{\mathcal{M}}(\mathcal{X}_{e})$. For the first term, the Jacobians are given by
\begin{equation}
\begin{aligned}
	\frac{\partial \mathbf{r}_{\mathcal{H}}(\mathbf{x}^{-1}_{p}\mathbf{x}_{k}, \mathbf{p})}{\partial\mathbf{x}_{k}}
	&=
	[\mathbf{W}\mathbf{R}_{p}^{\top},	    
	-\mathbf{W}\mathbf{R}_{p}^{\top}\mathbf{R}_{k}\mathbf{p}^{\wedge},
	\mathbf{0}_{3\times1}],\\
\end{aligned}
\end{equation}
where $	k \in [p+1, N+1]$.
Since the second term has the same form as \eqref{equ:objective_initialization}, the Jacobians are given by \eqref{equ:residual_jacobian} as
\begin{equation}
\begin{aligned}
	\frac{\partial \mathbf{r}_{\mathcal{H}}(\mathbf{x}^{b}_{l^{i}}, \mathbf{p})}{\partial\mathbf{x}^{b}_{l^{i}}}
	&=
	[\mathbf{W},\ 
	-\mathbf{W}\mathbf{R}_{l^{i}}^{b}\mathbf{p}^{\wedge},\ 
	\mathbf{0}_{3\times1}],
\end{aligned}
\end{equation}
where $i\in[2, I]$.

\subsubsection{Jacobians of residuals in $f_{\mathcal{M}}$ for Pure Odometry}
The Jacobians of the residuals in \eqref{equ:objective_pure_odometry} are computed as
\begin{equation}
\begin{aligned}
	&\frac{\partial \mathbf{r}_{\mathcal{H}}s(
		\mathbf{x}_{p}^{-1}
		\mathbf{x}_{k}
		\mathbf{x}^{b}_{l^{i}}, \mathbf{p})}{\partial \mathbf{x}_{k}}
	=\\
	&\ \ \ \ \ \ \ \ \ \ \ \ 
	[\mathbf{W}\mathbf{R}_{p}^{\top},\ 
	-\mathbf{W}\mathbf{R}_{p}^{\top}\mathbf{R}_{k}(\mathbf{R}_{l^{i}}^{b}\mathbf{p} + \mathbf{t}^{b}_{l^{i}})^{\wedge},\ 
	\mathbf{0}_{3\times1}],\\	
\end{aligned}
\end{equation}
where $i \in [1, I]$ and $k \in [p+1, N+1]$.

\subsection{Marginalization}
\label{app.marginalization}
The sliding-window estimator need to margialize out several states and add new states after optimization.
For the whole state vector $\mathcal{X}$, 
we denote $\mathcal{X}_{m}$ as the set of marginalized states and
$\mathcal{X}_{r}$ as the set of remaining states.
By linearizing and expanding the cost function \eqref{equ:objective_function_multilo_whole} at an initial point, 
we obtain the \textit{normal equation}: $\bm{\Lambda}\delta\mathcal{X}=-\mathbf{g}$, 
where $\bm{\Lambda}=\sum_{}^{}\mathbf{J}^{\top}\bm{\Sigma}^{-1}\mathbf{J}$ is the \textit{information matrix}, 
$\mathbf{g}=\mathbf{J}^{\top}\bm{\Sigma}^{-1}\mathbf{r}$.
By representing the equation using block matrices, we have
\begin{equation}
\begin{aligned}
	\begin{bmatrix}
		\bm{\Lambda}_{mm} & \bm{\Lambda}_{mr}\\
		\bm{\Lambda}_{rm} & \bm{\Lambda}_{rr}
	\end{bmatrix}
	\begin{bmatrix}
		\delta\mathcal{X}_{m}\\
		\delta\mathcal{X}_{r}		
	\end{bmatrix}	
	=
	-
	\begin{bmatrix}
	\mathbf{g}_{m}\\
	\mathbf{g}_{r}		
	\end{bmatrix}.
\end{aligned}	
\end{equation}
where we apply the Schur complement to yield:
\begin{equation}
	\begin{bmatrix}
		\bm{\Lambda}_{mm} & \bm{\Lambda}_{mr}\\
		\mathbf{0} & \bm{\Lambda}_{rr}^{*}
	\end{bmatrix}
	\begin{bmatrix}
		\delta\mathcal{X}_{m}\\
		\delta\mathcal{X}_{r}		
	\end{bmatrix}	
	=
	-
	\begin{bmatrix}
		\mathbf{g}_{m}\\
		\mathbf{g}_{r}^{*}		
	\end{bmatrix},
\end{equation}
where 
\begin{equation}
\label{equ:app_marginalization_value}
\begin{aligned}
	\bm{\Lambda}_{rr}^{*} 
	&=
	\bm{\Lambda}_{rr} - \bm{\Lambda}_{rm}\bm{\Lambda}_{mm}^{-1}\bm{\Lambda}_{mr},\\
	\mathbf{g}_{r}^{*}
	&=
	\mathbf{g}_{r} -
	\bm{\Lambda}_{rm}\bm{\Lambda}_{mm}^{-1}\mathbf{g}_{m}.
\end{aligned}	
\end{equation}


The resulting $\bm{\Lambda}_{rr}^{*}$ and $\mathbf{g}_{r}^{*}$ encode the dependency of the marginalized states. 
Taking the linear residual into the next optimization, 
we can maintain the consistency of $\mathcal{X}_{r}$.
Since the Ceres solver \cite{ceres-solver} uses Jacobians to update variables, when implementing the marginalization, 
we rewrite the \textit{information matrices} using Jacobians.
After obtaining $\bm{\Lambda}_{rr}^{*}$, 
we factorize it with eigenvalues and eigenvectors:
\begin{equation}
	\bm{\Lambda}_{rr}^{*} = \mathbf{P}\bm{\Psi}\mathbf{P}^{\top}.
\end{equation}

Let $\mathbf{J}^{*}=\sqrt{\bm{\Psi}}\mathbf{P}^{\top}$, $\mathbf{r}^{*}=\sqrt{\bm{\Psi}^{-1}}\mathbf{P}^{\top}\mathbf{g}_{r}^{*}$, we have 
\begin{equation}
\begin{aligned}
	\mathbf{J}^{*\top}\mathbf{J}^{*}
	=
	\bm{\Lambda}_{rr}^{*}, \ \ \
	\mathbf{J}^{*\top}\mathbf{r}^{*}
	=
	\mathbf{g}_{r}^{*}.
\end{aligned}	
\end{equation}

In the next optimization, the prior residual 
$\|\mathbf{r}_{pri}(\mathcal{X}_{r})\|^{2}$ is equal to
$\|\mathbf{r}^{*} + \mathbf{J}^{*}\Delta\mathcal{X}_{r}\|^{2}$, where 
$\Delta\mathcal{X}_{r}$ is the ``distance'' between the current state and inital state.

%% file: acknowledgment.tex
\section*{Acknowledgment}
The authors would like to thank members in HKUST Robotics Institute and Shenzhen Unity-Drive Innovation for the help and suggestions 
in algorithm development, temporal synchronization, and data collection.
They would also like to thank the editors and anonymous reviewers of IEEE TRO for their suggestions, which led us to improve this article.